\documentclass[11pt]{article}

\usepackage[final]{acl}

\usepackage{microtype}

\usepackage{graphicx}
\usepackage{marvosym}
\usepackage{makecell}
\usepackage{booktabs}
\usepackage{xltabular}
\usepackage{rotating}
\usepackage{pfnote}

\usepackage[english,bidi=default]{babel} 
\babelfont{rm}{TeXGyreTermesX} 
\babelprovide[import]{tibetan}
\babelfont[tibetan]{rm}{NotoSerifTibetan.ttf}

\title{{\protect\includegraphics[height=3.3\fontcharht\font`\0]{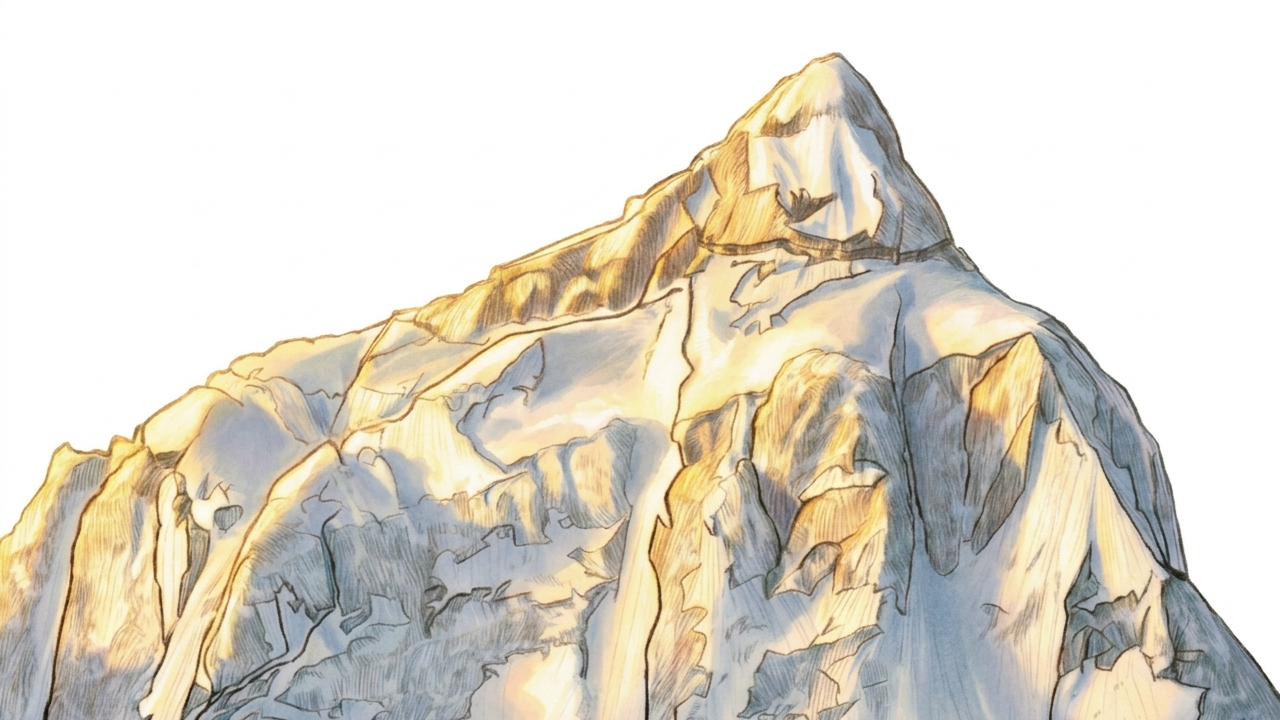}} Towards the \textit{Roof of the World}: NLP Progress in\\Tibetan Languages}

\author{
	\textbf{Xi Cao$^{\heartsuit1,\heartsuit3,\spadesuit}$}\textbf{,}~
	\textbf{Yijie Li$^{\heartsuit2,\spadesuit}$}\textbf{,}~
	\textbf{Yuan Sun$^{\heartsuit2,\heartsuit3,\spadesuit,}$\textsuperscript{\Letter}}\textbf{,}~\\
	\textbf{Shiwei Dong$^{\diamondsuit,\clubsuit}$}\textbf{,}~
	\textbf{Zeguang Huang$^{\diamondsuit,\clubsuit}$}\textbf{,}~
	\textbf{Nuo Qun$^{\diamondsuit,\clubsuit}$}\\
	$^{\heartsuit1}$School of Philosophy and Religious Studies, Minzu University of China\\
	$^{\heartsuit2}$School of Information Engineering, Minzu University of China\\
	$^{\heartsuit3}$Institute of National Security, Minzu University of China\\
	$^{\spadesuit}$National Language Resource Monitoring \& Research Center | Minority Languages Branch\\
	$^{\diamondsuit}$School of Information Science and Technology, Xizang University\\
	$^{\clubsuit}$The State Key Laboratory of Tibetan Intelligence\\
	\normalsize{\texttt{caoxi@muc.edu.cn},
		\texttt{sunyuan@muc.edu.cn}}
	\thanks{\Letter~Corresponding Author}
}

\makeatletter
\def\thanks#1{\protected@xdef\@thanks{\@thanks\protect\footnotetext{#1}}}
\makeatother

\newcommand{\cutoffdate}{00:00 UTC, September 25, 2026}

\begin{document}

\maketitle

\begin{abstract}
Tibetan, spoken by some seven million people on and around the Qinghai-Xizang Plateau, remains underrepresented in natural language processing (NLP). We present, to our knowledge, the first peer-reviewed systematic bibliometric survey of Tibetan NLP: 140 publications from 2004 to 2026 in the ACL Anthology and DBLP, annotated for venue type, research area, first-affiliation country, and resource release. Output has grown rapidly, with 84\% of the publications appearing since 2018, and has expanded bottom-up along the NLP pipeline, from corpora and word segmentation through script, speech, and applications to large language models (LLMs). Researchers based in China first-authored 87\% of the publications, which span all 25 represented research areas, whereas work from other countries concentrates on Classical Tibetan. We argue that this full-pipeline paradigm, realized under the compounded constraints of a complex script, dialectal diversity, and geographic remoteness, offers a reference model for other low-resource languages of the Global South. We also identify gaps in dialect coverage, responsible NLP, native and comparable evaluation, cross-community visibility, and resource release (only 29\% of the publications release data, code, or models), and recommend open science, shared standards, and community participation for the LLM era.
\end{abstract}

\section{Introduction}
\label{sec_intro}

The benefits of NLP remain unevenly distributed: most of the world's languages have little or no labelled data \citep{joshi-etal-2020-state}, and LLMs, whose capabilities depend on each language's digital text, threaten to widen this gap. The risk is compounded in remote highlands, where geographic isolation, dialectal fragmentation, and limited digital infrastructure add to data scarcity; the Qinghai-Xizang Plateau, the world's highest, known as the \textit{Roof of the World} and the \textit{Third Pole}, is an extreme instance.

Tibetan, the principal language of the plateau, is spoken by some seven million people in the Xizang Autonomous Region and the provinces of Qinghai, Sichuan, Gansu, and Yunnan in China, and by communities in Pakistan, India, Nepal, and Bhutan. Its script does not mark word boundaries, its spoken varieties differ substantially, and its digital text is scarce (\S\ref{sec_language}); in the taxonomy of \citet{joshi-etal-2020-state}, it belongs to Class~1 (``The Scraping-Bys''), with some unlabelled but almost no labelled data, a class that, the authors argue, needs ``a solid, organized movement'' to collect labelled data. Tibetan thus tests whether such a movement can bring a low-resource language into the LLM era.

Although Tibetan NLP has been studied for more than two decades, its literature is fragmented across communities and venues: 103 of the 140 publications we identify (74\%) lie outside the ACL Anthology, the sole source of a recent survey of Sino-Tibetan languages \citep{liu-best-2025-survey}, and the closest Tibetan-specific overview is a brief editorial \citep{congjun_recent_2021}. We address this gap with a bibliometric survey of these publications (\S\ref{sec_method}), guided by four research questions:

\paragraph{RQ1.} How has Tibetan NLP output evolved over time and across venues? (\S\ref{sec_trends})

\paragraph{RQ2.} Which research areas have received the most and least attention, and how have priorities shifted? (\S\ref{sec_areas})

\paragraph{RQ3.} Which countries have contributed, and what role has China played? (\S\ref{sec_country})

\paragraph{RQ4.} What are the key open challenges, and what could accelerate progress? (\S\S\ref{sec_discussion}--\ref{sec_recommendations})

Our central finding is that Tibetan NLP exemplifies what we call a \textit{full-pipeline paradigm}: starting from annotated corpora, lexicography, and word segmentation, the field has successively added script and speech processing, applications, and, since 2022, pre-trained and large language models, while continuing to build resources. We argue that this paradigm offers a reference model for other low-resource languages of the Global South, identify the gaps to be closed in the LLM era, and release our annotated bibliography with an audit of publicly released resources.

\section{Background}
\label{sec_background}

\subsection{The Tibetan Language}
\label{sec_language}

Tibetan belongs to the Tibeto-Burman branch of the Sino-Tibetan language family. Written continuously since the seventh century, its literary language, rooted in Classical Tibetan, serves as a shared written standard across three main dialect groups, Ü-Tsang, Kham, and Amdo, whose spoken forms differ so much in phonology that mutual intelligibility is limited \citep{schmidt_grading_2020,ren_research_2026}; even lexical tone varies, from tonal Ü-Tsang through Kham to largely atonal Amdo \citep{liang-zerong-2026-tonogenesis}. We use \textit{Tibetan} collectively for the classical and modern written language and its spoken varieties, often classified as distinct Tibetic languages (hence the plural in our title). This diglossia challenges NLP: models trained on written text may transfer poorly to speech, and variety-specific systems generalize poorly.

\begin{figure}[t]
	\centering
	\includegraphics[width=\linewidth]{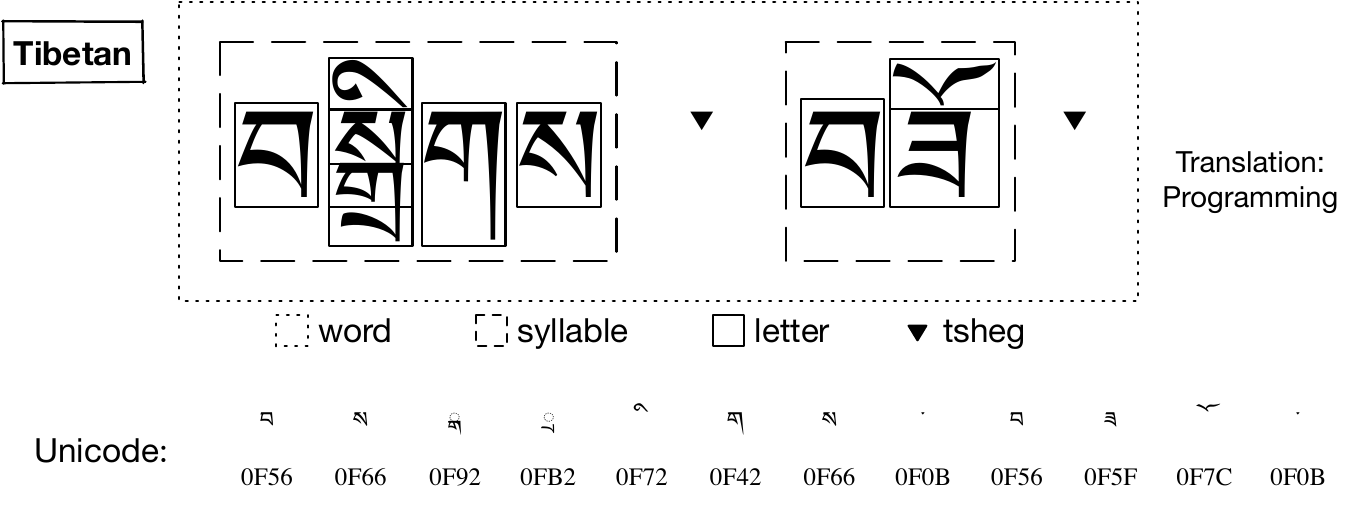}
	\captionsetup{justification=justified, singlelinecheck=false}
	\caption{The Tibetan word \foreignlanguage{Tibetan}{བསྒྲིགས་བཟོ་} (\textit{programming}), illustrating the three-level structure of Tibetan script: letters (30 consonants and 4 vowel signs, each encoded by a Unicode code point, shown at the bottom) combine into syllables according to strict orthographic rules, and syllables, each followed by the inter-syllable mark \textit{tsheg} (\foreignlanguage{Tibetan}{་}), form words whose boundaries are not marked.}
	\label{fig_tibetan_word}
\end{figure}

As Figure~\ref{fig_tibetan_word} shows, Tibetan script is an abugida in which 30 consonant letters and 4 vowel signs \citep{liu-etal-2014-zipfs} combine under strict orthographic rules to form syllables. Syllables are delimited by the \textit{tsheg}, but word boundaries are not marked, so word segmentation has long come first in text processing \citep{liu-etal-2011-tibetan}, although neural models can operate directly on syllables \citep{lai-etal-2018-tibetan}. Even syllable boundaries can be implicit: several case particles fuse with the preceding syllable and resemble ordinary final letters, a major source of errors in segmentation, syllable analysis, and subword machine translation (MT) \citep{li_study_2013,zhu_new_2019,gyatso_improved_2025}. Encoding raises further problems: unlike Unicode, Chinese national extension standards encode whole stacks, 7,198 in all, as single characters \citep{liu-etal-2014-zipfs,wang_online_2019}; stacks may be stored in variant code-point orders \citep{faggionato-meelen-2019-developing}; and replacing the \textit{tsheg} with spaces degrades fine-tuned models \citep{cao-etal-2023-pay-attention}. Tibetan is verb-final with ergative marking \citep{an_neural_2021}, case-particle forms depend on the preceding syllable's final letter \citep{khysru_tibetan_2021}, and verbs have up to four stems \citep{khysru_morphological_2019}; its conservative orthography retains prefixed, superscribed, and subscribed letters and leaves Lhasa tones unwritten \citep{yang_using_2015}, both of which grapheme-to-phoneme conversion must model. Together with scarce digital text, each property has spawned a sustained line of research (\S\ref{sec_areas}).

\subsection{Related Surveys}
\label{sec_related}

Surveys of geographically defined, under-resourced language groups, such as African languages \citep{adebara-abdul-mageed-2022-towards,alabi-etal-2025-charting}, the indigenous languages of Latin America \citep{tonja-etal-2024-nlp}, and the languages of India \citep{kumar-etal-2026-bhashasutra}, inform our design. Our survey differs from the two closest overviews of Tibetan NLP in scope, coverage of the LLM era, sources, and annotation (Table~\ref{tab_survey_comparison}).

\begin{table*}[t]
	\centering
	\small
	\renewcommand{\arraystretch}{1.1}
	\begin{tabular}{>{\bfseries}p{1.6cm} p{4.1cm} p{4.1cm} p{4.1cm}}
		\toprule
		\textbf{Feature}
		& \textbf{Our Survey}
		& \textbf{\makecell[l]{NAACL Survey\\\citep{liu-best-2025-survey}}}
		& \textbf{\makecell[l]{TALLIP Editorial\\\citep{congjun_recent_2021}}} \\
		\midrule
		
		Scope
		& \textbf{Tibetan only}
		& Sino-Tibetan family (49 low-resource languages)
		& Tibetan only \\
		
		Coverage Period
		& \textbf{2004--2026}: from the earliest identified paper through \cutoffdate
		& 1952--2022: the ACL OCL corpus \citep{rohatgi-etal-2023-acl}; venue trends for 2000--2021
		& 2016--2021: editorial for a special issue of \textit{ACM TALLIP} \\
		
		Corpus
		& \textbf{140 publications}, annotated for venue type, research area, country, and resource release
		& 253 papers (274 language--paper matches for 23 of the 49 languages), annotated for topic
		& No fixed count: narrative overview of key developments \\
		
		Sources
		& \textbf{Hybrid}: ACL Anthology BibTeX and DBLP Search API
		& ACL Anthology only, via the ACL OCL corpus
		& Selected by the special-issue editors \\
		
		\bottomrule
	\end{tabular}
	\caption{Comparison of our survey with the two most closely related overviews of Tibetan NLP.}
	\label{tab_survey_comparison}
\end{table*}

\section{Methodology}
\label{sec_method}

\paragraph{Scope and inclusion criteria.} We include peer-reviewed publications in which Tibetan, modern or classical, is the primary language of study or a principal experimental object, and whose contribution concerns computational processing of Tibetan text, speech, or script; multilingual studies qualify when Tibetan is among their principal experimental languages. Coverage runs from 2004, the year of the earliest Tibetan NLP paper in our sources, through \cutoffdate.

\begin{figure}[t]
	\centering
	\includegraphics[width=\linewidth]{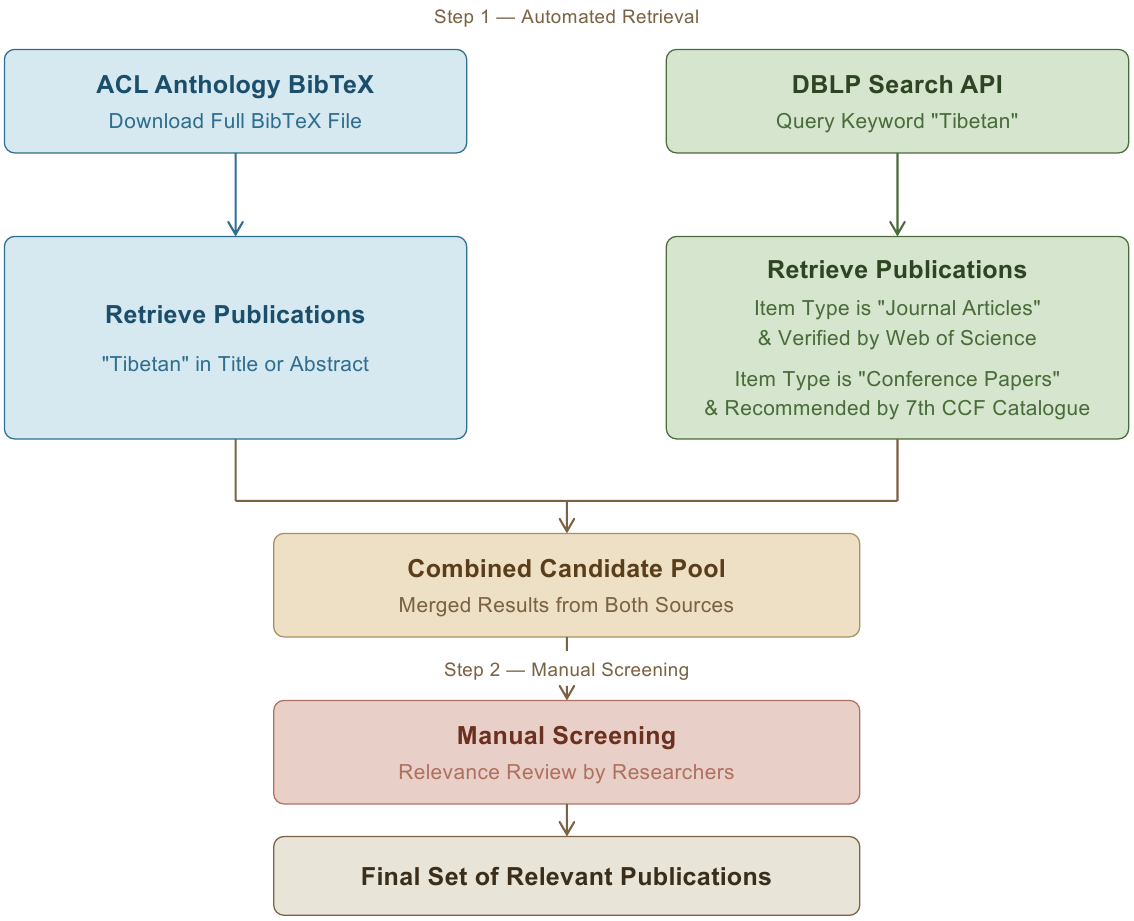}
	\captionsetup{justification=justified, singlelinecheck=false}
	\caption{Two-step construction of the survey corpus. \textbf{Step~1 (automated retrieval):} entries with \textit{Tibetan} in the title or abstract are retrieved from the full ACL Anthology BibTeX file, and publications matching the keyword \textit{Tibetan} are retrieved through the DBLP Search API, retaining journal articles verified by Web of Science and conference papers from venues recommended in the 7th edition of the CCF catalogue. \textbf{Step~2 (manual screening):} the merged and deduplicated candidate pool is reviewed for relevance, yielding the final set of 140 publications.}
	\label{fig_flowchart}
\end{figure}

\paragraph{Retrieval and screening.} We proceeded in two stages (Figure~\ref{fig_flowchart}). First, we retrieved all entries with the keyword \textit{Tibetan} in the title or abstract from the ACL Anthology's full BibTeX export\footnote{\url{https://aclanthology.org/faq/bib/}}. To cover research outside it, we queried the DBLP Search API\footnote{\url{https://dblp.org/faq/13501473.html}} with the same keyword and, to restrict results to established peer-reviewed venues, retained only Web of Science\footnote{\url{https://www.webofscience.com}} indexed journal articles and papers from international conferences recommended in the 7th edition of the China Computer Federation (CCF) catalogue\footnote{\url{https://www.ccf.org.cn/Academic_Evaluation/By_category/}}. Second, four researchers independently screened the deduplicated candidates against the inclusion criteria, considering only English-language papers and resolving disagreements by discussion. The final corpus has 140 publications: 37 from the ACL Anthology (25 conference and 12 workshop papers; Appendix \ref{appendix_acl}) and 103 from DBLP (29 conference papers, 74 journal articles; Appendix \ref{appendix_dblp}).

\paragraph{Annotation.} Each publication was annotated along three dimensions: (1)~\textit{venue type} (conference, workshop, or journal); (2)~\textit{country} of the first author's primary affiliation (Appendix \ref{appendix_country}); and (3)~\textit{research area}: up to two of the 29 areas of the ACL Rolling Review (ARR) taxonomy\footnote{\url{https://aclrollingreview.org/areas}} (Appendix \ref{appendix_research_area}). In practice, 115 publications received two areas and 25 received one, yielding 255 area assignments, so area counts sum to more than 140. We also recorded whether the authors release code, data, or models at a public URL (Appendix~\ref{appendix_open_source}).

\section{Results}
\label{sec_results}

\subsection{Publication Trends by Year and Venue}
\label{sec_trends}

\begin{figure}[h]
	\centering
	\includegraphics[width=\linewidth]{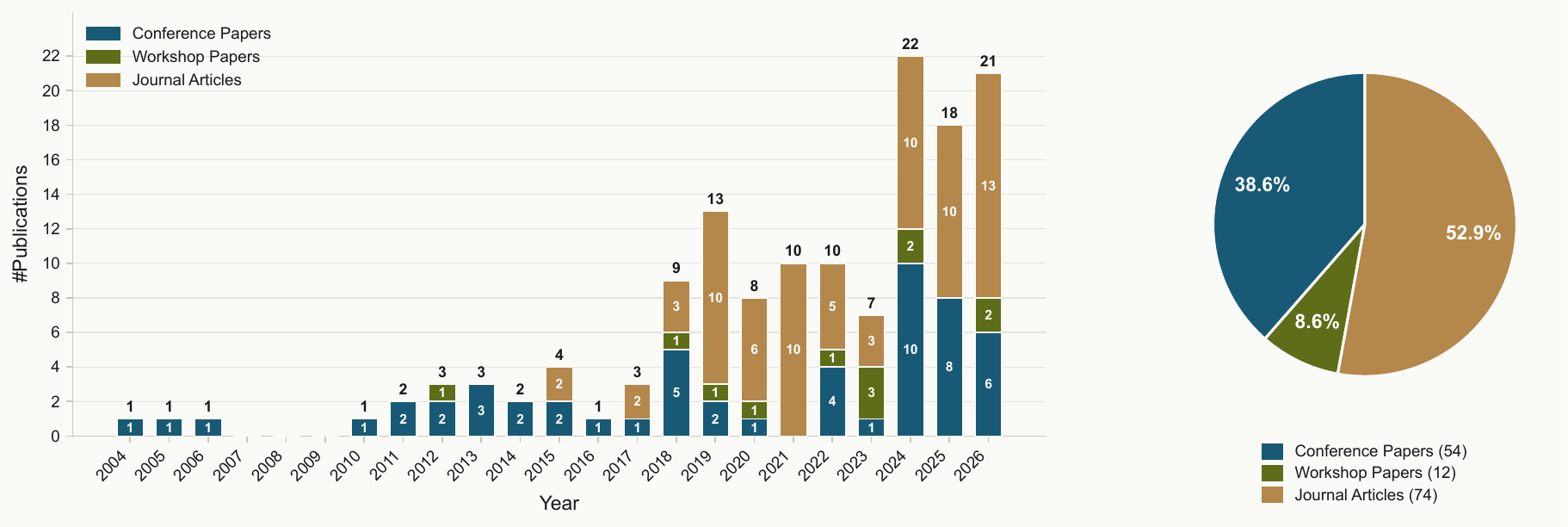}
	\captionsetup{justification=justified, singlelinecheck=false}
	\caption{Annual number of Tibetan NLP publications by venue type (left; stacked bars) and overall share of each venue type (right; counts in the legend), 2004--2026. The year 2026 is only partially covered. The full-scale version is Figure~\ref{fig_year_venue_appendix} in Appendix~\ref{appendix_figure}.}
	\label{fig_year_venue}
\end{figure}

Annual output (Figure~\ref{fig_year_venue}) was sporadic for nearly two decades, with three papers in 2004--2006, none in 2007--2009, and 19 in 2010--2017, before rising sharply: 118 publications (84.3\%) date from 2018 onward, and 61 (43.6\%) from 2024--2026 alone, although 2026 is only partially covered. This growth coincides with the wider field's adoption of deep learning and, later, of pre-trained Transformers and LLMs. Several annual peaks, however, reflect clusters in single venues: five of the nine papers of 2018 appeared at PRCV~2018, seven of the 13 of 2019 in \textit{IEEE Access}, eight of the ten of 2021 in TALLIP, six of them in the special issue introduced by \citet{congjun_recent_2021}, and five of the 21 of 2026 in one issue of \textit{Data Intelligence}. Dedicated venues have thus shaped growth, and yearly fluctuations should not be over-interpreted.

\begin{figure}[h]
	\centering
	\includegraphics[width=\linewidth]{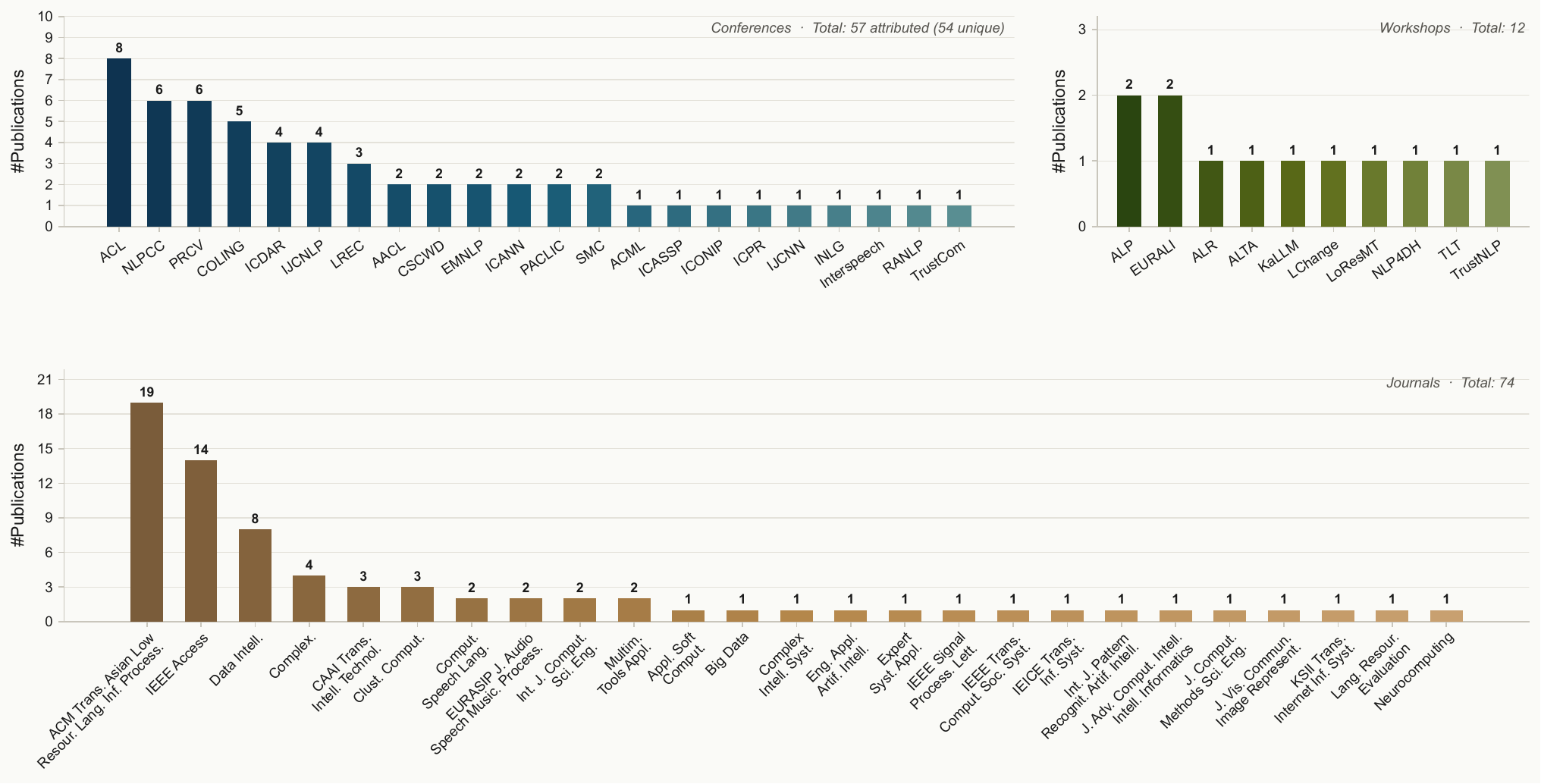}
	\captionsetup{justification=justified, singlelinecheck=false}
	\caption{Number of publications per conference (top left), workshop (top right), and journal (bottom). Papers at joint conferences (ACL-IJCNLP 2015, IJCNLP-AACL 2025) are counted under each constituent venue, so the conference counts sum to 57 for 54 papers. The full-scale version is Figure~\ref{fig_venues_appendix} in Appendix~\ref{appendix_figure}.}
	\label{fig_venues}
\end{figure}

Journal articles form the largest share of the corpus (74; 52.9\%), followed by conference papers (54; 38.6\%) and workshop papers (12; 8.6\%), dispersed across 10 workshops, 22 conferences, and 25 journals, led by TALLIP (19 articles) and \textit{IEEE Access} (14 articles) (Figure~\ref{fig_venues}). Only 37 publications (26.4\%) appear in the ACL Anthology, but 17 of these date from 2024--2026, including papers at ACL, EMNLP, AACL, COLING, and LREC: Tibetan NLP is entering the core NLP venues, although most of its literature circulates elsewhere.

\subsection{Research Areas}
\label{sec_areas}

\begin{figure}[h]
	\centering
	\includegraphics[width=\linewidth]{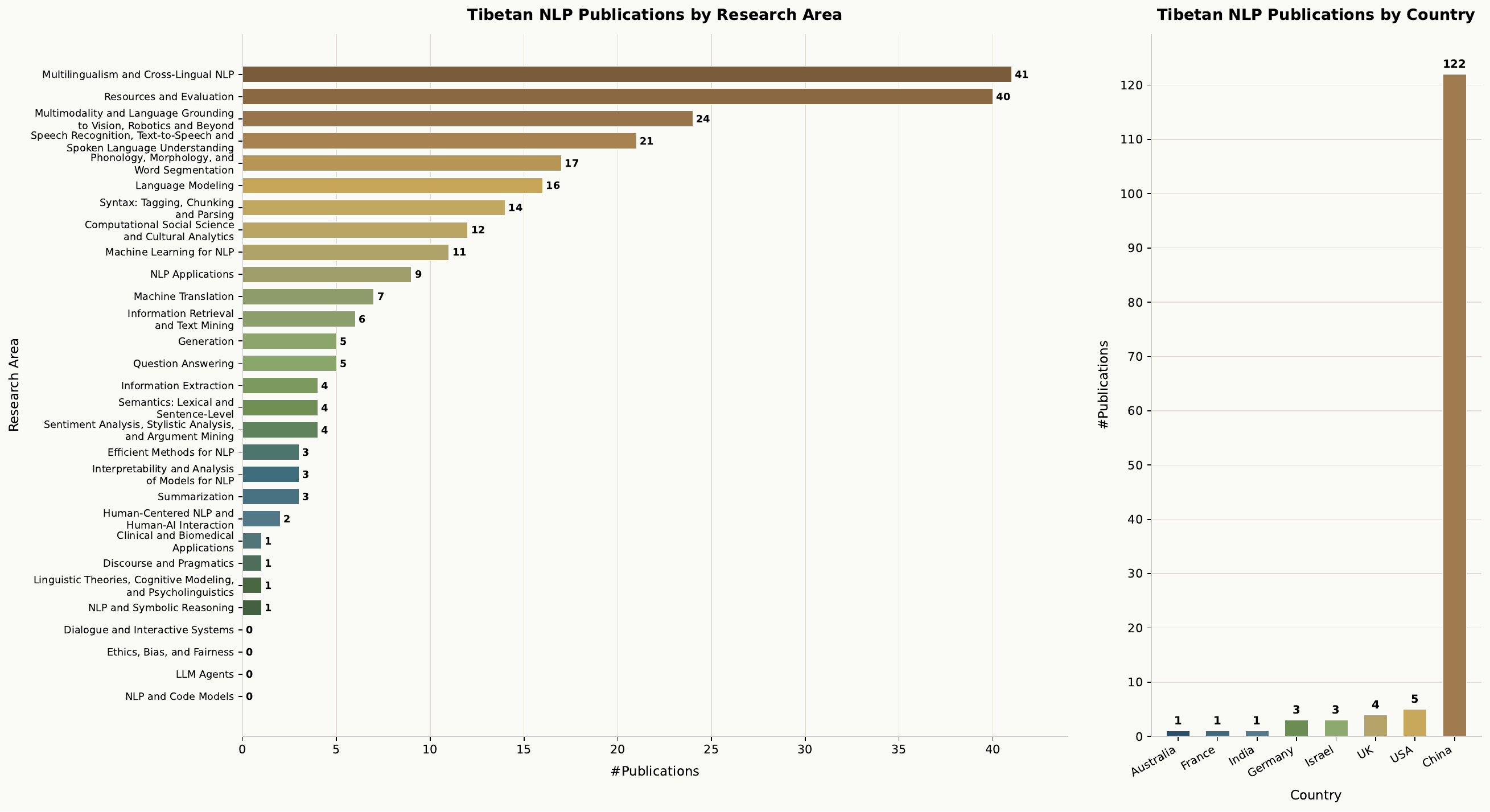}
	\captionsetup{justification=justified, singlelinecheck=false}
	\caption{Number of publications per ARR research area (left; descending order) and per country of the first author's affiliation (right; ascending order). As a publication can belong to more than one area, area counts sum to 255. The full-scale version is Figure~\ref{fig_area_country_appendix} in Appendix~\ref{appendix_figure}.}
	\label{fig_area_country}
\end{figure}

\begin{figure}[h]
	\centering
	\includegraphics[width=\linewidth]{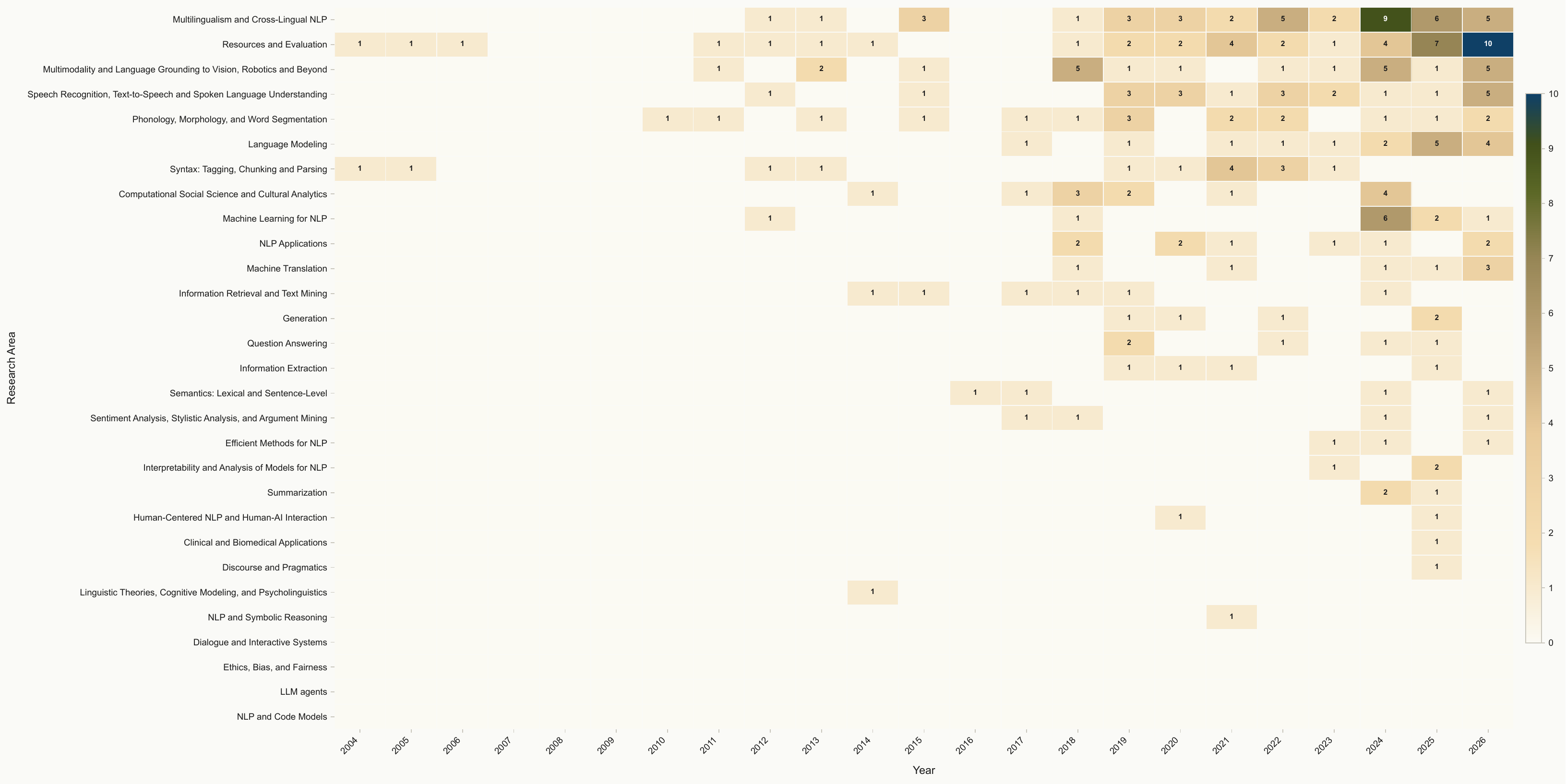}
	\captionsetup{justification=justified, singlelinecheck=false}
	\caption{Number of publications per ARR research area and year, 2004--2026; darker cells indicate more publications. Areas are ordered by total count (descending). The full-scale version is Figure~\ref{fig_heatmap_appendix} in Appendix~\ref{appendix_figure}.}
	\label{fig_heatmap}
\end{figure}

The 255 area assignments (Figures~\ref{fig_area_country}, left, and~\ref{fig_heatmap}) cover 25 of the 29 ARR areas but are highly skewed: Multilingualism and Cross-Lingual NLP (41 publications) and Resources and Evaluation (40) account for 31.8\% of all assignments and the five largest areas for 56.1\%, whereas 11 areas have between one and four publications and four (Dialogue and Interactive Systems; Ethics, Bias, and Fairness; LLM Agents; and NLP and Code Models) have none. We discuss them in four groups corresponding broadly to layers of the NLP pipeline.

\paragraph{Foundations: resources, segmentation, and syntax.} Resources and Evaluation is the most persistent area, active in 16 of the 20 years with any Tibetan NLP publication, from a syntactically annotated corpus and lexicographic resources \citep{wagner-zeisler-2004-syntactically,jiang-etal-2005-verbal,jiang-2006-current} to the training and evaluation data of the LLM era \citep{zhang-etal-2024-mc2,zhang-etal-2025-milic,gao-etal-2025-tlue,zhuang_tifd_2025,zhuang-sun-2025-cute,li-etal-2026-diversity}. Corpus scale has grown from 35M syllables of news text \citep{liu-etal-2012-building} to the 166M-token diachronic PACTib \citep{meelen-roux-2020-meta} and 72\,GB for LLM pre-training \citep{yang-etal-2026-curated}, and 17 of its 40 publications appeared in 2025--2026. Phonology, Morphology, and Word Segmentation (17) centres on the absence of word boundaries. For lack of a public segmented corpus, early work trained a conditional random field segmenter on machine-segmented text \citep{liu-etal-2011-tibetan}, and Chinese national standards for Tibetan word segmentation and part-of-speech tags followed only in 2018 \citep{zhu_new_2019,li_character-based_2022}. The area has since moved from syllable tagging \citep{liu-etal-2011-tibetan,li_study_2013} and unknown-word identification \citep{nuo-etal-2015-tibetan} to morphology-aware language models \citep{khysru_morphological_2019}, subword methods for MT \citep{li_finding_2021,gyatso_improved_2025}, joint segmentation and part-of-speech tagging \citep{li_character-based_2022}, neural segmentation \citep{brown-2024-improved}, and segmentation evaluation \citep{zhuoma_multi-granularity_2026}. Syntax: Tagging, Chunking and Parsing (14) follows the opposite trajectory: anchored in treebanks of Old and Classical Tibetan \citep{faggionato-meelen-2019-developing,meelen-roux-2020-meta} and in dependency parsing \citep{an_neural_2021,an-2023-tibetan}, it has no publications after 2023, mirroring its estimated decline across the ACL Anthology from 20\% to 2\% of papers in 2000--2021 \citep{rohatgi-etal-2023-acl}.

\paragraph{Script and speech.} Nearly a third of the corpus (45 publications; 32.1\%) processes visual or acoustic input. Multimodality and Language Grounding (24) is dominated by script recognition: online and offline handwriting \citep{ma_mrg-ohtc_2011,ma_online_2015,cai_online_2018,wang_online_2019,li_tibetan_2024,tashi_construction_2024,zhou_tibhcr_2026}, historical documents in woodblock and manuscript styles \citep{wang_research_2018,zhao_touching_2018,wang_recognition_2018,li_character_2022,zhao_incremental_2024}, scene text \citep{ke_align_2025,lhundrub_tibstr-former_2026}, and, recently, vision--language and multimodal optical character recognition (OCR) resources \citep{xu-etal-2026-ftibsuite,jie_tibocr-bench_2026}. Speech Recognition, Text-to-Speech and Spoken Language Understanding (21) has progressed from isolated-word recognition with at most 20 speakers \citep{zhao_unsupervised_2012} to end-to-end recognition for the Lhasa and Amdo varieties and in multi-dialect settings \citep{zhao_end--end-based_2019,qin_finer-grained_2022,liu_multi-task_2023,zhu_end--end_2020,liu_exploring_2024}, alongside speech synthesis \citep{yang_using_2015,zhao_lhasa-tibetan_2019,xu_end--end_2021,zhang_cro-mtvits_2026}, speech translation \citep{fan_tibetan-chinese_2026,ren_research_2026}, and word-level evaluation \citep{moore_evaluating_2026}. A recurring strategy exploits the compositional structure of the script: decomposing stacks into components reduces 562 handwritten character classes to 120 components \citep{ma_semi-automatic_2013,ma_online_2015} and aids few-shot printed-text recognition \citep{han_research_2018}, and sub-syllable radicals serve as units for language modelling and speech recognition \citep{shen_exploiting_2017,qin_finer-grained_2022,qin_improving_2022}. Both areas' prominence plausibly reflects a large manuscript and print heritage and spoken varieties poorly served by written resources.

\paragraph{Cross-lingual transfer and language modelling.} Multilingualism and Cross-Lingual NLP (41) is the largest area and has gained the most publications since 2022 (27). Early work uses the Tibetan--Chinese pair to project base noun phrases without a Tibetan tagger or treebank \citep{nuo-etal-2012-tibetan}, to gloss keywords in OCR output in Chinese in lieu of unreliable MT \citep{su_cross-language_2013}, and for bilingual clustering and cross-lingual speech synthesis \citep{ye_chinese-tibetan_2015,yang_using_2015}, while Sanskrit--Tibetan resources serve the classical tradition \citep{nehrdich-2022-sanstib}. Since 2024, the area has been shaped by multilingual corpora and benchmarks for China's minority languages \citep{zhang-etal-2024-mc2,zhang-etal-2025-milic,li-etal-2026-diversity} and by adapting multilingual LLMs \citep{gurgurov-etal-2024-adapting,yang-etal-2026-curated}. Language Modeling (16; 13 since 2022) recapitulates the wider field's arc, from recurrent models exploiting radicals and morphology \citep{shen_exploiting_2017,khysru_morphological_2019}, through BERT-style models \citep{liu_tibert_2022,liang_tibetan-bert-wwm_2024,zhang_tibetan_2025}, to LLaMA-based LLMs \citep{lv_t-llama_2025,sha_tibetan-llama_2025}, alignment datasets \citep{zhuang_tifd_2025,chen_dataset_2026}, and studies of vocabulary size and continual pre-training \citep{han-etal-2026-scaling,yang-etal-2026-curated}. Machine Learning for NLP (11; nine since 2024) and Efficient Methods for NLP (3; all since 2023) reflect a turn to few-shot, prompt-based, and parameter-efficient adaptation \citep{bao_toward_2024,tang_tibetan_2025,zhou_efficient_2024}.

\paragraph{Applications.} Computational Social Science and Cultural Analytics (12) and Information Retrieval and Text Mining (6) emerged with mining Tibetan web and microblog text in 2014--2019 \citep{ye_tm-tot_2014,qiu_tibetan_2017,xu_efficient_2018}, and the former has returned with offensive-language and rumour detection \citep{ren_method_2024,yu_novel_2024}. Eight of the nine publications in NLP Applications concern historical or classical materials, such as interlinearized ancient texts \citep{long-an-2023-development} and OCR spelling correction \citep{luo_cleansing_2024}. Machine Translation (7), has moved from syllable- and subword-based neural MT \citep{lai-etal-2018-tibetan,li_finding_2021} to filtering noisy parallel data with a cross-lingual model \citep{zhou_tibetan-chinese_2024}, speech translation from Amdo or all three dialect groups \citep{ren_research_2026,fan_tibetan-chinese_2026}, and comparisons of dedicated MT engines with a prompted LLM \citep{wang_quality_2026}. Question Answering (5), Generation (5), Information Extraction (4), and Summarization (3) remain small \citep[e.g.,][]{dan_tibetanqa20_2024,xu-etal-2025-cmhg,zhang_tibetan_2025-1}; all summarization studies use news titles, leads, or LLM-generated summaries as references \citep{ouyang_pre-trained_2024,hao_tiltstibetan_2024,wang_disg_2025}.

\paragraph{Persistent gaps.} Besides the four empty areas, topics central to the responsible deployment of language technology are nearly absent. Interpretability and Analysis of Models (3) is represented only by adversarial-robustness studies from a single research team \citep{cao-etal-2023-pay-attention,cao-etal-2025-human,cao_tscheater_2025}, Human-Centered NLP (2) by a readability tool and human-in-the-loop adversarial-text generation \citep{schmidt_grading_2020,cao-etal-2025-human}, and Clinical and Biomedical Applications (1) by medical named-entity recognition, where F1 falls to 71\% from 94\% on news \citep{zhang_tibetan_2025-1}; Discourse and Pragmatics, Linguistic Theories, and Symbolic Reasoning have one each.

\paragraph{A layered trajectory.} Figure~\ref{fig_heatmap} and Table~\ref{tab_phases} show that the field has expanded broadly bottom-up along the NLP pipeline, adding new layers rather than replacing old ones: resources and syntax appeared in 2004; segmentation, script, speech, and cross-lingual work in 2010--2012; web mining in 2014; language modelling, machine translation, question answering, generation, and information extraction in 2017--2019; and efficient methods, interpretability, summarization, clinical applications, and discourse only in 2023--2025. Earlier layers persisted: resources remain the second most frequent area in the LLM era, and segmentation research continues. Only syntax has declined, as across the ACL Anthology, where resources have also waned but dialogue, interpretability, and ethics have grown \citep{rohatgi-etal-2023-acl}, the very areas in which Tibetan NLP lags.

\begin{table}[t]
	\centering
	\small
	\begin{tabular*}{\linewidth}{@{\extracolsep{\fill}}l c c r@{}}
		\toprule
		\textbf{Phase} & \textbf{Publications} & \textbf{Areas} & \textbf{China} \\
		\midrule
		2004--2017 & 22 & 13 & 95.5\% \\
		\multicolumn{4}{@{}>{\raggedright\arraybackslash}p{\linewidth}@{}}{\textit{Top areas:} Resources (7), Phon./Morph./Seg. (5), Multilingual (5), Syntax (4), Multimodality (4)} \\
		\midrule
		2018--2021 & 40 & 18 & 82.5\% \\
		\multicolumn{4}{@{}>{\raggedright\arraybackslash}p{\linewidth}@{}}{\textit{Top areas:} Resources (9), Multilingual (9), Multimodality (7), Speech (7)} \\
		\midrule
		2022--2026 & 78 & 23 & 87.2\% \\
		\multicolumn{4}{@{}>{\raggedright\arraybackslash}p{\linewidth}@{}}{\textit{Top areas:} Multilingual (27), Resources (24), Language Modeling (13), Multimodality (13)} \\
		\bottomrule
	\end{tabular*}
	\caption{Three phases of Tibetan NLP. \textbf{Areas}: ARR areas represented; \textbf{China}: share first-authored in China; \textbf{Top areas}: four most frequent areas, with ties. Boundaries: the 2018 output rise and the first BERT-style model, TiBERT \citep{liu_tibert_2022}, in 2022.}
	\label{tab_phases}
\end{table}

\subsection{Country-Level Analysis}
\label{sec_country}

Research is highly concentrated (Figure~\ref{fig_area_country}, right): 122 publications (87.1\%) are first-authored in China (hereafter, China-affiliated publications), and the remaining 18 in the USA (5), the UK (4), Germany (3), Israel (3), and Australia, France, and India (one each). China's share has exceeded 80\% in every phase (Table~\ref{tab_phases}), consistent with its lead in research teams and public funding for Tibetan information processing \citep{congjun_recent_2021}.

China's contribution is broad as well as large: China-affiliated publications cover all 25 represented areas, 15 of them exclusively, including Language Modeling, Multimodality, Machine Translation, Question Answering, and Generation. The 18 publications from other countries fall into ten areas, mainly Resources and Evaluation (9), Syntax (5), NLP Applications (5), and Multilingualism (5), and most concern Old or Classical Tibetan and Buddhist texts \citep[cf.][]{congjun_recent_2021}: UK-based researchers built treebanks and annotated corpora \citep{faggionato-meelen-2019-developing,meelen-roux-2020-meta,meelen_optimisation_2021}; Israel-based researchers, with German Tibetologists, developed evaluation resources and methods for Buddhist texts \citep{hashiloni-etal-2025-dharmabench,cohen-etal-2026-scaling,bilitski-etal-2026-automatic}; and researchers in Germany, France, and the USA contributed a Sanskrit--Tibetan parallel corpus, e-text tools, text analytics, and OCR spelling correction \citep{nehrdich-2022-sanstib,trinley_taming_2021,krishna_applying_2021,luo_cleansing_2024}. US-based work also covers modern Tibetan segmentation, speech recognition evaluation, and tonogenesis \citep{brown-2024-improved,moore_evaluating_2026,liang-zerong-2026-tonogenesis}.

The two groups also publish in different venues: two-thirds of the publications from outside China (12 of 18) appear in the ACL Anthology, seven of them at workshops, whereas 97 of the 122 China-affiliated publications (79.5\%) appear in journals or in conferences outside it. Tibetan NLP thus comprises two partly separate communities that differ in research focus as well as in venue, although exchange is productive where it occurs, as in the 2021 TALLIP special issue with papers from China, France, and the USA \citep{congjun_recent_2021}.

\section{Discussion}
\label{sec_discussion}

\subsection{A Full-Pipeline Paradigm and Its Limits}
\label{sec_paradigm}

Our results characterize Tibetan NLP by \textit{concentration} (87.1\% first-authored in China), \textit{breadth} (China-affiliated work covers all 25 represented areas), and \textit{layering} (new layers added, earlier ones kept). We call this pattern a full-pipeline paradigm: research spans every layer from script processing to LLMs, though not every task in each layer.
The paradigm also has limits. First, it is predominantly engineering-oriented, with little fundamental research \citep[cf.][]{congjun_recent_2021}. Second, it is only partly visible: most China-affiliated research lies outside the ACL Anthology, and hence outside surveys restricted to it \citep{liu-best-2025-survey}. Third, its outputs are only partly reusable: 41 of the 140 publications (29.3\%) release data, code, or models at a public URL (Appendix~\ref{appendix_open_source}). Release has become more common, from none of the 31 publications of 2004--2018 to 27 of the 61 of 2024--2026 (44.3\%), but remains less frequent.

\subsection{Lessons for the Global South}
\label{sec_lessons}

Many low-resource languages of the Global South face obstacles usually studied in isolation: dispersed speakers, dialectal diversity, scripts that complicate encoding and segmentation, and limited infrastructure and investment. All co-occur on the Qinghai-Xizang Plateau, yet Tibetan research has advanced across the entire pipeline, suggesting four transferable lessons: (1)~invest first in the base layers of script processing, segmentation, and annotated corpora, on which later work depends; (2)~treat resource and benchmark construction as a continuing frontier, not a one-off step (\S\ref{sec_areas}); (3)~exploit a high-resource contact language, here Mandarin Chinese, for cross-lingual transfer, while guarding against translated data and benchmarks, which can lose cultural content and misstate difficulty \citep{zhuang-sun-2025-cute,li-etal-2026-diversity}; and (4)~use dedicated venues to consolidate a research community (\S\ref{sec_trends}). Transfer is conditional: Tibetan NLP has had sustained public funding (\S\ref{sec_country}), whereas African NLP has grown through community initiatives and targeted funding \citep{alabi-etal-2025-charting}; communities with neither will rely more on international collaboration and open resources.

\subsection{The Transition to LLMs}
\label{sec_llm}

Since 2022, the field has pivoted toward pre-trained models and LLMs, from three separate LLaMA-2 extensions with Tibetan vocabularies and 6--26\,GB of text \citep{lv_t-llama_2025,sha_tibetan-llama_2025,zhuang_tifd_2025} to dense and mixture-of-experts models trained on a curated 72\,GB corpus \citep{yang-etal-2026-curated}, vocabulary sizing \citep{han-etal-2026-scaling}, reinforcement learning that limits forgetting of Chinese \citep{su-etal-2026-reinforcement}, and vision--language models \citep{xu-etal-2026-ftibsuite}. Two obstacles recur. Tokenization is unresolved: a syllable vocabulary of 35K types still leaves 15\% of tokens unknown \citep{liu_tibert_2022}, whereas multilingual tokenizers fragment Tibetan, so that a 27-token English sentence takes 100 or more in Tibetan \citep{zhang-etal-2025-milic}, and expanding byte-level byte-pair-encoding vocabularies helps only beyond about 3,000 new tokens per language \citep{han-etal-2026-scaling}. Data are noisy and partly closed: 42\% of sampled Tibetan CulturaX pages retain headers or footers \citep{zhang-etal-2024-mc2}, CINO's pre-training corpus is not public \citep{zhuang-sun-2025-cute}, and 41\,GB of the 72\,GB corpus are private \citep{yang-etal-2026-curated}.

Evaluation reveals a large gap. GPT-3.5 reaches 3.6\% F1 on Tibetan WikiAnn named-entity recognition, the lowest of ten low-resource Sino-Tibetan languages \citep{liu-best-2025-survey}; on TLUE, most LLMs perform below a random baseline, and Qwen2.5-72B drops from 84.7\% on the Chinese CMMLU to 16.5\% on its Tibetan counterpart \citep{gao-etal-2025-tlue}; and general multimodal LLMs recognize 49--66\% of characters on TibOCR-Bench, against 92\% for a 7B model tuned on 4.5K Tibetan images \citep{jie_tibocr-bench_2026}. Evaluation is itself a bottleneck. Encoders are compared chiefly on one news-classification corpus, TNCC \citep{liu_tibert_2022,zhou_efficient_2024,hao_tibetan_2024,cao_tscheater_2025}, and task data are increasingly obtained by pivoting through other languages, as in offensive-language and sentiment datasets translated from English or Chinese \citep{ren_method_2024,zhang_tibetan_2026} and summaries bootstrapped through Chinese with GPT-3.5 \citep{hao_tiltstibetan_2024}. Alignment data, too, are wholly or partly translated, from GPT-4 answers to English prompts \citep{zhuang_tifd_2025} or Chinese preference data \citep{chen_dataset_2026}. Many LLM benchmarks are likewise translated \citep{gao-etal-2025-tlue,xu-etal-2026-ftibsuite,yang-etal-2026-curated}, yet machine-translated test items depress scores, most of all for Tibetan \citep{zhang-etal-2025-milic}, and tasks built from native minority-language knowledge are 11.4--13.9 points harder than translated ones \citep{li-etal-2026-diversity}. The move to LLMs thus risks outpacing its foundations, with dialects underserved and evaluation standards only emerging (\S\ref{sec_recommendations}).

\subsection{Classical and Modern Tibetan}
\label{sec_classical}

The corpus contains two research traditions, centred on modern Tibetan and its spoken varieties and on Old and Classical Tibetan, which differ in who conducts them and where they appear (\S\ref{sec_country}). China-affiliated work on historical materials has focused on document-image analysis \citep{wang_recognition_2018,ma_segmentation_2020,li_character_2022}, the frontier of character recognition once OCR of modern print was in effect solved \citep{congjun_recent_2021}, and on interlinearization \citep{long-an-2023-development}. Shared archives already link the traditions: Buddhist Digital Resource Center scans underlie both a China-built character dataset \citep{li_character_2022} and US-based OCR post-correction \citep{luo_cleansing_2024}. Transfer, however, works more readily within writing than for speech: Old Tibetan has been normalized to Classical orthography to reuse its tools \citep{faggionato-meelen-2019-developing}, and a single pipeline annotates texts from the 11th to the 21st century because literary Tibetan has changed little \citep{meelen-roux-2020-meta}, whereas pre-training on Classical text did not help segment spoken Standard Tibetan \citep{brown-2024-improved}, and a segmenter with a classical and literary dictionary marks 87\% of syllable boundaries in transcribed Lhasa speech as word boundaries, against 41\% for an annotator \citep{moore_evaluating_2026}. Domain also matters within writing: general-domain Tibetan--Chinese MT drops from about 50 BLEU to at most 5 on Buddhist text \citep{gyatso_improved_2025}. Classical treebanks and parallel corpora could support modern written Tibetan, and modern OCR and parsing the digitization of classical texts, while spoken varieties need their own resources.

\section{Recommendations}
\label{sec_recommendations}

\paragraph{Open science and cross-community exchange.} As fewer than a third of surveyed publications release resources (\S\ref{sec_paradigm}), we encourage public release, not only on request, of datasets, models, and code under documented licences, and of pre-training corpora where rights permit \citep[as in][]{trinley_taming_2021,zhang-etal-2024-mc2}, following the practices in Appendix~\ref{appendix_open_source}. To make work findable, authors should name the varieties studied in titles and abstracts, and venues could record them as metadata \citep{liu-best-2025-survey}; joint workshops, special issues, and shared tasks could connect China-affiliated research with research elsewhere, and the classical with the modern tradition (\S\ref{sec_classical}).

\paragraph{Shared standards and native evaluation.} To make results comparable across teams, the field needs shared conventions for encoding normalization, segmentation, and unit terminology; public test sets for core tasks; and reporting of splits, metrics, and the varieties and scripts covered \citep{kumar-etal-2026-bhashasutra}. Shared segmentation tasks offer a basis, but their single gold segmentation covers only 22\% of valid granularities \citep{zhuoma_multi-granularity_2026}. LLM benchmarks should increasingly be built from native material with expert involvement, as in CMiLBench's minority-knowledge tasks and DharmaBench \citep{li-etal-2026-diversity,hashiloni-etal-2025-dharmabench}, rather than by translation alone.

\paragraph{Community participation.} No surveyed publication assesses Tibetan speakers' needs, yet needs assessments show that priorities vary across speaker groups, even across dialects of one language \citep{al-kautsar-etal-2025-indonesians}; German dialect speakers, for instance, favour tools that take spoken dialect input over those producing dialect output \citep{blaschke-etal-2024-dialect}, a preference worth testing in diglossic Tibetan. Resource creation and evaluation should involve speakers of all three major dialect groups, including diaspora communities, throughout research \citep{tonja-etal-2024-nlp}, as handwriting datasets with up to 235 writers from five regions \citep{tashi_construction_2024,zhou_tibhcr_2026} and a readability tool built on children's speech \citep{schmidt_grading_2020} begin to do.

\paragraph{Responsible NLP.} Beyond the gaps noted in \S\ref{sec_areas}, safety is evaluated only within recent benchmarks, which show that it does not carry over to Tibetan: most LLMs fall below chance on TLUE's safety set \citep{gao-etal-2025-tlue}, and safety alignment transfers unreliably across minority languages \citep{li-etal-2026-diversity}. Bias is unmeasured, although zero-shot LLMs tend to label Tibetan news as politics or religion \citep{tang_tibetan_2025}. Applications that mine user-level social-media data for public-opinion monitoring or to locate user groups \citep{ye_tm-tot_2014,qiu_tibetan_2018,xu_tibetan_2024}, microblog corpora stored with user names \citep{qiu_tibetan_2017,qiu_advanced_2018}, speaker verification motivated partly by criminal investigation \citep{gan_tibetan-dependent_2022}, and an OCR benchmark with identity cards and bank receipts but no stated anonymization \citep{jie_tibocr-bench_2026} raise privacy and dual-use questions that the absence of ethics research leaves unexamined and that call for the consent and privacy safeguards urged for African languages \citep{adebara-abdul-mageed-2022-towards}. Bias, safety, privacy, and human-centred evaluation need study before Tibetan LLMs are deployed in education, healthcare, or public services.

\paragraph{Dialects and related languages.} Only eight publications name Lhasa, Amdo, or a multi-dialect setting in their titles, all on speech recognition, synthesis, or translation. Open speech data for all three groups total about 80--100 hours, against over 10,000 for Mandarin or English \citep{liu_exploring_2024}, and Kham is their smallest, hardest part: the main multi-dialect corpus, TIBMD@MUC, was released with 8.7 hours of Kham speech against 30 of Lhasa speech \citep{zhao_open_2020}, and even the best recognizer trained on it errs on 23\% of Kham characters, against 3--5\% for the other groups \citep{liu_exploring_2024}. Kham and Amdo resources should be prioritized, publicly released, and evaluated per variety. Dzongkha and Ladakhi use Tibetan script, so OCR and segmentation tools may transfer directly, and Burmese could share low-level resources \citep{fang_mlcl_2024}.

\section{Conclusion}
\label{sec_conclusion}

We have surveyed 140 Tibetan NLP publications from 2004 to 2026. Since 2018, the field has grown rapidly and expanded bottom-up along the NLP pipeline while continuing to build resources, with 87\% of the work first-authored in China. This full-pipeline paradigm can serve as a reference model for other low-resource languages of the Global South; its limits in dialect coverage, responsible NLP, native and comparable evaluation, resource release, and cross-community visibility set the agenda for keeping plateau-region languages from being left behind in the LLM era.

\section*{Limitations}

This survey has several limitations. First, its coverage is bounded by its sources: we include only publications indexed in the ACL Anthology or DBLP, so work in Chinese-language venues, in venues indexed by neither source, and in preprints is not covered, and topics published mainly in such venues may be underrepresented. Second, keyword-based retrieval may miss multilingual studies that include Tibetan without naming it in the searched metadata fields; this risk is greater for DBLP, whose records do not include abstracts. Third, assigning research areas required interpretive judgement, especially for work spanning several topics. Fourth, the country-level analysis relies on the first author's primary affiliation, which may not reflect all contributors or the national context of the research. Fifth, publication counts measure research output rather than impact or quality, and we did not perform a citation analysis; our audit of released resources (Appendix~\ref{appendix_open_source}) records whether resources are publicly available at a URL given by their authors, not whether they are complete, documented, or maintained, and may miss resources released elsewhere or whose links have lapsed. Sixth, our qualitative synthesis, including the discussion of evaluation practice and of the LLM era, relies on results reported by the surveyed publications, which use different models, data, and metrics and which we did not reproduce. Finally, the survey reflects the literature available up to \cutoffdate, so 2026 is only partially covered. We release the full bibliography, in raw and refined versions, to facilitate community-driven correction and extension, and we will continue to track new work: \url{https://github.com/metaphors/TibetanNLPSurvey}.

\section*{Ethical Considerations}

This survey analyses publicly available bibliographic metadata and peer-reviewed publications; it involves no human subjects and no personal data beyond author names and affiliations as they appear in the public record. Country-level statistics describe institutional affiliations, not the nationality or ethnicity of researchers, and differences in resource release rates may reflect venue norms and legitimate constraints, such as consent and privacy requirements for speech data or missing permission to share collected data, rather than the practices of individual teams. Where we discuss the data practices or stated applications of individual publications, we do so to motivate shared norms, not to single out particular teams. We do not reproduce or redistribute copyrighted material beyond standard academic citation practices. We have aimed to describe research trends accurately without advancing normative political positions, and we recognize that decisions about language technology for Tibetan should involve the communities who speak it.

\section*{Acknowledgments}

This work is supported by the Science and Technology Strategic Consulting Project of the Chinese Academy of Engineering (2025-XZ-16-06), the Project of the China Tibetology Literature Resources Data Center (2025SJ003), the National Social Science Foundation of China (22\&ZD035), the Key Project of the Natural Science Foundation of Xizang Autonomous Region (XZ202401ZR0040), and the Major Project of the Science and Technology Program of Xizang Autonomous Region (XZ202601ZD0010).

\bibliography{custom_acl,custom_dblp_conference,custom_dblp_journal,custom_others,custom_balti,custom_ladakhi,custom_dzongkha,custom_bhutia,custom_tamang}

@inproceedings{li-etal-2026-diversity,
	title = "Diversity in Unity, Theory in Practice: Hierarchical Multitask Benchmarks for {C}hinese Minority Languages",
	author = "Li, Yijie  and
	Cao, Xi  and
	Sun, Yuan  and
	Minggad, Quulgan  and
	Ablikim, Abdulla  and
	Wang, Jia Qing Cai",
	editor = "Liakata, Maria  and
	Moreira, Viviane P.  and
	Zhang, Jiajun  and
	Jurgens, David",
	booktitle = "Proceedings of the 64th Annual Meeting of the {A}ssociation for {C}omputational {L}inguistics (Volume 1: Long Papers)",
	month = jul,
	year = "2026",
	address = "San Diego, California, United States",
	publisher = "Association for Computational Linguistics",
	url = "https://aclanthology.org/2026.acl-long.1684/",
	doi = "10.18653/v1/2026.acl-long.1684",
	pages = "36353--36373",
	ISBN = "979-8-89176-390-6"
}

@inproceedings{yang-etal-2026-curated,
	title = "From Curated Data to Scalable Models: Continual Pre-training of Dense and {M}o{E} Large Language Models for {T}ibetan",
	author = "Yang, Lei  and
	Pan, Leiyu  and
	Xiong, Bojian  and
	Jin, Renren  and
	Zhang, Shaowei  and
	Chen, Yue  and
	Shi, Ling  and
	Zhou, Jiang  and
	Wu, Junru  and
	Wang, Zhen  and
	Peng, Jianxiang  and
	Xiao, Juesi  and
	Dong, Tianyu  and
	Han, Zhuowen  and
	Chen, Zhuo  and
	Ren, Yuqi  and
	Xiong, Deyi",
	editor = "Liakata, Maria  and
	Moreira, Viviane P.  and
	Zhang, Jiajun  and
	Jurgens, David",
	booktitle = "Proceedings of the 64th Annual Meeting of the {A}ssociation for {C}omputational {L}inguistics (Volume 1: Long Papers)",
	month = jul,
	year = "2026",
	address = "San Diego, California, United States",
	publisher = "Association for Computational Linguistics",
	url = "https://aclanthology.org/2026.acl-long.1866/",
	doi = "10.18653/v1/2026.acl-long.1866",
	pages = "40167--40182",
	ISBN = "979-8-89176-390-6"
}

@inproceedings{su-etal-2026-reinforcement,
	title = "Reinforcement Learning with Semantic Rewards Enables Low-Resource Language Expansion without Alignment Tax",
	author = "Su, Zeli  and
	Zhang, Ziyin  and
	Liu, Zhou  and
	Song, Xuexian  and
	Xu, Zhankai  and
	Zheng, Longfei  and
	Zhang, Xiaolu  and
	Fu, Rong  and
	Xu, Guixian  and
	Zhang, Wentao",
	editor = "Liakata, Maria  and
	Moreira, Viviane P.  and
	Zhang, Jiajun  and
	Jurgens, David",
	booktitle = "Findings of the {A}ssociation for {C}omputational {L}inguistics: {ACL} 2026",
	month = jul,
	year = "2026",
	address = "San Diego, California, United States",
	publisher = "Association for Computational Linguistics",
	url = "https://aclanthology.org/2026.findings-acl.880/",
	doi = "10.18653/v1/2026.findings-acl.880",
	pages = "17772--17786",
	ISBN = "979-8-89176-395-1"
}

@inproceedings{xu-etal-2026-ftibsuite,
	title = "{FT}ib{S}uite: A Comprehensive Resource Suite for {T}ibetan Vision{--}Language Modeling",
	author = "Xu, Guixian  and
	Liang, Yide  and
	Su, Zeli  and
	Song, Xuexian  and
	Zhang, Ziyin  and
	Dong, Yushuang  and
	Zhang, Ting  and
	Han, Xu",
	editor = "Liakata, Maria  and
	Moreira, Viviane P.  and
	Zhang, Jiajun  and
	Jurgens, David",
	booktitle = "Findings of the {A}ssociation for {C}omputational {L}inguistics: {ACL} 2026",
	month = jul,
	year = "2026",
	address = "San Diego, California, United States",
	publisher = "Association for Computational Linguistics",
	url = "https://aclanthology.org/2026.findings-acl.903/",
	doi = "10.18653/v1/2026.findings-acl.903",
	pages = "18143--18159",
	ISBN = "979-8-89176-395-1"
}

@inproceedings{han-etal-2026-scaling,
	title = "Scaling Laws or Threshold Effects: Exploring the Optimal Vocabulary Size for Balancing Performance and Efficiency in Low-Resource Languages",
	author = "Han, Ao  and
	Chen, Andong  and
	Sun, Yuan  and
	Zhao, Xiaobing",
	editor = "Liakata, Maria  and
	Moreira, Viviane P.  and
	Zhang, Jiajun  and
	Jurgens, David",
	booktitle = "Findings of the {A}ssociation for {C}omputational {L}inguistics: {ACL} 2026",
	month = jul,
	year = "2026",
	address = "San Diego, California, United States",
	publisher = "Association for Computational Linguistics",
	url = "https://aclanthology.org/2026.findings-acl.1588/",
	doi = "10.18653/v1/2026.findings-acl.1588",
	pages = "31741--31758",
	ISBN = "979-8-89176-395-1"
}

@inproceedings{cohen-etal-2026-scaling,
	title = "Scaling Sentence Similarity for {C}lassical {T}ibetan with Automatic Annotations",
	author = "Cohen, Shay  and
	Yang, Jingyi  and
	Rabinovitz, Gal  and
	Choden, Sonam  and
	Shtrosberg, Ofir  and
	Bajetta, Nicola  and
	Horin, Goody Ben  and
	Sund{\'e}n, Rebecca  and
	Drori, Omri  and
	Jamtsho, Sonam  and
	Wangchuk, Dorji  and
	Bar, Kfir  and
	Almogi, Orna  and
	Fine, Shai",
	editor = {Hamilton, Sil  and
	{\"O}hman, Emily  and
	Hicke, Rebecca M. M.  and
	Bizzoni, Yuri  and
	Bax, Axel  and
	Matthews, Jacob A.  and
	H{\"a}m{\"a}l{\"a}inen, Mika},
	booktitle = "Proceedings of the 6th International Conference on Natural Language Processing for the Digital Humanities",
	month = jul,
	year = "2026",
	address = "San Diego, USA",
	publisher = "Association for Computational Linguistics",
	url = "https://aclanthology.org/2026.nlp4dh-1.15/",
	doi = "10.18653/v1/2026.nlp4dh-1.15",
	pages = "150--166",
	ISBN = "979-8-89176-427-9"
}

@inproceedings{bilitski-etal-2026-automatic,
	title = "Automatic Segmentation of {C}lassical {T}ibetan Texts into Autochthonous and Allochthonous Regions",
	author = "Bilitski, Guy  and
	Shechter, Lev  and
	Jamtsho, Sonam  and
	Marciano, Nir  and
	Bajetta, Nicola  and
	Sunden, Rebecca  and
	Drori, Omri  and
	Hashiloni, Kai Golan  and
	Zwebner, Orr  and
	Shina, Asaf  and
	Almogi, Orna  and
	Wangchuk, Dorji  and
	Bar, Kfir",
	editor = "Piperidis, Stelios  and
	Bel, N{\'u}ria  and
	van den Heuvel, Henk  and
	Ide, Nancy  and
	Krek, Simon  and
	Toral, Antonio",
	booktitle = "Proceedings of the Fifteenth Language Resources and Evaluation Conference",
	month = may,
	year = "2026",
	address = "Palma de Mallorca, Spain",
	publisher = "ELRA Language Resource Association",
	url = "https://aclanthology.org/2026.lrec-1.79/",
	doi = "10.63317/2iyfjjv9boc6",
	pages = "1017--1030"
}

@inproceedings{liang-zerong-2026-tonogenesis,
	title = "The Tonogenesis Continuum in {T}ibetan: A Computational Investigation",
	author = "Liang, Siyu  and
	Zerong, Zhaxi",
	editor = "Tahmasebi, Nina  and
	Cassotti, Pierluigi  and
	Montariol, Syrielle  and
	Kutuzov, Andrey  and
	Huebscher, Netta  and
	Spaziani, Elena  and
	Baes, Naomi",
	booktitle = "The Proceedings for the 6th International Workshop on Computational Approaches to Language Change ({LC}hange{'}26)",
	month = mar,
	year = "2026",
	address = "Rabat, Morocco",
	publisher = "Association for Computational Linguistics",
	url = "https://aclanthology.org/2026.lchange-1.3/",
	doi = "10.18653/v1/2026.lchange-1.3",
	pages = "27--34",
	ISBN = "979-8-89176-362-3"
}

@inproceedings{hashiloni-etal-2025-dharmabench,
	title = "{D}harma{B}ench: Evaluating Language Models on {B}uddhist Texts in {S}anskrit and {T}ibetan",
	author = "Hashiloni, Kai Golan  and
	Cohen, Shay  and
	Shina, Asaf  and
	Yang, Jingyi  and
	Zwebner, Orr Meir  and
	Bajetta, Nicola  and
	Bilitski, Guy  and
	Sund{\'e}n, Rebecca  and
	Maduel, Guy  and
	Conlon, Ryan  and
	Barzilai, Ari  and
	Mass, Daniel  and
	Jia, Shanshan  and
	Naaman, Aviv  and
	Choden, Sonam  and
	Jamtsho, Sonam  and
	Qu, Yadi  and
	Isaacson, Harunaga  and
	Wangchuk, Dorji  and
	Fine, Shai  and
	Almogi, Orna  and
	Bar, Kfir",
	editor = "Inui, Kentaro  and
	Sakti, Sakriani  and
	Wang, Haofen  and
	Wong, Derek F.  and
	Bhattacharyya, Pushpak  and
	Banerjee, Biplab  and
	Ekbal, Asif  and
	Chakraborty, Tanmoy  and
	Singh, Dhirendra Pratap",
	booktitle = "Proceedings of the 14th International Joint Conference on Natural Language Processing and the 4th Conference of the Asia-Pacific Chapter of the Association for Computational Linguistics",
	month = dec,
	year = "2025",
	address = "Mumbai, India",
	publisher = "The Asian Federation of Natural Language Processing and The Association for Computational Linguistics",
	url = "https://aclanthology.org/2025.ijcnlp-long.114/",
	doi = "10.18653/v1/2025.ijcnlp-long.114",
	pages = "2088--2110",
	ISBN = "979-8-89176-298-5"
}

@inproceedings{cao-etal-2025-human,
	title = "Human-in-the-Loop Generation of Adversarial Texts: A Case Study on {T}ibetan Script",
	author = "Cao, Xi  and
	Sun, Yuan  and
	Li, Jiajun  and
	Gesang, Quzong  and
	Qun, Nuo  and
	Tashi, Nyima",
	editor = "Liu, Xuebo  and
	Purwarianti, Ayu",
	booktitle = "Proceedings of The 14th International Joint Conference on Natural Language Processing and The 4th Conference of the Asia-Pacific Chapter of the Association for Computational Linguistics: System Demonstrations",
	month = dec,
	year = "2025",
	address = "Mumbai, India",
	publisher = "Association for Computational Linguistics",
	url = "https://aclanthology.org/2025.ijcnlp-demo.2/",
	doi = "10.18653/v1/2025.ijcnlp-demo.2",
	pages = "9--16",
	ISBN = "979-8-89176-301-2"
}

@inproceedings{xu-etal-2025-cmhg,
	title = "{CMHG}: A Dataset and Benchmark for Headline Generation of Minority Languages in {C}hina",
	author = "Xu, Guixian  and
	Su, Zeli  and
	Zhang, Ziyin  and
	Liu, Jianing  and
	Han, Xu  and
	Zhang, Ting  and
	Dong, Yushuang",
	editor = "Christodoulopoulos, Christos  and
	Chakraborty, Tanmoy  and
	Rose, Carolyn  and
	Peng, Violet",
	booktitle = "Proceedings of the 2025 Conference on Empirical Methods in Natural Language Processing",
	month = nov,
	year = "2025",
	address = "Suzhou, China",
	publisher = "Association for Computational Linguistics",
	url = "https://aclanthology.org/2025.emnlp-main.622/",
	doi = "10.18653/v1/2025.emnlp-main.622",
	pages = "12339--12346",
	ISBN = "979-8-89176-332-6"
}

@inproceedings{gao-etal-2025-tlue,
	title = "{TLUE}: A {T}ibetan Language Understanding Evaluation Benchmark",
	author = "Gao, Fan  and
	Huang, Cheng  and
	Liu, Yutong  and
	Tashi, Nyima  and
	Wang, Xiangxiang  and
	Tsering, Thupten  and
	Ma-bao, Ban  and
	Duojie, Renzeng  and
	Luosang, Gadeng  and
	Dongrub, Rinchen  and
	Tashi, Dorje  and
	Cd, Xiao Feng  and
	Yu, Yongbin  and
	Wang, Hao",
	editor = "Christodoulopoulos, Christos  and
	Chakraborty, Tanmoy  and
	Rose, Carolyn  and
	Peng, Violet",
	booktitle = "Proceedings of the 2025 Conference on Empirical Methods in Natural Language Processing",
	month = nov,
	year = "2025",
	address = "Suzhou, China",
	publisher = "Association for Computational Linguistics",
	url = "https://aclanthology.org/2025.emnlp-main.1777/",
	doi = "10.18653/v1/2025.emnlp-main.1777",
	pages = "35071--35097",
	ISBN = "979-8-89176-332-6"
}

@inproceedings{zhang-etal-2025-milic,
	title = "{M}i{L}i{C}-Eval: Benchmarking Multilingual {LLM}s for {C}hina{'}s Minority Languages",
	author = "Zhang, Chen  and
	Tao, Mingxu  and
	Liao, Zhiyuan  and
	Feng, Yansong",
	editor = "Che, Wanxiang  and
	Nabende, Joyce  and
	Shutova, Ekaterina  and
	Pilehvar, Mohammad Taher",
	booktitle = "Findings of the Association for Computational Linguistics: ACL 2025",
	month = jul,
	year = "2025",
	address = "Vienna, Austria",
	publisher = "Association for Computational Linguistics",
	url = "https://aclanthology.org/2025.findings-acl.578/",
	doi = "10.18653/v1/2025.findings-acl.578",
	pages = "11086--11102",
	ISBN = "979-8-89176-256-5"
}

@inproceedings{zhuang-sun-2025-cute,
	title = "{CUTE}: A Multilingual Dataset for Enhancing Cross-Lingual Knowledge Transfer in Low-Resource Languages",
	author = "Zhuang, Wenhao  and
	Sun, Yuan",
	editor = "Rambow, Owen  and
	Wanner, Leo  and
	Apidianaki, Marianna  and
	Al-Khalifa, Hend  and
	Eugenio, Barbara Di  and
	Schockaert, Steven",
	booktitle = "Proceedings of the 31st International Conference on Computational Linguistics",
	month = jan,
	year = "2025",
	address = "Abu Dhabi, UAE",
	publisher = "Association for Computational Linguistics",
	url = "https://aclanthology.org/2025.coling-main.670/",
	pages = "10037--10046"
}

@inproceedings{zhang-etal-2024-mc2,
	title = "{MC}$^2$: Towards Transparent and Culturally-Aware {NLP} for Minority Languages in {C}hina",
	author = "Zhang, Chen  and
	Tao, Mingxu  and
	Huang, Quzhe  and
	Lin, Jiuheng  and
	Chen, Zhibin  and
	Feng, Yansong",
	editor = "Ku, Lun-Wei  and
	Martins, Andre  and
	Srikumar, Vivek",
	booktitle = "Proceedings of the 62nd Annual Meeting of the Association for Computational Linguistics (Volume 1: Long Papers)",
	month = aug,
	year = "2024",
	address = "Bangkok, Thailand",
	publisher = "Association for Computational Linguistics",
	url = "https://aclanthology.org/2024.acl-long.479/",
	doi = "10.18653/v1/2024.acl-long.479",
	pages = "8832--8850"
}

@inproceedings{gurgurov-etal-2024-adapting,
	title = "Adapting Multilingual {LLM}s to Low-Resource Languages with Knowledge Graphs via Adapters",
	author = "Gurgurov, Daniil  and
	Hartmann, Mareike  and
	Ostermann, Simon",
	editor = "Biswas, Russa  and
	Kaffee, Lucie-Aim{\'e}e  and
	Agarwal, Oshin  and
	Minervini, Pasquale  and
	Singh, Sameer  and
	de Melo, Gerard",
	booktitle = "Proceedings of the 1st Workshop on Knowledge Graphs and Large Language Models (KaLLM 2024)",
	month = aug,
	year = "2024",
	address = "Bangkok, Thailand",
	publisher = "Association for Computational Linguistics",
	url = "https://aclanthology.org/2024.kallm-1.7/",
	doi = "10.18653/v1/2024.kallm-1.7",
	pages = "63--74"
}

@inproceedings{brown-2024-improved,
	title = "Improved Neural Word Segmentation for Standard {T}ibetan",
	author = "Brown, Collin J.",
	editor = "Ojha, Atul Kr.  and
	Ahmadi, Sina  and
	Cinkov{\'a}, Silvie  and
	Fransen, Theodorus  and
	Liu, Chao-Hong  and
	McCrae, John P.",
	booktitle = "Proceedings of the 2nd Workshop on Resources and Technologies for Indigenous, Endangered and Lesser-resourced Languages in Eurasia (EURALI) @ LREC-COLING 2024",
	month = may,
	year = "2024",
	address = "Torino, Italia",
	publisher = "ELRA and ICCL",
	url = "https://aclanthology.org/2024.eurali-1.2/",
	pages = "12--17"
}

@inproceedings{an-2023-tibetan,
	title = "{T}ibetan Dependency Parsing with Graph Convolutional Neural Networks",
	author = "An, Bo",
	editor = "Anderson, Adam  and
	Gordin, Shai  and
	Li, Bin  and
	Liu, Yudong  and
	Passarotti, Marco C.",
	booktitle = "Proceedings of the Ancient Language Processing Workshop",
	month = sep,
	year = "2023",
	address = "Varna, Bulgaria",
	publisher = "INCOMA Ltd., Shoumen, Bulgaria",
	url = "https://aclanthology.org/2023.alp-1.24/",
	pages = "213--221"
}

@inproceedings{long-an-2023-development,
	title = "On the Development of Interlinearized Ancient Literature of Ethnic Minorities: A Case Study of the Interlinearization of Ancient Written {T}ibetan Literature",
	author = "Long, Congjun  and
	An, Bo",
	editor = "Anderson, Adam  and
	Gordin, Shai  and
	Li, Bin  and
	Liu, Yudong  and
	Passarotti, Marco C.",
	booktitle = "Proceedings of the Ancient Language Processing Workshop",
	month = sep,
	year = "2023",
	address = "Varna, Bulgaria",
	publisher = "INCOMA Ltd., Shoumen, Bulgaria",
	url = "https://aclanthology.org/2023.alp-1.25/",
	pages = "222--231"
}

@inproceedings{cao-etal-2023-pay-attention,
	title = "Pay Attention to the Robustness of {C}hinese Minority Language Models! {S}yllable-level Textual Adversarial Attack on {T}ibetan Script",
	author = "Cao, Xi  and
	Dawa, Dolma  and
	Qun, Nuo  and
	Nyima, Trashi",
	editor = "Ovalle, Anaelia  and
	Chang, Kai-Wei  and
	Mehrabi, Ninareh  and
	Pruksachatkun, Yada  and
	Galystan, Aram  and
	Dhamala, Jwala  and
	Verma, Apurv  and
	Cao, Trista  and
	Kumar, Anoop  and
	Gupta, Rahul",
	booktitle = "Proceedings of the 3rd Workshop on Trustworthy Natural Language Processing (TrustNLP 2023)",
	month = jul,
	year = "2023",
	address = "Toronto, Canada",
	publisher = "Association for Computational Linguistics",
	url = "https://aclanthology.org/2023.trustnlp-1.4/",
	doi = "10.18653/v1/2023.trustnlp-1.4",
	pages = "35--46"
}

@inproceedings{sun-etal-2022-question,
	title = "Question Generation Based on Grammar Knowledge and Fine-grained Classification",
	author = "Sun, Yuan  and
	Liu, Sisi  and
	Dan, Zhengcuo  and
	Zhao, Xiaobing",
	editor = "Calzolari, Nicoletta  and
	Huang, Chu-Ren  and
	Kim, Hansaem  and
	Pustejovsky, James  and
	Wanner, Leo  and
	Choi, Key-Sun  and
	Ryu, Pum-Mo  and
	Chen, Hsin-Hsi  and
	Donatelli, Lucia  and
	Ji, Heng  and
	Kurohashi, Sadao  and
	Paggio, Patrizia  and
	Xue, Nianwen  and
	Kim, Seokhwan  and
	Hahm, Younggyun  and
	He, Zhong  and
	Lee, Tony Kyungil  and
	Santus, Enrico  and
	Bond, Francis  and
	Na, Seung-Hoon",
	booktitle = "Proceedings of the 29th International Conference on Computational Linguistics",
	month = oct,
	year = "2022",
	address = "Gyeongju, Republic of Korea",
	publisher = "International Committee on Computational Linguistics",
	url = "https://aclanthology.org/2022.coling-1.562/",
	pages = "6457--6467"
}

@inproceedings{nehrdich-2022-sanstib,
	title = "{S}ans{T}ib, a {S}anskrit - {T}ibetan Parallel Corpus and Bilingual Sentence Embedding Model",
	author = "Nehrdich, Sebastian",
	editor = "Calzolari, Nicoletta  and
	B{\'e}chet, Fr{\'e}d{\'e}ric  and
	Blache, Philippe  and
	Choukri, Khalid  and
	Cieri, Christopher  and
	Declerck, Thierry  and
	Goggi, Sara  and
	Isahara, Hitoshi  and
	Maegaard, Bente  and
	Mariani, Joseph  and
	Mazo, H{\'e}l{\`e}ne  and
	Odijk, Jan  and
	Piperidis, Stelios",
	booktitle = "Proceedings of the Thirteenth Language Resources and Evaluation Conference",
	month = jun,
	year = "2022",
	address = "Marseille, France",
	publisher = "European Language Resources Association",
	url = "https://aclanthology.org/2022.lrec-1.724/",
	pages = "6728--6734"
}

@inproceedings{faggionato-etal-2022-nlp,
	title = "{NLP} Pipeline for Annotating (Endangered) {T}ibetan and {N}ewar Varieties",
	author = "Faggionato, Christian  and
	Hill, Nathan  and
	Meelen, Marieke",
	editor = "Ojha, Atul Kr.  and
	Ahmadi, Sina  and
	Liu, Chao-Hong  and
	McCrae, John P.",
	booktitle = "Proceedings of the Workshop on Resources and Technologies for Indigenous, Endangered and Lesser-resourced Languages in Eurasia within the 13th Language Resources and Evaluation Conference",
	month = jun,
	year = "2022",
	address = "Marseille, France",
	publisher = "European Language Resources Association",
	url = "https://aclanthology.org/2022.eurali-1.1/",
	pages = "1--6"
}

@inproceedings{kuanzhuo-etal-2020-simplenlg,
	title = "{S}imple{NLG}-{TI}: Adapting {S}imple{NLG} to {T}ibetan",
	author = "Kuanzhuo, Zewang  and
	Lin, Li  and
	Weina, Zhao",
	editor = "Davis, Brian  and
	Graham, Yvette  and
	Kelleher, John  and
	Sripada, Yaji",
	booktitle = "Proceedings of the 13th International Conference on Natural Language Generation",
	month = dec,
	year = "2020",
	address = "Dublin, Ireland",
	publisher = "Association for Computational Linguistics",
	url = "https://aclanthology.org/2020.inlg-1.12/",
	doi = "10.18653/v1/2020.inlg-1.12",
	pages = "86--90"
}

@inproceedings{meelen-roux-2020-meta,
	title = "Meta-dating the {PA}rsed Corpus of {T}ibetan ({PACT}ib)",
	author = "Meelen, Marieke  and
	Roux, {\'E}lie",
	editor = "Evang, Kilian  and
	Kallmeyer, Laura  and
	Ehren, Rafael  and
	Petitjean, Simon  and
	Seyffarth, Esther  and
	Seddah, Djam{\'e}",
	booktitle = "Proceedings of the 19th International Workshop on Treebanks and Linguistic Theories",
	month = oct,
	year = "2020",
	address = {D{\"u}sseldorf, Germany},
	publisher = "Association for Computational Linguistics",
	url = "https://aclanthology.org/2020.tlt-1.3/",
	doi = "10.18653/v1/2020.tlt-1.3",
	pages = "31--42"
}

@inproceedings{di-etal-2019-modelling,
	title = "Modelling {T}ibetan Verbal Morphology",
	author = "Di, Qianji  and
	Vylomova, Ekaterina  and
	Baldwin, Tim",
	editor = "Mistica, Meladel  and
	Piccardi, Massimo  and
	MacKinlay, Andrew",
	booktitle = "Proceedings of the 17th Annual Workshop of the Australasian Language Technology Association",
	month = "4--6 " # dec,
	year = "2019",
	address = "Sydney, Australia",
	publisher = "Australasian Language Technology Association",
	url = "https://aclanthology.org/U19-1005/",
	pages = "35--40"
}

@inproceedings{faggionato-meelen-2019-developing,
	title = "Developing the {O}ld {T}ibetan Treebank",
	author = "Faggionato, Christian  and
	Meelen, Marieke",
	editor = "Mitkov, Ruslan  and
	Angelova, Galia",
	booktitle = "Proceedings of the International Conference on Recent Advances in Natural Language Processing (RANLP 2019)",
	month = sep,
	year = "2019",
	address = "Varna, Bulgaria",
	publisher = "INCOMA Ltd.",
	url = "https://aclanthology.org/R19-1035/",
	doi = "10.26615/978-954-452-056-4_035",
	pages = "304--312"
}

@inproceedings{lai-etal-2018-tibetan,
	title = "{T}ibetan-{C}hinese Neural Machine Translation based on Syllable Segmentation",
	author = "Lai, Wen  and
	Zhao, Xiaobing  and
	Bao, Wei",
	editor = "Liu, Chao-Hong",
	booktitle = "Proceedings of the {AMTA} 2018 Workshop on Technologies for {MT} of Low Resource Languages ({L}o{R}es{MT} 2018)",
	month = mar,
	year = "2018",
	address = "Boston, MA",
	publisher = "Association for Machine Translation in the Americas",
	url = "https://aclanthology.org/W18-2203/",
	pages = "21--29"
}

@inproceedings{nuo-etal-2015-tibetan,
	title = "{T}ibetan Unknown Word Identification from News Corpora for Supporting Lexicon-based {T}ibetan Word Segmentation",
	author = "Nuo, Minghua  and
	Liu, Huidan  and
	Long, Congjun  and
	Wu, Jian",
	editor = "Zong, Chengqing  and
	Strube, Michael",
	booktitle = "Proceedings of the 53rd Annual Meeting of the Association for Computational Linguistics and the 7th International Joint Conference on Natural Language Processing (Volume 2: Short Papers)",
	month = jul,
	year = "2015",
	address = "Beijing, China",
	publisher = "Association for Computational Linguistics",
	url = "https://aclanthology.org/P15-2074/",
	doi = "10.3115/v1/P15-2074",
	pages = "451--457"
}

@inproceedings{liu-etal-2014-zipfs,
	title = "{Z}ipf{'}s Law and Statistical Data on {M}odern {T}ibetan",
	author = "Liu, Huidan  and
	Nuo, Minghua  and
	Wu, Jian",
	editor = "Tsujii, Junichi  and
	Hajic, Jan",
	booktitle = "Proceedings of {COLING} 2014, the 25th International Conference on Computational Linguistics: Technical Papers",
	month = aug,
	year = "2014",
	address = "Dublin, Ireland",
	publisher = "Dublin City University and Association for Computational Linguistics",
	url = "https://aclanthology.org/C14-1032/",
	pages = "322--333"
}

@inproceedings{nuo-etal-2012-tibetan,
	title = "{T}ibetan Base Noun Phrase Identification Framework Based on {C}hinese-{T}ibetan Sentence Aligned Corpus",
	author = "Nuo, Ming Hua  and
	Liu, Hui Dan  and
	Zhao, Wei Na  and
	Ma, Long Long  and
	Wu, Jian  and
	Ding, Zhi Ming",
	editor = "Kay, Martin  and
	Boitet, Christian",
	booktitle = "Proceedings of {COLING} 2012",
	month = dec,
	year = "2012",
	address = "Mumbai, India",
	publisher = "The COLING 2012 Organizing Committee",
	url = "https://aclanthology.org/C12-1131/",
	pages = "2141--2158"
}

@inproceedings{liu-etal-2012-building,
	title = "Building Large Scale Text Corpus for {T}ibetan Natural Language Processing by Extracting Text from Web Pages",
	author = "Liu, Huidan  and
	Nuo, Minghua  and
	Wu, Jian  and
	He, Yeping",
	editor = "Weerasinghe, Ruvan  and
	Hussain, Sarmad  and
	Sornlertlamvanich, Virach  and
	Roxas, Rachel Edita O.",
	booktitle = "Proceedings of the 10th Workshop on {A}sian Language Resources",
	month = dec,
	year = "2012",
	address = "Mumbai, India",
	publisher = "The COLING 2012 Organizing Committee",
	url = "https://aclanthology.org/W12-5202/",
	pages = "11--20"
}

@inproceedings{liu-etal-2011-tibetan,
	title = "{T}ibetan Word Segmentation as Syllable Tagging Using Conditional Random Field",
	author = "Liu, Huidan  and
	Nuo, Minghua  and
	Ma, Longlong  and
	Wu, Jian  and
	He, Yeping",
	editor = "Gao, Helena Hong  and
	Dong, Minghui",
	booktitle = "Proceedings of the 25th Pacific Asia Conference on Language, Information and Computation",
	month = dec,
	year = "2011",
	address = "Singapore",
	publisher = "Institute of Digital Enhancement of Cognitive Processing, Waseda University",
	url = "https://aclanthology.org/Y11-1018/",
	pages = "168--177"
}

@inproceedings{liu-etal-2010-tibetan,
	title = "{T}ibetan Number Identification Based on Classification of Number Components in {T}ibetan Word Segmentation",
	author = "Liu, Huidan  and
	Zhao, Weina  and
	Nuo, Minghua  and
	Jiang, Li  and
	Wu, Jian  and
	He, Yeping",
	editor = "Huang, Chu-Ren  and
	Jurafsky, Dan",
	booktitle = "Coling 2010: Posters",
	month = aug,
	year = "2010",
	address = "Beijing, China",
	publisher = "Coling 2010 Organizing Committee",
	url = "https://aclanthology.org/C10-2082/",
	pages = "719--724"
}

@inproceedings{jiang-2006-current,
	title = "The Current Status of Sorting Order of {T}ibetan Dictionaries and Standardization",
	author = "Jiang, Di",
	booktitle = "Proceedings of the 20th Pacific Asia Conference on Language, Information and Computation",
	month = nov,
	year = "2006",
	address = "Huazhong Normal University, Wuhan, China",
	publisher = "Tsinghua University Press",
	url = "https://aclanthology.org/Y06-1030/",
	doi = "http://hdl.handle.net/2065/29059",
	pages = "231--236"
}

@inproceedings{jiang-etal-2005-verbal,
	title = "The Verbal Entries and Their Description in a Grammatical Information-Dictionary of Contemporary {T}ibetan",
	author = "Jiang, Di  and
	Long, Congjun  and
	Zhang, Jichuan",
	booktitle = "Second International Joint Conference on Natural Language Processing: Full Papers",
	year = "2005",
	url = "https://aclanthology.org/I05-1076/",
	doi = "10.1007/11562214_76"
}

@inproceedings{wagner-zeisler-2004-syntactically,
	title = "A Syntactically Annotated Corpus of {T}ibetan",
	author = "Wagner, Andreas  and
	Zeisler, Bettina",
	editor = "Lino, Maria Teresa  and
	Xavier, Maria Francisca  and
	Ferreira, F{\'a}tima  and
	Costa, Rute  and
	Silva, Raquel",
	booktitle = "Proceedings of the Fourth International Conference on Language Resources and Evaluation ({LREC}{'}04)",
	month = may,
	year = "2004",
	address = "Lisbon, Portugal",
	publisher = "European Language Resources Association (ELRA)",
	url = "https://aclanthology.org/L04-1156/"
}

@inproceedings{sharif_balti-tamko_2025,
	series = {Lecture {Notes} in {Computer} {Science}},
	title = {Balti-{Tamko}: {A} {Spoken} {Words} {Dataset} {Development} for {Automatic} {Recognition} of the {Endangered} {Balti} {Language}},
	volume = {16313},
	url = {https://doi.org/10.1007/978-981-95-4445-5\_10},
	doi = {10.1007/978-981-95-4445-5_10},
	booktitle = {Neural {Information} {Processing} - 32nd {International} {Conference}, {ICONIP} 2025, {Okinawa}, {Japan}, {November} 20-24, 2025, {Proceedings}, {Part} {V}},
	publisher = {Springer},
	author = {Sharif, Muhammad and Liu, Bin and Naqvi, Sardar Shan Ali and Abbas, Zeeshan and Liu, Cheng-Lin},
	editor = {Taniguchi, Tadahiro and Leung, Andrew Chi-Sing and Kozuno, Tadashi and Yoshimoto, Junichiro and Mahmud, Mufti and Doborjeh, Maryam and Doya, Kenji},
	year = {2025},
	pages = {140--150},
}

@inproceedings{ashida-etal-2020-building,
	title = "Building a Part-of-Speech Tagged Corpus for Drenjongke (Bhutia)",
	author = "Ashida, Mana  and
	Lee, Seunghun  and
	Namgyal, Kunzang",
	editor = "Shmueli, Boaz  and
	Huang, Yin Jou",
	booktitle = "Proceedings of the 1st Conference of the Asia-Pacific Chapter of the Association for Computational Linguistics and the 10th International Joint Conference on Natural Language Processing: Student Research Workshop",
	month = dec,
	year = "2020",
	address = "Suzhou, China",
	publisher = "Association for Computational Linguistics",
	url = "https://aclanthology.org/2020.aacl-srw.9/",
	doi = "10.18653/v1/2020.aacl-srw.9",
	pages = "57--63"
}

@inproceedings{cao_tscheater_2025,
	title = {{TSCheater}: {Generating} {High}-{Quality} {Tibetan} {Adversarial} {Texts} via {Visual} {Similarity}},
	url = {https://doi.org/10.1109/ICASSP49660.2025.10889732},
	doi = {10.1109/ICASSP49660.2025.10889732},
	booktitle = {2025 {IEEE} {International} {Conference} on {Acoustics}, {Speech} and {Signal} {Processing}, {ICASSP} 2025, {Hyderabad}, {India}, {April} 6-11, 2025},
	publisher = {IEEE},
	author = {Cao, Xi and Gesang, Quzong and Sun, Yuan and Qun, Nuo and Nyima, Tashi},
	year = {2025},
	pages = {1--5},
}

@inproceedings{wang_disg_2025,
	series = {Lecture {Notes} in {Computer} {Science}},
	title = {{DiSG}: {A} {Discourse} {Structure}-{Aware} {Multi}-stage {Approach} for {Long} {Tibetan} {Text} {Summarization}},
	volume = {16105},
	url = {https://doi.org/10.1007/978-981-95-3352-7\_13},
	doi = {10.1007/978-981-95-3352-7_13},
	booktitle = {Natural {Language} {Processing} and {Chinese} {Computing} - 14th {National} {CCF} {Conference}, {NLPCC} 2025, {Urumqi}, {China}, {August} 7-9, 2025, {Proceedings}, {Part} {IV}},
	publisher = {Springer},
	author = {Wang, Yiwen and Hao, Yanrong and Chen, Bo and Xu, Yang and Zhao, Xiaobing},
	editor = {Mao, Xian-Ling and Ren, Zhaochun and Yang, Muyun},
	year = {2025},
	pages = {159--171},
}

@inproceedings{xu_tibetan_2024,
	title = {Tibetan microblogging user data analysis and topic identification},
	url = {https://doi.org/10.1109/TrustCom63139.2024.00154},
	doi = {10.1109/TRUSTCOM63139.2024.00154},
	booktitle = {23rd {IEEE} {International} {Conference} on {Trust}, {Security} and {Privacy} in {Computing} and {Communications}, {TrustCom} 2024, {Sanya}, {China}, {December} 17-21, 2024},
	publisher = {IEEE},
	author = {Xu, Guixian and Gao, Wenhui},
	year = {2024},
	pages = {1077--1082},
}

@inproceedings{hao_tibetan_2024,
	title = {A {Tibetan} {Text} {Classification} {Method} {Based} on {Hybrid} {Model} and {Channel} {Attention} {Mechanism}},
	url = {https://doi.org/10.1109/CSCWD61410.2024.10580150},
	doi = {10.1109/CSCWD61410.2024.10580150},
	booktitle = {27th {International} {Conference} on {Computer} {Supported} {Cooperative} {Work} in {Design}, {CSCWD} 2024, {Tianjin}, {China}, {May} 8-10, 2024},
	publisher = {IEEE},
	author = {Hao, Minghui and Yan, Xiaodong and Ouyang, Xinpeng},
	editor = {Shen, Weiming and Barthès, Jean-Paul A. and Luo, Junzhou and Qiu, Tie and Zhou, Xiaobo and Zhang, Jinghui and Zhu, Haibin and Peng, Kunkun and Xu, Tianyi and Chen, Ning},
	year = {2024},
	pages = {1522--1527},
}

@inproceedings{ouyang_pre-trained_2024,
	title = {A {Pre}-{Trained} {Language} {Model} {Based} on {LED} for {Tibetan} {Long} {Text} {Summarization}},
	url = {https://doi.org/10.1109/CSCWD61410.2024.10580451},
	doi = {10.1109/CSCWD61410.2024.10580451},
	booktitle = {27th {International} {Conference} on {Computer} {Supported} {Cooperative} {Work} in {Design}, {CSCWD} 2024, {Tianjin}, {China}, {May} 8-10, 2024},
	publisher = {IEEE},
	author = {Ouyang, Xinpeng and Yan, Xiaodong and Hao, Minghui},
	editor = {Shen, Weiming and Barthès, Jean-Paul A. and Luo, Junzhou and Qiu, Tie and Zhou, Xiaobo and Zhang, Jinghui and Zhu, Haibin and Peng, Kunkun and Xu, Tianyi and Chen, Ning},
	year = {2024},
	pages = {992--997},
}

@inproceedings{zhou_efficient_2024,
	series = {Lecture {Notes} in {Computer} {Science}},
	title = {Efficient {Fine}-{Tuning} for {Low}-{Resource} {Tibetan} {Pre}-trained {Language} {Models}},
	volume = {15022},
	url = {https://doi.org/10.1007/978-3-031-72350-6\_28},
	doi = {10.1007/978-3-031-72350-6_28},
	booktitle = {Artificial {Neural} {Networks} and {Machine} {Learning} - {ICANN} 2024 - 33rd {International} {Conference} on {Artificial} {Neural} {Networks}, {Lugano}, {Switzerland}, {September} 17-20, 2024, {Proceedings}, {Part} {VII}},
	publisher = {Springer},
	author = {Zhou, Mingjun and Zhuoma, Daiqing and Qun, Nuo and Nyima, Tashi},
	editor = {Wand, Michael and Malinovská, Kristína and Schmidhuber, Jürgen and Tetko, Igor V.},
	year = {2024},
	pages = {410--422},
}

@inproceedings{fang_mlcl_2024,
	series = {Proceedings of {Machine} {Learning} {Research}},
	title = {{MLCL}: {A} {Framework} for {Reducing} {Language} {Imbalance} in {Sino}-{Tibetan} {Languages} through {Adapter} {Structures}},
	volume = {260},
	url = {https://proceedings.mlr.press/v260/fang25a.html},
	booktitle = {Asian {Conference} on {Machine} {Learning}, 5-8 {December} 2024, {Hanoi}, {Vietnam}},
	publisher = {PMLR},
	author = {Fang, Jiajun and Huang, Wentao and Yang, Aimin and Zhou, Dong and Lin, Nankai},
	editor = {Nguyen, Vu and Lin, Hsuan-Tien},
	year = {2024},
	pages = {431--446},
}

@inproceedings{zhou_tibetan-chinese_2024,
	title = {Tibetan-{Chinese} {Machine} {Translation} {Enhanced} on {Cross}-{Lingual} {Pre}-{Trained} {Model}},
	url = {https://doi.org/10.1109/SMC54092.2024.11169736},
	doi = {10.1109/SMC54092.2024.11169736},
	booktitle = {{IEEE} {International} {Conference} on {Systems}, {Man}, and {Cybernetics}, {SMC} 2024, {Kuching}, {Malaysia}, {October} 6-10, 2024},
	publisher = {IEEE},
	author = {Zhou, Mingjun and Gesang, Quzong and Qun, Nuo and Nyima, Tashi and Dongrub, Rinchen},
	year = {2024},
	pages = {1618--1623},
}

@inproceedings{ren_method_2024,
	title = {A {Method} for {Tibetan} {Offensive} {Language} {Detection} {Based} on {Prompt} {Learning} and {Information} {Theory}},
	url = {https://doi.org/10.1109/IJCNN60899.2024.10651369},
	doi = {10.1109/IJCNN60899.2024.10651369},
	booktitle = {International {Joint} {Conference} on {Neural} {Networks}, {IJCNN} 2024, {Yokohama}, {Japan}, {June} 30 - {July} 5, 2024},
	publisher = {IEEE},
	author = {Ren, Hang and Yang, Jin and Ni, Shengqiao and Yang, Qin and Zhang, Jing and Qun, Nuo},
	year = {2024},
	pages = {1--8},
}

@inproceedings{hao_tiltstibetan_2024,
	series = {Lecture {Notes} in {Computer} {Science}},
	title = {{TiLTS}:{Tibetan} {Long} {Text} {Summarization} {Dataset}},
	volume = {15362},
	url = {https://doi.org/10.1007/978-981-97-9440-9\_21},
	doi = {10.1007/978-981-97-9440-9_21},
	booktitle = {Natural {Language} {Processing} and {Chinese} {Computing} - 13th {National} {CCF} {Conference}, {NLPCC} 2024, {Hangzhou}, {China}, {November} 1-3, 2024, {Proceedings}, {Part} {IV}},
	publisher = {Springer},
	author = {Hao, Yanrong and Chen, Bo and Zhao, Xiaobing},
	editor = {Wong, Derek F. and Wei, Zhongyu and Yang, Muyun},
	year = {2024},
	pages = {265--276},
}

@inproceedings{li_tibetan_2024,
	series = {Lecture {Notes} in {Computer} {Science}},
	title = {Tibetan {Handwriting} {Recognition} {Method} {Based} on {Structural} {Re}-{Parameterization} {ViT} and {Vertical} {Attention}},
	volume = {15037},
	url = {https://doi.org/10.1007/978-981-97-8511-7\_12},
	doi = {10.1007/978-981-97-8511-7_12},
	booktitle = {Pattern {Recognition} and {Computer} {Vision} - 7th {Chinese} {Conference}, {PRCV} 2024, {Urumqi}, {China}, {October} 18-20, 2024, {Proceedings}, {Part} {VII}},
	publisher = {Springer},
	author = {Li, Binglin and Zhu, Jie and Zhao, Dongcai},
	editor = {Lin, Zhouchen and Cheng, Ming-Ming and He, Ran and Ubul, Kurban and Silamu, Wushouer and Zha, Hongbin and Zhou, Jie and Liu, Cheng-Lin},
	year = {2024},
	pages = {156--169},
}

@inproceedings{liu_multi-task_2023,
	series = {Lecture {Notes} in {Computer} {Science}},
	title = {Multi-task {Pre}-training for {Lhasa}-{Tibetan} {Speech} {Recognition}},
	volume = {14262},
	url = {https://doi.org/10.1007/978-3-031-44201-8\_7},
	doi = {10.1007/978-3-031-44201-8_7},
	booktitle = {Artificial {Neural} {Networks} and {Machine} {Learning} - {ICANN} 2023 - 32nd {International} {Conference} on {Artificial} {Neural} {Networks}, {Heraklion}, {Crete}, {Greece}, {September} 26-29, 2023, {Proceedings}, {Part} {IX}},
	publisher = {Springer},
	author = {Liu, Yigang and Zhao, Yue and Xu, Xiaona and Xu, Liang and Zhang, Xubei},
	editor = {Iliadis, Lazaros and Papaleonidas, Antonios and Angelov, Plamen and Jayne, Chrisina},
	year = {2023},
	pages = {78--90},
}

@inproceedings{qin_finer-grained_2022,
	title = {Finer-grained {Modeling} units-based {Meta}-{Learning} for {Low}-resource {Tibetan} {Speech} {Recognition}},
	url = {https://doi.org/10.21437/Interspeech.2022-10015},
	doi = {10.21437/INTERSPEECH.2022-10015},
	booktitle = {23rd {Annual} {Conference} of the {International} {Speech} {Communication} {Association}, {Interspeech} 2022, {Incheon}, {Korea}, {September} 18-22, 2022},
	publisher = {ISCA},
	author = {Qin, Siqing and Wang, Longbiao and Li, Sheng and Lin, Yuqin and Dang, Jianwu},
	editor = {Ko, Hanseok and Hansen, John H. L.},
	year = {2022},
	pages = {2133--2137},
}

@inproceedings{liu_tibert_2022,
	title = {{TiBERT}: {Tibetan} {Pre}-trained {Language} {Model}},
	url = {https://doi.org/10.1109/SMC53654.2022.9945074},
	doi = {10.1109/SMC53654.2022.9945074},
	booktitle = {{IEEE} {International} {Conference} on {Systems}, {Man}, and {Cybernetics}, {SMC} 2022, {Prague}, {Czech} {Republic}, {October} 9-12, 2022},
	publisher = {IEEE},
	author = {Liu, Sisi and Deng, Junjie and Sun, Yuan and Zhao, Xiaobing},
	year = {2022},
	pages = {2956--2961},
}

@inproceedings{zhu_new_2019,
	series = {Lecture {Notes} in {Computer} {Science}},
	title = {A {New} {Algorithm} for {Component} {Decomposition} and {Type} {Recognition} of {Tibetan} {Syllable}},
	volume = {11839},
	url = {https://doi.org/10.1007/978-3-030-32236-6\_34},
	doi = {10.1007/978-3-030-32236-6_34},
	booktitle = {Natural {Language} {Processing} and {Chinese} {Computing} - 8th {CCF} {International} {Conference}, {NLPCC} 2019, {Dunhuang}, {China}, {October} 9-14, 2019, {Proceedings}, {Part} {II}},
	publisher = {Springer},
	author = {Zhu, Jie and Wang, Shugen and Wu, Yanru and Guan, Meijing and Feng, Yang},
	editor = {Tang, Jie and Kan, Min-Yen and Zhao, Dongyan and Li, Sujian and Zan, Hongying},
	year = {2019},
	pages = {381--392},
}

@inproceedings{cai_online_2018,
	series = {Lecture {Notes} in {Computer} {Science}},
	title = {Online {Handwriting} {Tibetan} {Character} {Recognition} {Based} on {Two}-{Dimensional} {Discriminant} {Locality} {Alignment}},
	volume = {11258},
	url = {https://doi.org/10.1007/978-3-030-03338-5\_8},
	doi = {10.1007/978-3-030-03338-5_8},
	booktitle = {Pattern {Recognition} and {Computer} {Vision} - {First} {Chinese} {Conference}, {PRCV} 2018, {Guangzhou}, {China}, {November} 23-26, 2018, {Proceedings}, {Part} {III}},
	publisher = {Springer},
	author = {Cai, Zhengqi and Wang, Weilan},
	editor = {Lai, Jian-Huang and Liu, Cheng-Lin and Chen, Xilin and Zhou, Jie and Tan, Tieniu and Zheng, Nanning and Zha, Hongbin},
	year = {2018},
	pages = {88--98},
}

@inproceedings{han_research_2018,
	series = {Lecture {Notes} in {Computer} {Science}},
	title = {Research on the {Method} of {Tibetan} {Recognition} {Based} on {Component} {Location} {Information}},
	volume = {11258},
	url = {https://doi.org/10.1007/978-3-030-03338-5\_6},
	doi = {10.1007/978-3-030-03338-5_6},
	booktitle = {Pattern {Recognition} and {Computer} {Vision} - {First} {Chinese} {Conference}, {PRCV} 2018, {Guangzhou}, {China}, {November} 23-26, 2018, {Proceedings}, {Part} {III}},
	publisher = {Springer},
	author = {Han, Yuehui and Wang, Weilan and Wang, Yiqun and Wang, Xiaojuan},
	editor = {Lai, Jian-Huang and Liu, Cheng-Lin and Chen, Xilin and Zhou, Jie and Tan, Tieniu and Zheng, Nanning and Zha, Hongbin},
	year = {2018},
	pages = {63--73},
}

@inproceedings{wang_recognition_2018,
	series = {Lecture {Notes} in {Computer} {Science}},
	title = {A {Recognition} {Method} of the {Similarity} {Character} for {Uchen} {Script} {Tibetan} {Historical} {Document} {Based} on {DNN}},
	volume = {11258},
	url = {https://doi.org/10.1007/978-3-030-03338-5\_5},
	doi = {10.1007/978-3-030-03338-5_5},
	booktitle = {Pattern {Recognition} and {Computer} {Vision} - {First} {Chinese} {Conference}, {PRCV} 2018, {Guangzhou}, {China}, {November} 23-26, 2018, {Proceedings}, {Part} {III}},
	publisher = {Springer},
	author = {Wang, Xiaojuan and Wang, Weilan and Li, Zhenjiang and Wang, Yiqun and Han, Yuehui and Hao, Zhanjun},
	editor = {Lai, Jian-Huang and Liu, Cheng-Lin and Chen, Xilin and Zhou, Jie and Tan, Tieniu and Zheng, Nanning and Zha, Hongbin},
	year = {2018},
	pages = {52--62},
}

@inproceedings{wang_research_2018,
	series = {Lecture {Notes} in {Computer} {Science}},
	title = {Research on {Text} {Line} {Segmentation} of {Historical} {Tibetan} {Documents} {Based} on the {Connected} {Component} {Analysis}},
	volume = {11258},
	url = {https://doi.org/10.1007/978-3-030-03338-5\_7},
	doi = {10.1007/978-3-030-03338-5_7},
	booktitle = {Pattern {Recognition} and {Computer} {Vision} - {First} {Chinese} {Conference}, {PRCV} 2018, {Guangzhou}, {China}, {November} 23-26, 2018, {Proceedings}, {Part} {III}},
	publisher = {Springer},
	author = {Wang, Yiqun and Wang, Weilan and Li, Zhenjiang and Han, Yuehui and Wang, Xiaojuan},
	editor = {Lai, Jian-Huang and Liu, Cheng-Lin and Chen, Xilin and Zhou, Jie and Tan, Tieniu and Zheng, Nanning and Zha, Hongbin},
	year = {2018},
	pages = {74--87},
}

@inproceedings{zhao_touching_2018,
	series = {Lecture {Notes} in {Computer} {Science}},
	title = {A {Touching} {Character} {Database} from {Tibetan} {Historical} {Documents} to {Evaluate} the {Segmentation} {Algorithm}},
	volume = {11259},
	url = {https://doi.org/10.1007/978-3-030-03341-5\_26},
	doi = {10.1007/978-3-030-03341-5_26},
	booktitle = {Pattern {Recognition} and {Computer} {Vision} - {First} {Chinese} {Conference}, {PRCV} 2018, {Guangzhou}, {China}, {November} 23-26, 2018, {Proceedings}, {Part} {IV}},
	publisher = {Springer},
	author = {Zhao, Quanchao and Ma, Long-Long and Duan, Lijuan},
	editor = {Lai, Jian-Huang and Liu, Cheng-Lin and Chen, Xilin and Zhou, Jie and Tan, Tieniu and Zheng, Nanning and Zha, Hongbin},
	year = {2018},
	pages = {309--321},
}

@inproceedings{shen_exploiting_2017,
	series = {Lecture {Notes} in {Computer} {Science}},
	title = {Exploiting the {Tibetan} {Radicals} in {Recurrent} {Neural} {Network} for {Low}-{Resource} {Language} {Models}},
	volume = {10635},
	url = {https://doi.org/10.1007/978-3-319-70096-0\_28},
	doi = {10.1007/978-3-319-70096-0_28},
	booktitle = {Neural {Information} {Processing} - 24th {International} {Conference}, {ICONIP} 2017, {Guangzhou}, {China}, {November} 14-18, 2017, {Proceedings}, {Part} {II}},
	publisher = {Springer},
	author = {Shen, Tongtong and Wang, Longbiao and Chen, Xie and Khysru, Kuntharrgyal and Dang, Jianwu},
	editor = {Liu, Derong and Xie, Shengli and Li, Yuanqing and Zhao, Dongbin and El-Alfy, El-Sayed M.},
	year = {2017},
	pages = {266--275},
}

@inproceedings{nuo_tibetan_2016,
	series = {Lecture {Notes} in {Computer} {Science}},
	title = {Tibetan {Multi}-word {Expressions} {Identification} {Framework} {Based} on {News} {Corpora}},
	volume = {10102},
	url = {https://doi.org/10.1007/978-3-319-50496-4\_2},
	doi = {10.1007/978-3-319-50496-4_2},
	booktitle = {Natural {Language} {Understanding} and {Intelligent} {Applications} - 5th {CCF} {Conference} on {Natural} {Language} {Processing} and {Chinese} {Computing}, {NLPCC} 2016, and 24th {International} {Conference} on {Computer} {Processing} of {Oriental} {Languages}, {ICCPOL} 2016, {Kunming}, {China}, {December} 2-6, 2016, {Proceedings}},
	publisher = {Springer},
	author = {Nuo, Minghua and Lun, Congjun and Liu, Huidan},
	editor = {Lin, Chin-Yew and Xue, Nianwen and Zhao, Dongyan and Huang, Xuanjing and Feng, Yansong},
	year = {2016},
	pages = {16--26},
}

@inproceedings{ma_online_2015,
	title = {Online handwritten {Tibetan} syllable recognition based on component segmentation method},
	url = {https://doi.org/10.1109/ICDAR.2015.7333723},
	doi = {10.1109/ICDAR.2015.7333723},
	booktitle = {13th {International} {Conference} on {Document} {Analysis} and {Recognition}, {ICDAR} 2015, {Nancy}, {France}, {August} 23-26, 2015},
	publisher = {IEEE Computer Society},
	author = {Ma, Long-Long and Wu, Jian},
	year = {2015},
	pages = {46--50},
}

@inproceedings{ye_tm-tot_2014,
	series = {Communications in {Computer} and {Information} {Science}},
	title = {{TM}-{ToT}: {An} {Effective} {Model} for {Topic} {Mining} from the {Tibetan} {Messages}},
	volume = {496},
	url = {https://doi.org/10.1007/978-3-662-45924-9\_40},
	doi = {10.1007/978-3-662-45924-9_40},
	booktitle = {Natural {Language} {Processing} and {Chinese} {Computing} - {Third} {CCF} {Conference}, {NLPCC} 2014, {Shenzhen}, {China}, {December} 5-9, 2014. {Proceedings}},
	publisher = {Springer},
	author = {Ye, Chengxu and Wen, Wushao and Yang, Ping},
	editor = {Zong, Chengqing and Nie, Jian-Yun and Zhao, Dongyan and Feng, Yansong},
	year = {2014},
	pages = {438--448},
}

@inproceedings{ma_semi-automatic_2013,
	title = {Semi-automatic {Tibetan} {Component} {Annotation} from {Online} {Handwritten} {Tibetan} {Character} {Database} by {Optimizing} {Segmentation} {Hypotheses}},
	url = {https://doi.org/10.1109/ICDAR.2013.271},
	doi = {10.1109/ICDAR.2013.271},
	booktitle = {12th {International} {Conference} on {Document} {Analysis} and {Recognition}, {ICDAR} 2013, {Washington}, {DC}, {USA}, {August} 25-28, 2013},
	publisher = {IEEE Computer Society},
	author = {Ma, Long-Long and Wu, Jian},
	year = {2013},
	pages = {1340--1344},
}

@inproceedings{su_cross-language_2013,
	title = {Cross-{Language} {Sensitive} {Words} {Distribution} {Map}: {A} {Novel} {Recognition}-{Based} {Document} {Understanding} {Method} for {Uighur} and {Tibetan}},
	url = {https://doi.org/10.1109/ICDAR.2013.58},
	doi = {10.1109/ICDAR.2013.58},
	booktitle = {12th {International} {Conference} on {Document} {Analysis} and {Recognition}, {ICDAR} 2013, {Washington}, {DC}, {USA}, {August} 25-28, 2013},
	publisher = {IEEE Computer Society},
	author = {Su, Bing and Ding, Xiaoqing and Peng, Liangrui and Liu, Changsong},
	year = {2013},
	pages = {255--259},
}

@inproceedings{li_study_2013,
	series = {Communications in {Computer} and {Information} {Science}},
	title = {Study on {Tibetan} {Word} {Segmentation} as {Syllable} {Tagging}},
	volume = {400},
	url = {https://doi.org/10.1007/978-3-642-41644-6\_34},
	doi = {10.1007/978-3-642-41644-6_34},
	booktitle = {Natural {Language} {Processing} and {Chinese} {Computing} - {Second} {CCF} {Conference}, {NLPCC} 2013, {Chongqing}, {China}, {November} 15-19, 2013, {Proceedings}},
	publisher = {Springer},
	author = {Li, Yachao and Yu, Hongzhi},
	editor = {Zhou, Guodong and Li, Juanzi and Zhao, Dongyan and Feng, Yansong},
	year = {2013},
	pages = {363--369},
}

@inproceedings{zhao_unsupervised_2012,
	title = {Unsupervised {Tibetan} speech features {Learning} based on {Dynamic} {Bayesian} {Networks}},
	url = {https://ieeexplore.ieee.org/document/6460629/},
	booktitle = {Proceedings of the 21st {International} {Conference} on {Pattern} {Recognition}, {ICPR} 2012, {Tsukuba}, {Japan}, {November} 11-15, 2012},
	publisher = {IEEE Computer Society},
	author = {Zhao, Yue and Xu, Xiaona and Yang, Guosheng},
	year = {2012},
	pages = {2319--2322},
}

@inproceedings{ma_mrg-ohtc_2011,
	title = {{MRG}-{OHTC} {Database} for {Online} {Handwritten} {Tibetan} {Character} {Recognition}},
	url = {https://doi.org/10.1109/ICDAR.2011.50},
	doi = {10.1109/ICDAR.2011.50},
	booktitle = {2011 {International} {Conference} on {Document} {Analysis} and {Recognition}, {ICDAR} 2011, {Beijing}, {China}, {September} 18-21, 2011},
	publisher = {IEEE Computer Society},
	author = {Ma, Long-Long and Liu, Huidan and Wu, Jian},
	year = {2011},
	pages = {207--211},
}

@article{tashi_tibetan_2026,
	title = {Tibetan {Data} {Augmentation} via {GAN}-{Based} {Handwritten} {Text} {Generation}},
	volume = {11},
	url = {https://doi.org/10.1049/cit2.70078},
	doi = {10.1049/CIT2.70078},
	number = {1},
	journal = {CAAI Trans. Intell. Technol.},
	author = {Tashi, Dorje and Chen, Bingtian and Sheng, Tianying and Yu, Yongbin and Wang, Xiangxiang and Zhang, Jin and Yeshi, Lobsang and Dongrub, Rinchen and Tsering, Thupten and Tashi, Nyima},
	year = {2026},
	pages = {55--65},
}

@article{zhang_cro-mtvits_2026,
	title = {Cro-{MTVITS}: {An} end-to-end cross-lingual speech synthesis model for {Mandarin} and multi-dialect {Tibetan} based on {VITS}},
	volume = {100},
	url = {https://doi.org/10.1016/j.csl.2026.101956},
	doi = {10.1016/J.CSL.2026.101956},
	journal = {Comput. Speech Lang.},
	author = {Zhang, Weizhao and Wang, Mengjuan and Li, Junzhi and Yang, Hongwu},
	year = {2026},
	pages = {101956},
}

@article{chen_dataset_2026,
	title = {Dataset {Used} for {Reinforcement} {Learning} {Tasks} of {Tibetan} {Large} {Models}},
	volume = {8},
	url = {https://doi.org/10.3724/2096-7004.di.2025.0044},
	doi = {10.3724/2096-7004.DI.2025.0044},
	number = {2},
	journal = {Data Intell.},
	author = {Chen, Long and Duojie, Gengga and Sun, Yuan},
	year = {2026},
	pages = {20250044},
}

@article{fan_tibetan-chinese_2026,
	title = {Tibetan-{Chinese} {Speech}-to-{Speech} {Translation} {Based} on {Large} {Speech} {Models}},
	volume = {8},
	url = {https://doi.org/10.3724/2096-7004.di.2025.0137},
	doi = {10.3724/2096-7004.DI.2025.0137},
	number = {2},
	journal = {Data Intell.},
	author = {Fan, Hetong and Zhao, Yue and Yu, Jing and Gong, Zairan and Xu, Xiaona and Li, Haizhou},
	year = {2026},
	pages = {20250137},
}

@article{jie_tibocr-bench_2026,
	title = {{TibOCR}-{Bench}: {A} {Comprehensive} {Benchmark} and {Training} {Pipeline} for {Tibetan} {Multimodal} {OCR}},
	volume = {8},
	url = {https://doi.org/10.3724/2096-7004.di.2025.0140},
	doi = {10.3724/2096-7004.DI.2025.0140},
	number = {2},
	journal = {Data Intell.},
	author = {Jie, Lama and Cairang, Manla and Khysru, Kuntharrgyal and Rang, Hua Guo Cai and Jiahui, Jiao and Yingkai, Yue and Qian, Tan},
	year = {2026},
	pages = {20250140},
}

@article{ren_research_2026,
	title = {Research on the {Quality} {Evaluation} of {Amdo} {Tibetan}-{Chinese} {Speech} {Translation} {Dataset}},
	volume = {8},
	url = {https://doi.org/10.3724/2096-7004.di.2025.0172},
	doi = {10.3724/2096-7004.DI.2025.0172},
	number = {2},
	journal = {Data Intell.},
	author = {Ren, Zengzhuoma and Zhu, Liping and Yang, Xiuying and Zhao, Xiaobing},
	year = {2026},
	pages = {20250172},
}

@article{wu_efficient_2026,
	title = {Efficient and {Natural} {Tibetan} {Speech} {Synthesis} via {Gaussian} {Noise}-{Improved} {Monotonic} {Alignment} {Search} and {Lightweight} {iSTFT} {Multiband} {Decoder}},
	volume = {8},
	url = {https://doi.org/10.3724/2096-7004.di.2025.0128},
	doi = {10.3724/2096-7004.DI.2025.0128},
	number = {2},
	journal = {Data Intell.},
	author = {Wu, Shiqi and Zhao, Yue and Yu, Jing and Xu, Xiaona and Li, Haizhou},
	year = {2026},
	pages = {20250128},
}

@article{zhuoma_multi-granularity_2026,
	title = {A multi-granularity word segment evaluation set construction method for {Tibetan} language},
	volume = {181},
	url = {https://doi.org/10.1016/j.engappai.2026.115301},
	doi = {10.1016/J.ENGAPPAI.2026.115301},
	journal = {Eng. Appl. Artif. Intell.},
	author = {Zhuoma, Cairang and Zhijie, Cai},
	year = {2026},
	pages = {115301},
}

@article{moore_evaluating_2026,
	title = {Evaluating {Tibetan} {ASR} {With} {Segmented} {Word} {Error} {Rate}: {Beyond} {Character}-{Level} {Metrics}},
	volume = {14},
	url = {https://doi.org/10.1109/ACCESS.2026.3709206},
	doi = {10.1109/ACCESS.2026.3709206},
	journal = {IEEE Access},
	author = {Moore, Jacob and Li, Sheng and Lauren, Paula},
	year = {2026},
	pages = {101790--101805},
}

@article{zhou_tibhcr_2026,
	title = {{TibHCR}: {A} {Large}-{Scale} {Dataset} for {Offline} {Handwritten} {Tibetan} {Character} {Recognition}},
	volume = {109},
	url = {https://doi.org/10.1587/transinf.2025edp7098},
	doi = {10.1587/TRANSINF.2025EDP7098},
	number = {1},
	journal = {IEICE Trans. Inf. Syst.},
	author = {Zhou, Maoke and Qi, Xiaoke and Bao, Wei and Zhao, Xiaobing},
	year = {2026},
	pages = {193--205},
}

@article{lhundrub_tibstr-former_2026,
	title = {{TibSTR}-{Former}: {Multi}-scale feature integration and attention mechanisms for {Tibetan} scene text recognition},
	volume = {120},
	url = {https://doi.org/10.1016/j.jvcir.2026.104907},
	doi = {10.1016/J.JVCIR.2026.104907},
	journal = {J. Vis. Commun. Image Represent.},
	author = {Lhundrub, Gyaltsan and Zhou, Yong and Long, Mohan and Ke, Wuping and Chen, Geng and Wang, Jianshu and Zheng, Desheng},
	year = {2026},
	pages = {104907},
}

@article{zhang_tibetan_2026,
	title = {Tibetan {Sentiment} {Analysis} {Based} on {Cross}-{Language} {Sentence} {Vector} {Contrastive} {Distillation} {Learning}},
	volume = {20},
	url = {https://doi.org/10.3837/tiis.2026.06.001},
	doi = {10.3837/TIIS.2026.06.001},
	number = {6},
	journal = {KSII Trans. Internet Inf. Syst.},
	author = {Zhang, Rui and Liu, Hongkai and Li, Gege and Wang, Zhaoyang and Ye, Zhonglin and Liu, Zhen and Jia, Cairang},
	year = {2026},
	pages = {2851--2873},
}

@article{wang_quality_2026,
	title = {Quality assessment of {Tibetan}-{Chinese} poetry translation: {Integrating} automated metrics and qualitative insights through a cross-system comparison of dedicated {NMT} engines and a prompted {LLM}},
	volume = {60},
	url = {https://doi.org/10.1007/s10579-026-09918-8},
	doi = {10.1007/S10579-026-09918-8},
	number = {2},
	journal = {Lang. Resour. Evaluation},
	author = {Wang, Qiufen and Wang, Ying and Amini, Mansour and Xian, Ming and Fu, Ziqiang},
	year = {2026},
	pages = {42},
}

@article{sha_tibetan-llama_2025,
	title = {Tibetan-{LLaMA} 2: {Large} {Language} {Model} for {Tibetan}},
	volume = {24},
	url = {https://doi.org/10.1145/3776748},
	doi = {10.1145/3776748},
	number = {12},
	journal = {ACM Trans. Asian Low Resour. Lang. Inf. Process.},
	author = {Sha, Jiu and Zhu, Mengxiao and Feng, Chong and Ci, Jizhuoma},
	year = {2025},
	pages = {141:1--141:19},
}

@article{tang_tibetan_2025,
	title = {Tibetan {Text} {Classification} based on {Prompt} {Learning} and {Ensemble} {Learning}},
	volume = {24},
	url = {https://doi.org/10.1145/3711827},
	doi = {10.1145/3711827},
	number = {2},
	journal = {ACM Trans. Asian Low Resour. Lang. Inf. Process.},
	author = {Tang, Chao and Tan, Zelin and Zhao, Xiaobing},
	year = {2025},
	pages = {19:1--19:17},
}

@article{zhuang_tibetan_2025,
	title = {Tibetan {Question} {Generation} {Based} on {Key} {Sentence} and {Knowledge} {Graph}},
	volume = {24},
	url = {https://doi.org/10.1145/3725531},
	doi = {10.1145/3725531},
	number = {5},
	journal = {ACM Trans. Asian Low Resour. Lang. Inf. Process.},
	author = {Zhuang, Yan and Sun, Yuan and Li, Yijie and Liu, Sisi and Zhao, Xiaobing},
	year = {2025},
	pages = {42:1--42:16},
}

@article{ke_align_2025,
	title = {Align, enhance and read: {Scene} {Tibetan} text recognition with cross-sequence reasoning},
	volume = {169},
	url = {https://doi.org/10.1016/j.asoc.2024.112548},
	doi = {10.1016/J.ASOC.2024.112548},
	journal = {Appl. Soft Comput.},
	author = {Ke, Wenjun and Liu, Yutian and Yang, Xiaokang and Wei, Jianguo and Hou, Qingzhi},
	year = {2025},
	pages = {112548},
}

@article{zhang_tibetan_2025-1,
	title = {Tibetan {Medical} {Named} {Entity} {Recognition} {Based} on {Syllable}-{Word}-{Sentence} {Embedding} {Transformer}},
	volume = {10},
	url = {https://doi.org/10.1049/cit2.70029},
	doi = {10.1049/CIT2.70029},
	number = {4},
	journal = {CAAI Trans. Intell. Technol.},
	author = {Zhang, Jin and Zhang, Ziyue and Yeshi, Lobsang and Tashi, Dorje and Wang, Xiangshi and Cai, Yuqing and Yu, Yongbin and Wang, Xiangxiang and Tashi, Nyima and Luosang, Gadeng},
	year = {2025},
	pages = {1148--1158},
}

@article{zhang_tibetan_2025,
	title = {Tibetan {Few}-{Shot} {Learning} {Model} {With} {Deep} {Contextualised} {Two}-{Level} {Word} {Embeddings}},
	volume = {10},
	url = {https://doi.org/10.1049/cit2.70047},
	doi = {10.1049/CIT2.70047},
	number = {5},
	journal = {CAAI Trans. Intell. Technol.},
	author = {Zhang, Ziyue and Yu, Yongbin and Wang, Xiangxiang and Feng, Xiao and Li, Yuze and Shen, Jiarun and Tashi, Dorje and Zhang, Jin and Yeshi, Lobsang and Li, Lei and Tashi, Nyima and Cai, Jingye},
	year = {2025},
	pages = {1394--1410},
}

@article{lv_t-llama_2025,
	title = {T-{LLaMA}: a {Tibetan} large language model based on {LLaMA2}},
	volume = {11},
	url = {https://doi.org/10.1007/s40747-024-01641-7},
	doi = {10.1007/S40747-024-01641-7},
	number = {1},
	journal = {Complex Intell. Syst.},
	author = {Lv, Hui and Pu, Chi and Duo, La and Li, Yan and Zhou, Qingguo and Shen, Jun},
	year = {2025},
}

@article{zhuang_tifd_2025,
	title = {{TIFD}: {Tibetan} {Instruction}-{Following} {Dataset} for {Large} {Language} {Models} {Supervised} {Fine}-{Tuning}},
	volume = {7},
	url = {https://doi.org/10.3724/2096-7004.di.2024.0010},
	doi = {10.3724/2096-7004.DI.2024.0010},
	number = {3},
	journal = {Data Intell.},
	author = {Zhuang, Wenhao and Cairen, Dawa and Sun, Yuan},
	year = {2025},
	pages = {776--785},
}

@article{khysru_research_2025,
	title = {Research on speech synthesis technology based on {Tibetan} rhythmic features},
	volume = {277},
	url = {https://doi.org/10.1016/j.eswa.2025.127181},
	doi = {10.1016/J.ESWA.2025.127181},
	journal = {Expert Syst. Appl.},
	author = {Khysru, Kuntharrgyal and {Yangzom} and Tang, Wenjie and Wei, Jianguo},
	year = {2025},
	pages = {127181},
}

@article{gyatso_improved_2025,
	title = {An {Improved} {Byte} {Pair} {Encoding} {Method} for {Tibetan}},
	volume = {29},
	url = {https://doi.org/10.20965/jaciii.2025.p1273},
	doi = {10.20965/JACIII.2025.P1273},
	number = {6},
	journal = {J. Adv. Comput. Intell. Intell. Informatics},
	author = {Gyatso, Kalzang and Tshering, Sonam and Norbu, Tashi and Tashi, Nyima and Xiao, Tong and Zhu, Jingbo and Tashi, Garma and Luosang, Gadeng},
	year = {2025},
	pages = {1273--1282},
}

@article{kong_rpepl_2024,
	title = {{RPEPL}: {Tibetan} {Sentiment} {Analysis} {Based} on {Relative} {Position} {Encoding} and {Prompt} {Learning}},
	volume = {23},
	url = {https://doi.org/10.1145/3698575},
	doi = {10.1145/3698575},
	number = {12},
	journal = {ACM Trans. Asian Low Resour. Lang. Inf. Process.},
	author = {Kong, Chunwei and Lv, Xueqiang and Zhang, Le and Zhao, Haixing and Cai, Zangtai and Chen, Yuzhong},
	year = {2024},
	pages = {167:1--167:18},
}

@article{luo_cleansing_2024,
	title = {Cleansing {Jewel}: {A} {Neural} {Spelling} {Correction} {Model} {Built} {On} {Google} {OCR}-ed {Tibetan} {Manuscripts}},
	volume = {23},
	url = {https://doi.org/10.1145/3654811},
	doi = {10.1145/3654811},
	number = {5},
	journal = {ACM Trans. Asian Low Resour. Lang. Inf. Process.},
	author = {Luo, Queenie and Chuang, Yung-Sung},
	year = {2024},
	pages = {73},
}

@article{lv_improved_2024,
	title = {Improved {Tibetan} {Word} {Vectors} {Models} {Based} on {Position} {Information} {Fusion}},
	volume = {23},
	url = {https://doi.org/10.1145/3681787},
	doi = {10.1145/3681787},
	number = {11},
	journal = {ACM Trans. Asian Low Resour. Lang. Inf. Process.},
	author = {Lv, Hui and Lv, Hao and Yang, Liu and Shen, Jun and Duo, La and Li, Yan and Zhou, Qingguo and Yong, Binbin},
	year = {2024},
	pages = {153:1--153:21},
}

@article{yu_novel_2024,
	title = {A novel {Chinese}-{Tibetan} mixed-language rumor detector with multi-extractor representations},
	volume = {87},
	url = {https://doi.org/10.1016/j.csl.2024.101625},
	doi = {10.1016/J.CSL.2024.101625},
	journal = {Comput. Speech Lang.},
	author = {Yu, Lisu and Li, Fei and Yu, Lixin and Li, Wei and Dong, Zhicheng and Cai, Donghong and Wang, Zhen},
	year = {2024},
	pages = {101625},
}

@article{dan_tibetanqa20_2024,
	title = {{TibetanQA2}.0: {Dataset} with {Unanswerable} {Questions} for {Tibetan} {Machine} {Reading} {Comprehension}},
	volume = {6},
	url = {https://doi.org/10.3724/2096-7004.di.2024.0029},
	doi = {10.3724/2096-7004.DI.2024.0029},
	number = {4},
	journal = {Data Intell.},
	author = {Dan, Zhengcuo and Sun, Yuan},
	year = {2024},
	pages = {1158--1167},
}

@article{tashi_construction_2024,
	title = {Construction of a {Tibetan} {Handwriting} {Khyug}-yig {Dataset}},
	volume = {6},
	url = {https://doi.org/10.3724/2096-7004.di.2024.0048},
	doi = {10.3724/2096-7004.DI.2024.0048},
	number = {3},
	journal = {Data Intell.},
	author = {Tashi, Dorje and Sheng, Tianying and Chen, Bingtian and Duojie, Renzeng and Dongrub, Rinchen and Yu, Yongbin and Wang, Xiangxiang and Tashi, Nyima},
	year = {2024},
	pages = {870--887},
}

@article{liu_exploring_2024,
	title = {Exploring task-diverse meta-learning on {Tibetan} multi-dialect speech recognition},
	volume = {2024},
	url = {https://doi.org/10.1186/s13636-024-00361-7},
	doi = {10.1186/S13636-024-00361-7},
	number = {1},
	journal = {EURASIP J. Audio Speech Music. Process.},
	author = {Liu, Yigang and Zhao, Yue and Xu, Xiaona and Xu, Liang and Zhang, Xubei and Ji, Qiang},
	year = {2024},
	pages = {37},
}

@article{bao_toward_2024,
	title = {Toward {Better} {Generalization} of {Cross}-{Domain} {Few}-{Shot} {Classification} in {Tibetan} {Character} {With} {Contrastive} {Learning} and {Meta} {Fine}-{Tuning}},
	volume = {12},
	url = {https://doi.org/10.1109/ACCESS.2024.3459933},
	doi = {10.1109/ACCESS.2024.3459933},
	journal = {IEEE Access},
	author = {Bao, Xun and Wang, Weilan and Wang, Xiaojuan and Zhao, Guanzhong and Li, Huarui and Liu, Meiling},
	year = {2024},
	pages = {134439--134452},
}

@article{zhao_incremental_2024,
	title = {Incremental {Recognition} of {Multi}-{Style} {Tibetan} {Character} {Based} on {Transfer} {Learning}},
	volume = {12},
	url = {https://doi.org/10.1109/ACCESS.2024.3381039},
	doi = {10.1109/ACCESS.2024.3381039},
	journal = {IEEE Access},
	author = {Zhao, Guanzhong and Wang, Weilan and Wang, Xiaojuan and Bao, Xun and Li, Huarui and Liu, Meiling},
	year = {2024},
	pages = {44190--44206},
}

@article{liang_tibetan-bert-wwm_2024,
	title = {Tibetan-{BERT}-wwm: {A} {Tibetan} {Pretrained} {Model} {With} {Whole} {Word} {Masking} for {Text} {Classification}},
	volume = {11},
	url = {https://doi.org/10.1109/TCSS.2024.3374633},
	doi = {10.1109/TCSS.2024.3374633},
	number = {5},
	journal = {IEEE Trans. Comput. Soc. Syst.},
	author = {Liang, Yatao and Lv, Hui and Li, Yan and Duo, La and Liu, Chuanyi and Zhou, Qingguo},
	year = {2024},
	pages = {6268--6277},
}

@article{an_prompt-based_2023,
	title = {Prompt-based for {Low}-{Resource} {Tibetan} {Text} {Classification}},
	volume = {22},
	url = {https://doi.org/10.1145/3603168},
	doi = {10.1145/3603168},
	number = {8},
	journal = {ACM Trans. Asian Low Resour. Lang. Inf. Process.},
	author = {An, Bo},
	year = {2023},
	pages = {207:1--207:13},
}

@article{zhang_improving_2023,
	title = {Improving {Sequence}-to-sequence {Tibetan} {Speech} {Synthesis} with {Prosodic} {Information}},
	volume = {22},
	url = {https://doi.org/10.1145/3616012},
	doi = {10.1145/3616012},
	number = {9},
	journal = {ACM Trans. Asian Low Resour. Lang. Inf. Process.},
	author = {Zhang, Weizhao and Yang, Hongwu},
	year = {2023},
	pages = {225:1--225:13},
}

@article{zhang_hutnet_2023,
	title = {{HUTNet}: {An} {Efficient} {Convolutional} {Neural} {Network} for {Handwritten} {Uchen} {Tibetan} {Character} {Recognition}},
	volume = {11},
	url = {https://doi.org/10.1089/big.2021.0333},
	doi = {10.1089/BIG.2021.0333},
	number = {5},
	journal = {Big Data},
	author = {Zhang, Guowei and Wang, Weilan and Zhang, Ce and Zhao, Penghai and Zhang, Mingkai},
	year = {2023},
	pages = {387--398},
}

@article{li_sentence_2022,
	title = {Sentence {Boundary} {Disambiguation} for {Tibetan} {Based} on {Attention} {Mechanism} at the {Syllable} {Level}},
	volume = {21},
	url = {https://doi.org/10.1145/3527663},
	doi = {10.1145/3527663},
	number = {6},
	journal = {ACM Trans. Asian Low Resour. Lang. Inf. Process.},
	author = {Li, Fenfang and Lv, Hui and Duo, La and Yong, Binbin and Zhou, Qingguo},
	year = {2022},
	pages = {1--18},
}

@article{li_character-based_2022,
	title = {Character-based {Joint} {Word} {Segmentation} and {Part}-of-{Speech} {Tagging} for {Tibetan} {Based} on {Deep} {Learning}},
	volume = {21},
	url = {https://doi.org/10.1145/3511600},
	doi = {10.1145/3511600},
	number = {5},
	journal = {ACM Trans. Asian Low Resour. Lang. Inf. Process.},
	author = {Li, Yan and Li, Xiaomin and Wang, Yiru and Lv, Hui and Li, Fenfang and Duo, La},
	year = {2022},
	pages = {95:1--95:15},
}

@article{qin_improving_2022,
	title = {Improving low-resource {Tibetan} end-to-end {ASR} by multilingual and multilevel unit modeling},
	volume = {2022},
	url = {https://doi.org/10.1186/s13636-021-00233-4},
	doi = {10.1186/S13636-021-00233-4},
	number = {1},
	journal = {EURASIP J. Audio Speech Music. Process.},
	author = {Qin, Siqing and Wang, Longbiao and Li, Sheng and Dang, Jianwu and Pan, Lixin},
	year = {2022},
	pages = {2},
}

@article{li_character_2022,
	title = {Character recognition of {Tibetan} {Historical} document in {Uchen} font: {Dataset} and bench mark},
	volume = {22},
	url = {https://doi.org/10.3233/JCM-226167},
	doi = {10.3233/JCM-226167},
	number = {5},
	journal = {J. Comput. Methods Sci. Eng.},
	author = {Li, Zhenjiang and Wang, Weilan and Wang, Yiqun and Zhang, Qianxue},
	year = {2022},
	pages = {1779--1794},
}

@article{gan_tibetan-dependent_2022,
	title = {A {Tibetan}-dependent speaker recognition method based on deep learning},
	volume = {81},
	url = {https://doi.org/10.1007/s11042-022-12540-9},
	doi = {10.1007/S11042-022-12540-9},
	number = {21},
	journal = {Multim. Tools Appl.},
	author = {Gan, Zhenye and Yu, Yue and Luo, Min},
	year = {2022},
	pages = {30821--30840},
}

@article{an_neural_2021,
	title = {Neural {Dependency} {Parser} for {Tibetan} {Sentences}},
	volume = {20},
	url = {https://doi.org/10.1145/3429456},
	doi = {10.1145/3429456},
	number = {2},
	journal = {ACM Trans. Asian Low Resour. Lang. Inf. Process.},
	author = {An, Bo and Long, Congjun},
	year = {2021},
	pages = {30:1--30:16},
}

@article{krishna_applying_2021,
	title = {Applying {Text} {Analytics} to the {Mind}-section {Literature} of the {Tibetan} {Tradition} of the {Great} {Perfection}},
	volume = {20},
	url = {https://doi.org/10.1145/3392047},
	doi = {10.1145/3392047},
	number = {2},
	journal = {ACM Trans. Asian Low Resour. Lang. Inf. Process.},
	author = {Krishna, Ravi and Mu, Norman and Keutzer, Kurt},
	year = {2021},
	pages = {21:1--21:32},
}

@article{li_finding_2021,
	title = {Finding {Better} {Subwords} for {Tibetan} {Neural} {Machine} {Translation}},
	volume = {20},
	url = {https://doi.org/10.1145/3448216},
	doi = {10.1145/3448216},
	number = {2},
	journal = {ACM Trans. Asian Low Resour. Lang. Inf. Process.},
	author = {Li, Yachao and Jiang, Jing and Yangji, Jia and Ma, Ning},
	year = {2021},
	pages = {24:1--24:11},
}

@article{long_recognition_2021,
	title = {Recognition of {Tibetan} {Maximal}-length {Noun} {Phrases} {Based} on {Syntax} {Tree}},
	volume = {20},
	url = {https://doi.org/10.1145/3423324},
	doi = {10.1145/3423324},
	number = {2},
	journal = {ACM Trans. Asian Low Resour. Lang. Inf. Process.},
	author = {Long, Congjun and Zhou, Xuewen and Zhou, Maoke},
	year = {2021},
	pages = {25:1--25:13},
}

@article{meelen_optimisation_2021,
	title = {Optimisation of the {Largest} {Annotated} {Tibetan} {Corpus} {Combining} {Rule}-based, {Memory}-based, and {Deep}-learning {Methods}},
	volume = {20},
	url = {https://doi.org/10.1145/3409488},
	doi = {10.1145/3409488},
	number = {1},
	journal = {ACM Trans. Asian Low Resour. Lang. Inf. Process.},
	author = {Meelen, Marieke and Roux, Élie and Hill, Nathan},
	year = {2021},
	pages = {7:1--7:11},
}

@article{shi_multi-level_2021,
	title = {Multi-level {Chunk}-based {Constituent}-to-{Dependency} {Treebank} {Transformation} for {Tibetan} {Dependency} {Parsing}},
	volume = {20},
	url = {https://doi.org/10.1145/3424247},
	doi = {10.1145/3424247},
	number = {2},
	journal = {ACM Trans. Asian Low Resour. Lang. Inf. Process.},
	author = {Shi, Shumin and Luo, Dan and Wu, Xing and Long, Congjun and Huang, Heyan},
	year = {2021},
	pages = {26:1--26:12},
}

@article{sun_joint_2021,
	title = {A {Joint} {Model} for {Representation} {Learning} of {Tibetan} {Knowledge} {Graph} {Based} on {Encyclopedia}},
	volume = {20},
	url = {https://doi.org/10.1145/3447248},
	doi = {10.1145/3447248},
	number = {2},
	journal = {ACM Trans. Asian Low Resour. Lang. Inf. Process.},
	author = {Sun, Yuan and Chen, Andong and Chen, Chaofan and Xia, Tianci and Zhao, Xiaobing},
	year = {2021},
	pages = {27:1--27:17},
}

@article{trinley_taming_2021,
	title = {Taming the {Wild} {Etext}: {Managing}, {Annotating}, and {Sharing} {Tibetan} {Corpora} in {Open} {Spaces}},
	volume = {20},
	url = {https://doi.org/10.1145/3418060},
	doi = {10.1145/3418060},
	number = {2},
	journal = {ACM Trans. Asian Low Resour. Lang. Inf. Process.},
	author = {Trinley, Ngawang and {Tenzin} and Schmidt, Dirk and Hildt, Helios and Kaldan, Tenzin},
	year = {2021},
	pages = {22:1--22:23},
}

@article{xu_end--end_2021,
	title = {End-to-{End} {Speech} {Synthesis} for {Tibetan} {Multidialect}},
	volume = {2021},
	url = {https://doi.org/10.1155/2021/6682871},
	doi = {10.1155/2021/6682871},
	journal = {Complex.},
	author = {Xu, Xiaona and Yang, Li and Zhao, Yue and Wang, Hui},
	year = {2021},
	pages = {6682871:1--6682871:8},
}

@article{khysru_tibetan_2021,
	title = {A {Tibetan} {Language} {Model} {That} {Considers} the {Relationship} {Between} {Suffixes} and {Functional} {Words}},
	volume = {28},
	url = {https://doi.org/10.1109/LSP.2021.3058896},
	doi = {10.1109/LSP.2021.3058896},
	journal = {IEEE Signal Process. Lett.},
	author = {Khysru, Kuntharrgyal and Jin, Di and Huang, Yuxiao and Feng, Hui and Dang, Jianwu},
	year = {2021},
	pages = {459--463},
}

@article{schmidt_grading_2020,
	title = {Grading {Tibetan} {Children}'s {Literature}: {A} {Test} {Case} {Using} the {NLP} {Readability} {Tool} ``{Dakje}''},
	volume = {19},
	url = {https://doi.org/10.1145/3392046},
	doi = {10.1145/3392046},
	number = {6},
	journal = {ACM Trans. Asian Low Resour. Lang. Inf. Process.},
	author = {Schmidt, Dirk},
	year = {2020},
	pages = {75:1--75:19},
}

@article{wang_multitask_2020,
	title = {Multitask {Learning} with {Local} {Attention} for {Tibetan} {Speech} {Recognition}},
	volume = {2020},
	url = {https://doi.org/10.1155/2020/8894566},
	doi = {10.1155/2020/8894566},
	journal = {Complex.},
	author = {Wang, Hui and Gao, Fei and Zhao, Yue and Yang, Li and Yue, Jianjian and Ma, Huilin},
	year = {2020},
	pages = {8894566:1--8894566:10},
}

@article{ma_segmentation_2020,
	title = {Segmentation and {Recognition} for {Historical} {Tibetan} {Document} {Images}},
	volume = {8},
	url = {https://doi.org/10.1109/ACCESS.2020.2975023},
	doi = {10.1109/ACCESS.2020.2975023},
	journal = {IEEE Access},
	author = {Ma, Long-Long and Long, Congjun and Duan, Lijuan and Zhang, Xiqun and Li, Yanxing and Zhao, Quanchao},
	year = {2020},
	pages = {52641--52651},
}

@article{zhu_end--end_2020,
	title = {End-to-{End} {Amdo}-{Tibetan} {Speech} {Recognition} {Based} on {Knowledge} {Transfer}},
	volume = {8},
	url = {https://doi.org/10.1109/ACCESS.2020.3023783},
	doi = {10.1109/ACCESS.2020.3023783},
	journal = {IEEE Access},
	author = {Zhu, Xiaojun and Huang, Heming},
	year = {2020},
	pages = {170991--171000},
}

@article{wang_transliteration_2020,
	title = {Transliteration recognition of {Tibetan} person name based on {Tibetan} cultural knowledge},
	volume = {22},
	url = {https://doi.org/10.1504/IJCSE.2020.107352},
	doi = {10.1504/IJCSE.2020.107352},
	number = {2/3},
	journal = {Int. J. Comput. Sci. Eng.},
	author = {Wang, Zhijuan and Fang, Wenguang and Zhao, Xiaobing and Song, Wei and Feng, Yinghui and Chang, Yining},
	year = {2020},
	pages = {305--312},
}

@article{zhao_open_2020,
	title = {An open speech resource for {Tibetan} multi-dialect and multitask recognition},
	volume = {22},
	url = {https://doi.org/10.1504/IJCSE.2020.107351},
	doi = {10.1504/IJCSE.2020.107351},
	number = {2/3},
	journal = {Int. J. Comput. Sci. Eng.},
	author = {Zhao, Yue and Xu, Xiaona and Yue, Jianjian and Song, Wei and Li, Xiali and Wu, Licheng and Ji, Qiang},
	year = {2020},
	pages = {297--304},
}

@article{xu_research_2019,
	title = {Research on {Tibetan} hot words, sensitive words tracking and public opinion classification},
	volume = {22},
	url = {https://doi.org/10.1007/s10586-017-1026-x},
	doi = {10.1007/S10586-017-1026-X},
	number = {Supplement},
	journal = {Clust. Comput.},
	author = {Xu, Guixian and Wang, Changzhi and Yao, Haishen and Qi, Qi},
	year = {2019},
	pages = {9977--9990},
}

@article{xu_public_2019,
	title = {Public opinion classification and text alignment based on {Chinese} and {Tibetan} corpus},
	volume = {22},
	url = {https://doi.org/10.1007/s10586-017-1267-8},
	doi = {10.1007/S10586-017-1267-8},
	number = {Supplement},
	journal = {Clust. Comput.},
	author = {Xu, Guixian and Yao, Haishen and Wu, Dongming and Li, Yuan and Ouyang, Deguang and Chen, Gaofeng},
	year = {2019},
	pages = {10263--10274},
}

@article{khysru_morphological_2019,
	title = {Morphological {Verb}-{Aware} {Tibetan} {Language} {Model}},
	volume = {7},
	url = {https://doi.org/10.1109/ACCESS.2019.2919328},
	doi = {10.1109/ACCESS.2019.2919328},
	journal = {IEEE Access},
	author = {Khysru, Kuntharrgyal and Jin, Di and Dang, Jianwu},
	year = {2019},
	pages = {72896--72904},
}

@article{sun_qugan_2019,
	title = {{QuGAN}: {Quasi} {Generative} {Adversarial} {Network} for {Tibetan} {Question} {Answering} {Corpus} {Generation}},
	volume = {7},
	url = {https://doi.org/10.1109/ACCESS.2019.2934581},
	doi = {10.1109/ACCESS.2019.2934581},
	journal = {IEEE Access},
	author = {Sun, Yuan and Chen, Chaofan and Xia, Tianci and Zhao, Xiaobing},
	year = {2019},
	pages = {116247--116255},
}

@article{sun_improved_2019,
	title = {Improved {Distant} {Supervised} {Model} in {Tibetan} {Relation} {Extraction} {Using} {ELMo} and {Attention}},
	volume = {7},
	url = {https://doi.org/10.1109/ACCESS.2019.2955977},
	doi = {10.1109/ACCESS.2019.2955977},
	journal = {IEEE Access},
	author = {Sun, Yuan and Wang, Like and Chen, Chaofan and Xia, Tianci and Zhao, Xiaobing},
	year = {2019},
	pages = {173054--173062},
}

@article{sun_hybrid_2019,
	title = {A {Hybrid} {Network} {Model} for {Tibetan} {Question} {Answering}},
	volume = {7},
	url = {https://doi.org/10.1109/ACCESS.2019.2911320},
	doi = {10.1109/ACCESS.2019.2911320},
	journal = {IEEE Access},
	author = {Sun, Yuan and Xia, Tianci},
	year = {2019},
	pages = {52769--52777},
}

@article{zhang_deep_2019,
	title = {Deep {Learning} for {Mandarin}-{Tibetan} {Cross}-{Lingual} {Speech} {Synthesis}},
	volume = {7},
	url = {https://doi.org/10.1109/ACCESS.2019.2954342},
	doi = {10.1109/ACCESS.2019.2954342},
	journal = {IEEE Access},
	author = {Zhang, Weizhao and Yang, Hongwu and Bu, Xiaolong and Wang, Lili},
	year = {2019},
	pages = {167884--167894},
}

@article{zhao_lhasa-tibetan_2019,
	title = {Lhasa-{Tibetan} {Speech} {Synthesis} {Using} {End}-to-{End} {Model}},
	volume = {7},
	url = {https://doi.org/10.1109/ACCESS.2019.2940125},
	doi = {10.1109/ACCESS.2019.2940125},
	journal = {IEEE Access},
	author = {Zhao, Yue and Hu, Panhua and Xu, Xiaona and Wu, Licheng and Li, Xiali},
	year = {2019},
	pages = {140305--140311},
}

@article{zhao_end--end-based_2019,
	title = {End-to-{End}-{Based} {Tibetan} {Multitask} {Speech} {Recognition}},
	volume = {7},
	url = {https://doi.org/10.1109/ACCESS.2019.2952406},
	doi = {10.1109/ACCESS.2019.2952406},
	journal = {IEEE Access},
	author = {Zhao, Yue and Yue, Jianjian and Xu, Xiaona and Wu, Licheng and Li, Xiali},
	year = {2019},
	pages = {162519--162529},
}

@article{wang_online_2019,
	title = {Online {Tibetan} {Handwriting} {Recognition} for {Large} {Character} {Set} on {New} {Databases}},
	volume = {33},
	url = {https://doi.org/10.1142/S0218001419530033},
	doi = {10.1142/S0218001419530033},
	number = {10},
	journal = {Int. J. Pattern Recognit. Artif. Intell.},
	author = {Wang, Weilan and Li, Zhengjiang and Cai, Zhengqi and Lv, Xiaobao and Zhaxi, Caike and Han, Yuehui},
	year = {2019},
	pages = {1953003:1--1953003:21},
}

@article{qiu_tibetan_2018,
	title = {Tibetan {Weibo} {User} {Group} {Division} {Based} on {Semantic} {Information} in the {Era} of {Big} {Data}},
	volume = {2018},
	url = {https://doi.org/10.1155/2018/5809043},
	doi = {10.1155/2018/5809043},
	journal = {Complex.},
	author = {Qiu, Lirong and Yu, Jia and Li, Jie and Jia, Haoran},
	year = {2018},
	pages = {5809043:1--5809043:11},
}

@article{qiu_advanced_2018,
	title = {Advanced {Sentiment} {Classification} of {Tibetan} {Microblogs} on {Smart} {Campuses} {Based} on {Multi}-{Feature} {Fusion}},
	volume = {6},
	url = {https://doi.org/10.1109/ACCESS.2018.2820163},
	doi = {10.1109/ACCESS.2018.2820163},
	journal = {IEEE Access},
	author = {Qiu, Lirong and Lei, Qiao and Zhang, Zhen},
	year = {2018},
	pages = {17896--17904},
}

@article{xu_efficient_2018,
	title = {Efficient {Sensitive} {Information} {Classification} and {Topic} {Tracking} {Based} on {Tibetan} {Web} {Pages}},
	volume = {6},
	url = {https://doi.org/10.1109/ACCESS.2018.2870122},
	doi = {10.1109/ACCESS.2018.2870122},
	journal = {IEEE Access},
	author = {Xu, Guixian and Yu, Ziheng and Qi, Qi},
	year = {2018},
	pages = {55643--55652},
}

@article{xu_semantic_2017,
	title = {Semantic classification method for network {Tibetan} corpus},
	volume = {20},
	url = {https://doi.org/10.1007/s10586-017-0742-6},
	doi = {10.1007/S10586-017-0742-6},
	number = {1},
	journal = {Clust. Comput.},
	author = {Xu, Guixian and Wang, Chang-Zhi and Wang, Li-Hui and Zhou, Yu-Hong and Li, Wei-Kang and Xu, Hao and Huang, Qing},
	year = {2017},
	pages = {155--165},
}

@article{qiu_tibetan_2017,
	title = {Tibetan {Microblog} {Emotional} {Analysis} {Based} on {Sequential} {Model} in {Online} {Social} {Platforms}},
	volume = {2017},
	url = {https://doi.org/10.1155/2017/5342601},
	doi = {10.1155/2017/5342601},
	journal = {Complex.},
	author = {Qiu, Lirong and Zhang, Huili and Zhang, Zhen and Pu, Qiumei},
	year = {2017},
	pages = {5342601:1--5342601:8},
}

@article{yang_using_2015,
	title = {Using speaker adaptive training to realize {Mandarin}-{Tibetan} cross-lingual speech synthesis},
	volume = {74},
	url = {https://doi.org/10.1007/s11042-014-2117-9},
	doi = {10.1007/S11042-014-2117-9},
	number = {22},
	journal = {Multim. Tools Appl.},
	author = {Yang, Hongwu and Oura, Keiichiro and Wang, Haiyan and Gan, Zhenye and Tokuda, Keiichi},
	year = {2015},
	pages = {9927--9942},
}

@article{ye_chinese-tibetan_2015,
	title = {Chinese-{Tibetan} bilingual clustering based on random walk},
	volume = {158},
	url = {https://doi.org/10.1016/j.neucom.2015.02.003},
	doi = {10.1016/J.NEUCOM.2015.02.003},
	journal = {Neurocomputing},
	author = {Ye, Chengxu and Wen, Wushao and Wang, Chang-Dong},
	year = {2015},
	pages = {32--41},
}

@inproceedings{dasgupta-etal-2012-forward,
	title = "Forward Transliteration of {D}zongkha Text to {B}raille",
	author = "Dasgupta, Tirthankar  and
	Sinha, Manjira  and
	Basu, Anupam",
	editor = "Bali, Kalika  and
	Choudhury, Monojit  and
	Okuno, Yoh",
	booktitle = "Proceedings of the Second Workshop on Advances in Text Input Methods",
	month = dec,
	year = "2012",
	address = "Mumbai, India",
	publisher = "The COLING 2012 Organizing Committee",
	url = "https://aclanthology.org/W12-4809/",
	pages = "97--106"
}

@inproceedings{norbu-etal-2010-dzongkha,
	title = "{D}zongkha Word Segmentation",
	author = "Norbu, Sithar  and
	Choejey, Pema  and
	Dendup, Tenzin  and
	Hussain, Sarmad  and
	Muaz, Ahmed",
	editor = "Hussain, Sarmad  and
	Sornlertlamvanich, Virach  and
	Riza, Hammam",
	booktitle = "Proceedings of the Eighth Workshop on {A}sian Language Resouces",
	month = aug,
	year = "2010",
	address = "Beijing, China",
	publisher = "Coling 2010 Organizing Committee",
	url = "https://aclanthology.org/W10-3213/",
	pages = "95--102"
}

@inproceedings{chungku-etal-2010-building,
	title = "Building {NLP} resources for {D}zongkha: A Tagset and A Tagged Corpus",
	author = "Chungku, Chungku  and
	Rabgay, Jurmey  and
	Faa{\ss}, Gertrud",
	editor = "Hussain, Sarmad  and
	Sornlertlamvanich, Virach  and
	Riza, Hammam",
	booktitle = "Proceedings of the Eighth Workshop on {A}sian Language Resouces",
	month = aug,
	year = "2010",
	address = "Beijing, China",
	publisher = "Coling 2010 Organizing Committee",
	url = "https://aclanthology.org/W10-3214/",
	pages = "103--110"
}

@article{wangchuk_next_2022,
	title = {Next syllables prediction system in {Dzongkha} using long short-term memory},
	volume = {34},
	url = {https://doi.org/10.1016/j.jksuci.2021.01.001},
	doi = {10.1016/J.JKSUCI.2021.01.001},
	number = {6 Part B},
	journal = {J. King Saud Univ. Comput. Inf. Sci.},
	author = {Wangchuk, Karma and Riyamongkol, Panomkhawn and Waranusast, Rattapoom},
	year = {2022},
	pages = {3800--3806},
}

@article{dhungyel_analysing_2017,
	title = {Analysing the {Methods} of {Dzongkha} {Word} {Segmentation}},
	volume = {21},
	url = {https://doi.org/10.1515/acss-2017-0008},
	doi = {10.1515/ACSS-2017-0008},
	number = {1},
	journal = {Appl. Comput. Syst.},
	author = {Dhungyel, Parshu Ram and Grundspenkis, Janis},
	year = {2017},
	pages = {61--65},
}

@article{arun_semantic_2014,
	title = {A semantic parsing approach for {Bhutanese} language of {Dzongkha}},
	volume = {17},
	url = {https://doi.org/10.1007/s10772-013-9218-0},
	doi = {10.1007/S10772-013-9218-0},
	number = {2},
	journal = {Int. J. Speech Technol.},
	author = {Arun, P. V.},
	year = {2014},
	pages = {161--165},
}

@article{thukroo_hybrid_2023,
	title = {A {Hybrid} {Convolutional} {Bi}-{Directional} {Gated} {Recurrent} {Unit} {System} for {Spoken} {Languages} of {JK} and {Ladakhi}},
	volume = {22},
	url = {https://doi.org/10.1142/S0219649223500284},
	doi = {10.1142/S0219649223500284},
	number = {4},
	journal = {J. Inf. Knowl. Manag.},
	author = {Thukroo, Irshad Ahmad and Bashir, Rumaan and Giri, Kaiser J.},
	year = {2023},
	pages = {2350028:1--2350028:23},
}

@inproceedings{liu-best-2025-survey,
	title = "A Survey of {NLP} Progress in {S}ino-{T}ibetan Low-Resource Languages",
	author = "Liu, Shuheng  and
	Best, Michael",
	editor = "Chiruzzo, Luis  and
	Ritter, Alan  and
	Wang, Lu",
	booktitle = "Proceedings of the 2025 Conference of the Nations of the Americas Chapter of the Association for Computational Linguistics: Human Language Technologies (Volume 1: Long Papers)",
	month = apr,
	year = "2025",
	address = "Albuquerque, New Mexico",
	publisher = "Association for Computational Linguistics",
	url = "https://aclanthology.org/2025.naacl-long.396/",
	doi = "10.18653/v1/2025.naacl-long.396",
	pages = "7804--7825",
	ISBN = "979-8-89176-189-6"
}

@article{congjun_recent_2021,
	title = {Recent {Developments} in {Tibetan} {NLP}},
	volume = {20},
	url = {https://doi.org/10.1145/3453692},
	doi = {10.1145/3453692},
	number = {2},
	journal = {ACM Trans. Asian Low Resour. Lang. Inf. Process.},
	author = {Long, Congjun and Hill, Nathan W.},
	year = {2021},
	pages = {19:1--19:3},
}

@inproceedings{alabi-etal-2025-charting,
	title = "Charting the Landscape of {A}frican {NLP}: Mapping Progress and Shaping the Road Ahead",
	author = "Alabi, Jesujoba Oluwadara  and
	Hedderich, Michael A.  and
	Adelani, David Ifeoluwa  and
	Klakow, Dietrich",
	editor = "Christodoulopoulos, Christos  and
	Chakraborty, Tanmoy  and
	Rose, Carolyn  and
	Peng, Violet",
	booktitle = "Proceedings of the 2025 Conference on Empirical Methods in Natural Language Processing",
	month = nov,
	year = "2025",
	address = "Suzhou, China",
	publisher = "Association for Computational Linguistics",
	url = "https://aclanthology.org/2025.emnlp-main.1414/",
	doi = "10.18653/v1/2025.emnlp-main.1414",
	pages = "27807--27841",
	ISBN = "979-8-89176-332-6"
}

@inproceedings{adebara-abdul-mageed-2022-towards,
	title = "Towards Afrocentric {NLP} for {A}frican Languages: Where We Are and Where We Can Go",
	author = "Adebara, Ife  and
	Abdul-Mageed, Muhammad",
	editor = "Muresan, Smaranda  and
	Nakov, Preslav  and
	Villavicencio, Aline",
	booktitle = "Proceedings of the 60th Annual Meeting of the Association for Computational Linguistics (Volume 1: Long Papers)",
	month = may,
	year = "2022",
	address = "Dublin, Ireland",
	publisher = "Association for Computational Linguistics",
	url = "https://aclanthology.org/2022.acl-long.265/",
	doi = "10.18653/v1/2022.acl-long.265",
	pages = "3814--3841"
}

@inproceedings{tonja-etal-2024-nlp,
	title = "{NLP} Progress in Indigenous {L}atin {A}merican Languages",
	author = "Tonja, Atnafu  and
	Balouchzahi, Fazlourrahman  and
	Butt, Sabur  and
	Kolesnikova, Olga  and
	Ceballos, Hector  and
	Gelbukh, Alexander  and
	Solorio, Thamar",
	editor = "Duh, Kevin  and
	Gomez, Helena  and
	Bethard, Steven",
	booktitle = "Proceedings of the 2024 Conference of the North American Chapter of the Association for Computational Linguistics: Human Language Technologies (Volume 1: Long Papers)",
	month = jun,
	year = "2024",
	address = "Mexico City, Mexico",
	publisher = "Association for Computational Linguistics",
	url = "https://aclanthology.org/2024.naacl-long.385/",
	doi = "10.18653/v1/2024.naacl-long.385",
	pages = "6972--6987"
}

@inproceedings{kumar-etal-2026-bhashasutra,
	title = "{B}hasha{S}utra: A Task-Centric Unified Survey of {I}ndian {NLP} Datasets, Corpora, and Resources",
	author = "Kumar, Raghvendra  and
	Raj, Devankar  and
	Saha, Sriparna",
	editor = "Liakata, Maria  and
	Moreira, Viviane P.  and
	Zhang, Jiajun  and
	Jurgens, David",
	booktitle = "Proceedings of the 64th Annual Meeting of the {A}ssociation for {C}omputational {L}inguistics (Volume 1: Long Papers)",
	month = jul,
	year = "2026",
	address = "San Diego, California, United States",
	publisher = "Association for Computational Linguistics",
	url = "https://aclanthology.org/2026.acl-long.551/",
	doi = "10.18653/v1/2026.acl-long.551",
	pages = "11998--12064",
	ISBN = "979-8-89176-390-6"
}

@inproceedings{blaschke-etal-2024-dialect,
	title = "What Do Dialect Speakers Want? {A} Survey of Attitudes Towards Language Technology for {G}erman Dialects",
	author = "Blaschke, Verena  and
	Purschke, Christoph  and
	Schuetze, Hinrich  and
	Plank, Barbara",
	editor = "Ku, Lun-Wei  and
	Martins, Andre  and
	Srikumar, Vivek",
	booktitle = "Proceedings of the 62nd Annual Meeting of the Association for Computational Linguistics (Volume 2: Short Papers)",
	month = aug,
	year = "2024",
	address = "Bangkok, Thailand",
	publisher = "Association for Computational Linguistics",
	url = "https://aclanthology.org/2024.acl-short.74/",
	doi = "10.18653/v1/2024.acl-short.74",
	pages = "823--841"
}

@inproceedings{al-kautsar-etal-2025-indonesians,
	title = "What Do {I}ndonesians Really Need from Language Technology? {A} Nationwide Survey",
	author = "Al Kautsar, Muhammad Dehan  and
	Susanto, Lucky  and
	Wijaya, Derry Tanti  and
	Koto, Fajri",
	editor = "Christodoulopoulos, Christos  and
	Chakraborty, Tanmoy  and
	Rose, Carolyn  and
	Peng, Violet",
	booktitle = "Proceedings of the 2025 Conference on Empirical Methods in Natural Language Processing",
	month = nov,
	year = "2025",
	address = "Suzhou, China",
	publisher = "Association for Computational Linguistics",
	url = "https://aclanthology.org/2025.emnlp-main.367/",
	doi = "10.18653/v1/2025.emnlp-main.367",
	pages = "7220--7245",
	ISBN = "979-8-89176-332-6"
}

@inproceedings{rohatgi-etal-2023-acl,
	title = "The {ACL} {OCL} Corpus: Advancing Open Science in Computational Linguistics",
	author = "Rohatgi, Shaurya  and
	Qin, Yanxia  and
	Aw, Benjamin  and
	Unnithan, Niranjana  and
	Kan, Min-Yen",
	editor = "Bouamor, Houda  and
	Pino, Juan  and
	Bali, Kalika",
	booktitle = "Proceedings of the 2023 Conference on Empirical Methods in Natural Language Processing",
	month = dec,
	year = "2023",
	address = "Singapore",
	publisher = "Association for Computational Linguistics",
	url = "https://aclanthology.org/2023.emnlp-main.640/",
	doi = "10.18653/v1/2023.emnlp-main.640",
	pages = "10348--10361"
}

@inproceedings{joshi-etal-2020-state,
	title = "The State and Fate of Linguistic Diversity and Inclusion in the {NLP} World",
	author = "Joshi, Pratik  and
	Santy, Sebastin  and
	Budhiraja, Amar  and
	Bali, Kalika  and
	Choudhury, Monojit",
	editor = "Jurafsky, Dan  and
	Chai, Joyce  and
	Schluter, Natalie  and
	Tetreault, Joel",
	booktitle = "Proceedings of the 58th Annual Meeting of the Association for Computational Linguistics",
	month = jul,
	year = "2020",
	address = "Online",
	publisher = "Association for Computational Linguistics",
	url = "https://aclanthology.org/2020.acl-main.560/",
	doi = "10.18653/v1/2020.acl-main.560",
	pages = "6282--6293"
}

@inproceedings{ghimire-etal-2026-neptam,
	title = "{N}ep{T}am: A {N}epali-{T}amang Parallel Corpus and Baseline Machine Translation Experiments",
	author = "Ghimire, Rupak Raj  and
	Subedi, Bipesh  and
	Prasain, Balaram  and
	Poudyal, Prakash  and
	Acharya, Praveen  and
	Karki, Nischal  and
	Tiwari, Rupak  and
	Sharma, Rishikesh Kumar  and
	Poudel, Jenny  and
	Bal, Bal Krishna",
	editor = "Piperidis, Stelios  and
	Bel, N{\'u}ria  and
	van den Heuvel, Henk  and
	Ide, Nancy  and
	Krek, Simon  and
	Toral, Antonio",
	booktitle = "Proceedings of the Fifteenth Language Resources and Evaluation Conference",
	month = may,
	year = "2026",
	address = "Palma de Mallorca, Spain",
	publisher = "ELRA Language Resource Association",
	url = "https://aclanthology.org/2026.lrec-1.696/",
	doi = "10.63317/37edei5qcjb3",
	pages = "8850--8861"
}

@inproceedings{acharya-etal-2026-english,
	title = "{E}nglish{--}{N}epali{--}{T}amang: A Trilingual Parallel Corpus and Benchmark for Low-Resource Machine Translation",
	author = "Acharya, Praveen  and
	Ghimire, Rupak Raj  and
	Poudyal, Prakash  and
	Prasain, Balaram  and
	Bal, Bal Krishna",
	editor = "Shterionov, Dimitar  and
	Vanmassenhove, Eva  and
	De Sisto, Mirella  and
	Blain, Fred  and
	Pourmostafa Roshan Sharami, Javad  and
	Lepp, Lisa  and
	Manna, Chiara  and
	Rescigno, Argentina Anna  and
	Karakanta, Alina  and
	Rigouts Terryn, Ayla  and
	Lardelli, Manuel  and
	Resende, Natalia  and
	Murgolo, Elena  and
	Hackenbuchner, Jani{\c{c}}a  and
	Zaretskaya, Anna  and
	Espl{\`a}-Gomis, Miquel  and
	Etchegoyhen, Thierry  and
	Gromann, Dagmar  and
	Bawden, Rachel  and
	Haddow, Barry  and
	Szoc, Sara  and
	Forcada, Mikel  and
	Moniz, Helena",
	booktitle = "Proceedings of the 26th Annual Conference of the {E}uropean Association for Machine Translation (Volume 2)",
	month = jun,
	year = "2026",
	address = "Tilburg, The Netherlands",
	publisher = "European Association for Machine Translation",
	url = "https://aclanthology.org/2026.eamt-2.11/",
	pages = "23--24",
	ISBN = "9789403901404"
}

@inproceedings{chaudhary-etal-2020-efforts,
	title = "Efforts Towards Developing a {T}amang {N}epali Machine Translation System",
	author = "Chaudhary, Binaya Kumar  and
	Bal, Bal Krishna  and
	Baidar, Rasil",
	editor = "Bhattacharyya, Pushpak  and
	Sharma, Dipti Misra  and
	Sangal, Rajeev",
	booktitle = "Proceedings of the 17th International Conference on Natural Language Processing (ICON)",
	month = dec,
	year = "2020",
	address = "Indian Institute of Technology Patna, Patna, India",
	publisher = "NLP Association of India (NLPAI)",
	url = "https://aclanthology.org/2020.icon-main.37/",
	pages = "281--286"
}

\appendix

\section{Tibetan NLP Conference/Workshop Papers Found from ACL Anthology}
\label{appendix_acl}

\subsection{Conference Papers}

\citet{li-etal-2026-diversity,yang-etal-2026-curated,su-etal-2026-reinforcement,xu-etal-2026-ftibsuite,han-etal-2026-scaling,bilitski-etal-2026-automatic,hashiloni-etal-2025-dharmabench,cao-etal-2025-human,xu-etal-2025-cmhg,gao-etal-2025-tlue,zhang-etal-2025-milic,zhuang-sun-2025-cute,zhang-etal-2024-mc2,sun-etal-2022-question,nehrdich-2022-sanstib,kuanzhuo-etal-2020-simplenlg,faggionato-meelen-2019-developing,nuo-etal-2015-tibetan,liu-etal-2014-zipfs,nuo-etal-2012-tibetan,liu-etal-2011-tibetan,liu-etal-2010-tibetan,jiang-2006-current,jiang-etal-2005-verbal,wagner-zeisler-2004-syntactically}.

\subsection{Workshop Papers}

\citet{cohen-etal-2026-scaling,liang-zerong-2026-tonogenesis,gurgurov-etal-2024-adapting,brown-2024-improved,an-2023-tibetan,long-an-2023-development,cao-etal-2023-pay-attention,faggionato-etal-2022-nlp,meelen-roux-2020-meta,di-etal-2019-modelling,lai-etal-2018-tibetan,liu-etal-2012-building}.

\subsection{Conference List}

\begin{flushleft}
$\cdot$ AACL: Asia-Pacific Chapter of the Association for Computational Linguistics
\\
$\cdot$ ACL: Annual Meeting of the Association for Computational Linguistics
\\
$\cdot$ COLING: International Conference on Computational Linguistics
\\
$\cdot$ EMNLP: Conference on Empirical Methods in Natural Language Processing
\\
$\cdot$ IJCNLP: International Joint Conference on Natural Language Processing
\\
$\cdot$ INLG: International Natural Language Generation Conference
\\
$\cdot$ LREC: International Conference on Language Resources and Evaluation
\\
$\cdot$ PACLIC: Pacific Asia Conference on Language, Information and Computation
\\
$\cdot$ RANLP: International Conference on Recent Advances in Natural Language Processing
\end{flushleft}

\subsection{Workshop List}

\begin{flushleft}
$\cdot$ ALP: Workshop on Ancient Language Processing
\\
$\cdot$ ALR: Workshop on Asian Language Resources
\\
$\cdot$ ALTA: Australasian Language Technology Association Workshop
\\
$\cdot$ EURALI: Workshop on Resources and Technologies for Indigenous, Endangered and Lesser-resourced Languages in Eurasia
\\
$\cdot$ KaLLM: Workshop on Knowledge Graphs and Large Language Models
\\
$\cdot$ LChange: International Workshop on Computational Approaches to Historical Language Change
\\
$\cdot$ LoResMT: Workshop on Technologies for MT of Low Resource Languages
\\
$\cdot$ NLP4DH: International Workshop on Natural Language Processing for Digital Humanities
\\
$\cdot$ TLT: International Workshop on Treebanks and Linguistic Theories
\\
$\cdot$ TrustNLP: Workshop on Trustworthy Natural Language Processing
\end{flushleft}

\section{Tibetan NLP Conference Papers / Journal Articles Found from DBLP}
\label{appendix_dblp}

\subsection{Conference Papers}

\citet{cao_tscheater_2025,wang_disg_2025,xu_tibetan_2024,hao_tibetan_2024,ouyang_pre-trained_2024,zhou_efficient_2024,fang_mlcl_2024,zhou_tibetan-chinese_2024,ren_method_2024,hao_tiltstibetan_2024,li_tibetan_2024,liu_multi-task_2023,qin_finer-grained_2022,liu_tibert_2022,zhu_new_2019,cai_online_2018,han_research_2018,wang_recognition_2018,wang_research_2018,zhao_touching_2018,shen_exploiting_2017,nuo_tibetan_2016,ma_online_2015,ye_tm-tot_2014,ma_semi-automatic_2013,su_cross-language_2013,li_study_2013,zhao_unsupervised_2012,ma_mrg-ohtc_2011}.

\subsection{Journal Articles}

\citet{tashi_tibetan_2026,zhang_cro-mtvits_2026,chen_dataset_2026,fan_tibetan-chinese_2026,jie_tibocr-bench_2026,ren_research_2026,wu_efficient_2026,zhuoma_multi-granularity_2026,moore_evaluating_2026,zhou_tibhcr_2026,lhundrub_tibstr-former_2026,zhang_tibetan_2026,wang_quality_2026,sha_tibetan-llama_2025,tang_tibetan_2025,zhuang_tibetan_2025,ke_align_2025,zhang_tibetan_2025-1,zhang_tibetan_2025,lv_t-llama_2025,zhuang_tifd_2025,khysru_research_2025,gyatso_improved_2025,kong_rpepl_2024,luo_cleansing_2024,lv_improved_2024,yu_novel_2024,dan_tibetanqa20_2024,tashi_construction_2024,liu_exploring_2024,bao_toward_2024,zhao_incremental_2024,liang_tibetan-bert-wwm_2024,an_prompt-based_2023,zhang_improving_2023,zhang_hutnet_2023,li_sentence_2022,li_character-based_2022,qin_improving_2022,li_character_2022,gan_tibetan-dependent_2022,an_neural_2021,krishna_applying_2021,li_finding_2021,long_recognition_2021,meelen_optimisation_2021,shi_multi-level_2021,sun_joint_2021,trinley_taming_2021,xu_end--end_2021,khysru_tibetan_2021,schmidt_grading_2020,wang_multitask_2020,ma_segmentation_2020,zhu_end--end_2020,wang_transliteration_2020,zhao_open_2020,xu_research_2019,xu_public_2019,khysru_morphological_2019,sun_qugan_2019,sun_improved_2019,sun_hybrid_2019,zhang_deep_2019,zhao_lhasa-tibetan_2019,zhao_end--end-based_2019,wang_online_2019,qiu_tibetan_2018,qiu_advanced_2018,xu_efficient_2018,xu_semantic_2017,qiu_tibetan_2017,yang_using_2015,ye_chinese-tibetan_2015}.

\subsection{Conference List}

\begin{flushleft}
$\cdot$ ACML: Asian Conference on Machine Learning
\\
$\cdot$ CSCWD: International Conference on Computer Supported Cooperative Work in Design
\\
$\cdot$ ICANN: International Conference on Artificial Neural Networks
\\
$\cdot$ ICASSP: IEEE International Conference on Acoustics, Speech, and Signal Processing
\\
$\cdot$ ICDAR: International Conference on Document Analysis and Recognition
\\
$\cdot$ ICONIP: International Conference on Neural Information Processing
\\
$\cdot$ ICPR: International Conference on Pattern Recognition
\\
$\cdot$ IJCNN: International Joint Conference on Neural Networks
\\
$\cdot$ Interspeech: Annual Conference of the International Speech Communication Association
\\
$\cdot$ NLPCC: CCF International Conference on Natural Language Processing and Chinese Computing
\\
$\cdot$ PRCV: Chinese Conference on Pattern Recognition and Computer Vision
\\
$\cdot$ SMC: IEEE International Conference on Systems, Man, and Cybernetics
\\
$\cdot$ TrustCom: IEEE International Conference on Trust, Security and Privacy in Computing and Communications
\end{flushleft}

\subsection{Journal List}

\begin{flushleft}
$\cdot$ ACM Transactions on Asian and Low-Resource Language Information Processing
\\
$\cdot$ Applied Soft Computing
\\
$\cdot$ Big Data
\\
$\cdot$ CAAI Transactions on Intelligence Technology
\\
$\cdot$ Cluster Computing
\\
$\cdot$ Complex \& Intelligent Systems
\\
$\cdot$ Complexity
\\
$\cdot$ Computer Speech \& Language
\\
$\cdot$ Data Intelligence
\\
$\cdot$ Engineering Applications of Artificial Intelligence
\\
$\cdot$ EURASIP Journal on Audio, Speech, and Music Processing
\\
$\cdot$ Expert Systems with Applications
\\
$\cdot$ IEEE Access
\\
$\cdot$ IEEE Signal Processing Letters
\\
$\cdot$ IEEE Transactions on Computational Social Systems
\\
$\cdot$ IEICE Transactions on Information and Systems
\\
$\cdot$ International Journal of Computational Science and Engineering
\\
$\cdot$ International Journal of Pattern Recognition and Artificial Intelligence
\\
$\cdot$ Journal of Advanced Computational Intelligence and Intelligent Informatics
\\
$\cdot$ Journal of Computational Methods in Sciences and Engineering
\\
$\cdot$ Journal of Visual Communication and Image Representation
\\
$\cdot$ KSII Transactions on Internet and Information Systems
\\
$\cdot$ Language Resources and Evaluation
\\
$\cdot$ Multimedia Tools and Applications
\\
$\cdot$ Neurocomputing
\end{flushleft}

\section{Publications Classified by Country of First Author's Affiliation}
\label{appendix_country}

\subsection{Australia}
\citet{di-etal-2019-modelling}.

\subsection{China}
\citet{li-etal-2026-diversity,yang-etal-2026-curated,su-etal-2026-reinforcement,xu-etal-2026-ftibsuite,han-etal-2026-scaling,cao-etal-2025-human,xu-etal-2025-cmhg,gao-etal-2025-tlue,zhang-etal-2025-milic,zhuang-sun-2025-cute,zhang-etal-2024-mc2,an-2023-tibetan,long-an-2023-development,cao-etal-2023-pay-attention,sun-etal-2022-question,kuanzhuo-etal-2020-simplenlg,lai-etal-2018-tibetan,nuo-etal-2015-tibetan,liu-etal-2014-zipfs,nuo-etal-2012-tibetan,liu-etal-2012-building,liu-etal-2011-tibetan,liu-etal-2010-tibetan,jiang-2006-current,jiang-etal-2005-verbal,wang_disg_2025,cao_tscheater_2025,li_tibetan_2024,hao_tiltstibetan_2024,ren_method_2024,zhou_tibetan-chinese_2024,fang_mlcl_2024,zhou_efficient_2024,ouyang_pre-trained_2024,hao_tibetan_2024,xu_tibetan_2024,liu_multi-task_2023,liu_tibert_2022,qin_finer-grained_2022,zhu_new_2019,zhao_touching_2018,wang_research_2018,wang_recognition_2018,han_research_2018,cai_online_2018,shen_exploiting_2017,nuo_tibetan_2016,ma_online_2015,ye_tm-tot_2014,li_study_2013,su_cross-language_2013,ma_semi-automatic_2013,zhao_unsupervised_2012,ma_mrg-ohtc_2011,lhundrub_tibstr-former_2026,zhang_tibetan_2026,zhuoma_multi-granularity_2026,wu_efficient_2026,ren_research_2026,jie_tibocr-bench_2026,fan_tibetan-chinese_2026,chen_dataset_2026,wang_quality_2026,tashi_tibetan_2026,zhang_cro-mtvits_2026,zhou_tibhcr_2026,sha_tibetan-llama_2025,tang_tibetan_2025,zhuang_tibetan_2025,ke_align_2025,zhang_tibetan_2025-1,zhang_tibetan_2025,lv_t-llama_2025,zhuang_tifd_2025,khysru_research_2025,gyatso_improved_2025,kong_rpepl_2024,lv_improved_2024,yu_novel_2024,dan_tibetanqa20_2024,tashi_construction_2024,liu_exploring_2024,bao_toward_2024,zhao_incremental_2024,liang_tibetan-bert-wwm_2024,an_prompt-based_2023,zhang_improving_2023,zhang_hutnet_2023,li_sentence_2022,li_character-based_2022,qin_improving_2022,li_character_2022,gan_tibetan-dependent_2022,an_neural_2021,li_finding_2021,long_recognition_2021,shi_multi-level_2021,sun_joint_2021,xu_end--end_2021,khysru_tibetan_2021,zhao_open_2020,wang_transliteration_2020,wang_multitask_2020,ma_segmentation_2020,zhu_end--end_2020,xu_research_2019,xu_public_2019,khysru_morphological_2019,sun_qugan_2019,sun_improved_2019,sun_hybrid_2019,zhang_deep_2019,zhao_lhasa-tibetan_2019,zhao_end--end-based_2019,wang_online_2019,qiu_tibetan_2018,qiu_advanced_2018,xu_efficient_2018,xu_semantic_2017,qiu_tibetan_2017,yang_using_2015,ye_chinese-tibetan_2015}.

\subsection{France}
\citet{trinley_taming_2021}.

\subsection{Germany}
\citet{gurgurov-etal-2024-adapting,nehrdich-2022-sanstib,wagner-zeisler-2004-syntactically}.

\subsection{India}
\citet{schmidt_grading_2020}.

\subsection{Israel}
\citet{cohen-etal-2026-scaling,bilitski-etal-2026-automatic,hashiloni-etal-2025-dharmabench}.

\subsection{UK}
\citet{faggionato-etal-2022-nlp,meelen-roux-2020-meta,faggionato-meelen-2019-developing,meelen_optimisation_2021}.

\subsection{USA}
\citet{liang-zerong-2026-tonogenesis,brown-2024-improved,moore_evaluating_2026,luo_cleansing_2024,krishna_applying_2021}.

\section{Publications Classified by Research Area in ARR}
\label{appendix_research_area}

\subsection{Clinical and Biomedical Applications}
\citet{zhang_tibetan_2025-1}.

\subsection{Computational Social Science and Cultural Analytics}
\citet{xu_tibetan_2024, ren_method_2024, ye_tm-tot_2014, yu_novel_2024, liang_tibetan-bert-wwm_2024, krishna_applying_2021, xu_research_2019, xu_public_2019, qiu_tibetan_2018, qiu_advanced_2018, xu_efficient_2018, qiu_tibetan_2017}.

\subsection{Dialogue and Interactive Systems}
None.

\subsection{Discourse and Pragmatics}
\citet{wang_disg_2025}.

\subsection{Efficient Methods for NLP}
\citet{han-etal-2026-scaling, zhou_efficient_2024, zhang_hutnet_2023}.

\subsection{Ethics, Bias, and Fairness}
None.

\subsection{Generation}
\citet{xu-etal-2025-cmhg, sun-etal-2022-question, kuanzhuo-etal-2020-simplenlg, zhuang_tibetan_2025, sun_qugan_2019}.

\subsection{Human-Centered NLP and Human-AI Interaction}
\citet{cao-etal-2025-human, schmidt_grading_2020}.

\subsection{Information Extraction}
\citet{zhang_tibetan_2025-1, sun_joint_2021, wang_transliteration_2020, sun_improved_2019}.

\subsection{Information Retrieval and Text Mining}
\citet{xu_tibetan_2024, ye_tm-tot_2014, xu_research_2019, xu_efficient_2018, xu_semantic_2017, ye_chinese-tibetan_2015}.

\subsection{Interpretability and Analysis of Models for NLP}
\citet{cao-etal-2025-human, cao-etal-2023-pay-attention, cao_tscheater_2025}.

\subsection{Language Modeling}
\citet{yang-etal-2026-curated, su-etal-2026-reinforcement, han-etal-2026-scaling, ouyang_pre-trained_2024, liu_tibert_2022, shen_exploiting_2017, chen_dataset_2026, sha_tibetan-llama_2025, tang_tibetan_2025, zhang_tibetan_2025, lv_t-llama_2025, zhuang_tifd_2025, liang_tibetan-bert-wwm_2024, an_prompt-based_2023, khysru_tibetan_2021, khysru_morphological_2019}.

\subsection{LLM agents}
None.

\subsection{Linguistic Theories, Cognitive Modeling, and Psycholinguistics}
\citet{liu-etal-2014-zipfs}.

\subsection{Machine Learning for NLP}
\citet{gurgurov-etal-2024-adapting, hao_tibetan_2024, ren_method_2024, cai_online_2018, zhao_unsupervised_2012, tashi_tibetan_2026, tang_tibetan_2025, zhang_tibetan_2025, lv_improved_2024, bao_toward_2024, zhao_incremental_2024}.

\subsection{Machine Translation}
\citet{lai-etal-2018-tibetan, zhou_tibetan-chinese_2024, fan_tibetan-chinese_2026, ren_research_2026, wang_quality_2026, gyatso_improved_2025, li_finding_2021}.

\subsection{Multilingualism and Cross-Lingual NLP}
\citet{li-etal-2026-diversity, yang-etal-2026-curated, su-etal-2026-reinforcement, hashiloni-etal-2025-dharmabench, gao-etal-2025-tlue, zhang-etal-2025-milic, zhuang-sun-2025-cute, zhang-etal-2024-mc2, gurgurov-etal-2024-adapting, cao-etal-2023-pay-attention, nehrdich-2022-sanstib, faggionato-etal-2022-nlp, nuo-etal-2012-tibetan, hao_tibetan_2024, zhou_efficient_2024, fang_mlcl_2024, zhou_tibetan-chinese_2024, li_tibetan_2024, liu_multi-task_2023, qin_finer-grained_2022, liu_tibert_2022, han_research_2018, ma_online_2015, su_cross-language_2013, zhang_cro-mtvits_2026, zhang_tibetan_2026, sha_tibetan-llama_2025, lv_t-llama_2025, yu_novel_2024, liu_exploring_2024, qin_improving_2022, trinley_taming_2021, xu_end--end_2021, wang_multitask_2020, zhu_end--end_2020, wang_transliteration_2020, xu_public_2019, zhang_deep_2019, zhao_end--end-based_2019, yang_using_2015, ye_chinese-tibetan_2015}.

\subsection{Multimodality and Language Grounding to Vision, Robotics and Beyond}
\citet{xu-etal-2026-ftibsuite, fang_mlcl_2024, li_tibetan_2024, cai_online_2018, han_research_2018, wang_recognition_2018, wang_research_2018, zhao_touching_2018, ma_online_2015, ma_semi-automatic_2013, su_cross-language_2013, ma_mrg-ohtc_2011, tashi_tibetan_2026, jie_tibocr-bench_2026, zhou_tibhcr_2026, lhundrub_tibstr-former_2026, ke_align_2025, tashi_construction_2024, bao_toward_2024, zhao_incremental_2024, zhang_hutnet_2023, li_character_2022, ma_segmentation_2020, wang_online_2019}.

\subsection{NLP and Code Models}
None.

\subsection{NLP and Symbolic Reasoning}
\citet{sun_joint_2021}.

\subsection{NLP Applications}
\citet{cohen-etal-2026-scaling, bilitski-etal-2026-automatic, long-an-2023-development, wang_recognition_2018, wang_research_2018, luo_cleansing_2024, krishna_applying_2021, schmidt_grading_2020, ma_segmentation_2020}.

\subsection{Phonology, Morphology, and Word Segmentation}
\citet{liang-zerong-2026-tonogenesis, brown-2024-improved, di-etal-2019-modelling, lai-etal-2018-tibetan, nuo-etal-2015-tibetan, liu-etal-2011-tibetan, liu-etal-2010-tibetan, zhu_new_2019, shen_exploiting_2017, li_study_2013, zhuoma_multi-granularity_2026, gyatso_improved_2025, li_sentence_2022, li_character-based_2022, li_finding_2021, khysru_tibetan_2021, khysru_morphological_2019}.

\subsection{Question Answering}
\citet{sun-etal-2022-question, zhuang_tibetan_2025, dan_tibetanqa20_2024, sun_qugan_2019, sun_hybrid_2019}.

\subsection{Resources and Evaluation}
\citet{li-etal-2026-diversity, xu-etal-2026-ftibsuite, bilitski-etal-2026-automatic, hashiloni-etal-2025-dharmabench, xu-etal-2025-cmhg, gao-etal-2025-tlue, zhang-etal-2025-milic, zhuang-sun-2025-cute, zhang-etal-2024-mc2, long-an-2023-development, nehrdich-2022-sanstib, meelen-roux-2020-meta, faggionato-meelen-2019-developing, liu-etal-2014-zipfs, liu-etal-2012-building, jiang-2006-current, jiang-etal-2005-verbal, wagner-zeisler-2004-syntactically, cao_tscheater_2025, hao_tiltstibetan_2024, zhao_touching_2018, ma_semi-automatic_2013, ma_mrg-ohtc_2011, chen_dataset_2026, jie_tibocr-bench_2026, ren_research_2026, zhuoma_multi-granularity_2026, moore_evaluating_2026, zhou_tibhcr_2026, wang_quality_2026, zhuang_tifd_2025, dan_tibetanqa20_2024, tashi_construction_2024, li_character_2022, an_neural_2021, meelen_optimisation_2021, shi_multi-level_2021, trinley_taming_2021, zhao_open_2020, wang_online_2019}.

\subsection{Semantics: Lexical and Sentence-Level}
\citet{cohen-etal-2026-scaling, nuo_tibetan_2016, lv_improved_2024, xu_semantic_2017}.

\subsection{Sentiment Analysis, Stylistic Analysis, and Argument Mining}
\citet{zhang_tibetan_2026, kong_rpepl_2024, qiu_advanced_2018, qiu_tibetan_2017}.

\subsection{Speech Recognition, Text-to-Speech and Spoken Language Understanding}
\citet{liang-zerong-2026-tonogenesis, liu_multi-task_2023, qin_finer-grained_2022, zhao_unsupervised_2012, zhang_cro-mtvits_2026, fan_tibetan-chinese_2026, wu_efficient_2026, moore_evaluating_2026, khysru_research_2025, liu_exploring_2024, zhang_improving_2023, qin_improving_2022, gan_tibetan-dependent_2022, xu_end--end_2021, wang_multitask_2020, zhu_end--end_2020, zhao_open_2020, zhang_deep_2019, zhao_lhasa-tibetan_2019, zhao_end--end-based_2019, yang_using_2015}.

\subsection{Summarization}
\citet{wang_disg_2025, ouyang_pre-trained_2024, hao_tiltstibetan_2024}.

\subsection{Syntax: Tagging, Chunking and Parsing}
\citet{an-2023-tibetan, faggionato-etal-2022-nlp, meelen-roux-2020-meta, faggionato-meelen-2019-developing, nuo-etal-2012-tibetan, jiang-etal-2005-verbal, wagner-zeisler-2004-syntactically, li_study_2013, li_sentence_2022, li_character-based_2022, an_neural_2021, long_recognition_2021, meelen_optimisation_2021, shi_multi-level_2021}.

\section{Full-Scale Versions of Figure \ref{fig_year_venue}, \ref{fig_venues}, \ref{fig_area_country}, and \ref{fig_heatmap}}
\label{appendix_figure}

\begin{sidewaysfigure*}
		\includegraphics[width=\linewidth]{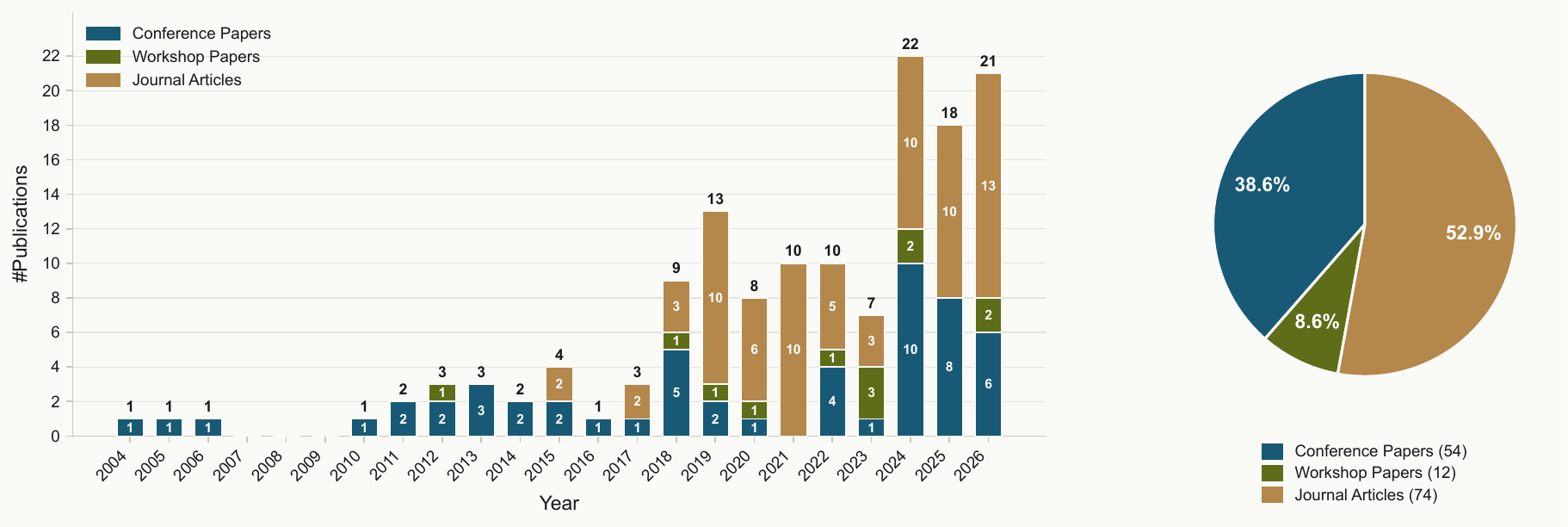}
		\captionsetup{justification=justified, singlelinecheck=false}
		\caption{Full-scale version of Figure~\ref{fig_year_venue}: distribution of the 140 Tibetan NLP publications by year and venue type, 2004--2026. \textbf{Left:} stacked bars give the annual number of conference papers (blue), workshop papers (green), and journal articles (tan); numbers inside the segments are per-type counts, and numbers above the bars are annual totals. No publications were identified for 2007--2009. Output rises from three papers in 2017 to nine in 2018 and peaks at 22 in 2024; 2026 is covered only up to \cutoffdate. \textbf{Right:} overall share of each venue type: journal articles 52.9\% (74), conference papers 38.6\% (54), and workshop papers 8.6\% (12).}
		\label{fig_year_venue_appendix}
\end{sidewaysfigure*}

\begin{sidewaysfigure*}
	\includegraphics[width=\linewidth]{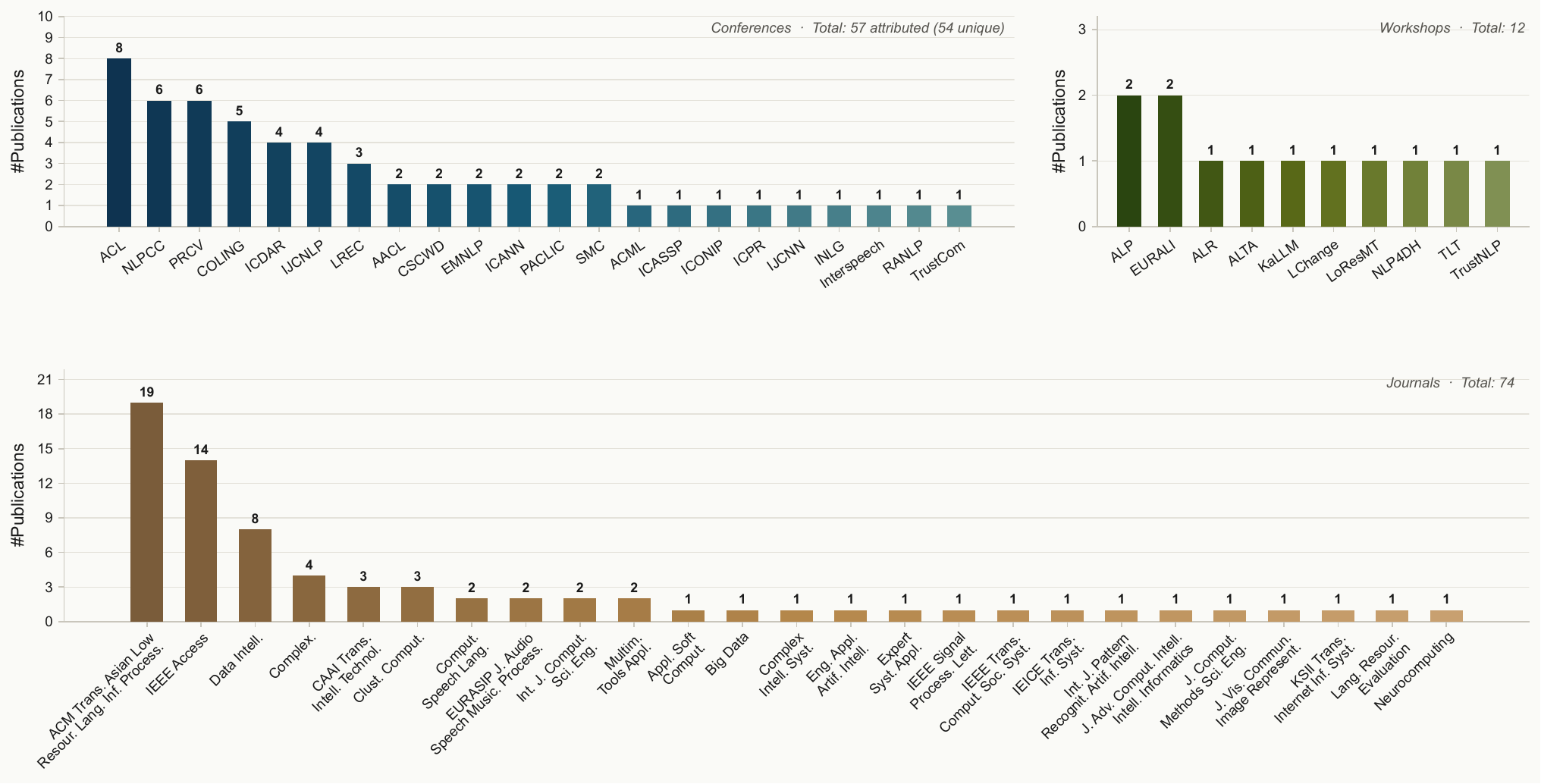}
	\captionsetup{justification=justified, singlelinecheck=false}
	\caption{Full-scale version of Figure~\ref{fig_venues}: number of Tibetan NLP publications per venue; within each panel, venues are sorted by count in descending order, with ties listed alphabetically. \textbf{Top left:} 22 conferences, led by ACL (8, including Findings of ACL), NLPCC (6), PRCV (6), and COLING (5). Papers at joint conferences (ACL-IJCNLP 2015 and IJCNLP-AACL 2025) are counted under each constituent venue, so the bars sum to 57 attributions for 54 conference papers. \textbf{Top right:} 10 workshops with 12 papers, led by ALP and EURALI (2 each). \textbf{Bottom:} 25 journals with 74 articles, led by \textit{ACM Transactions on Asian and Low-Resource Language Information Processing} (19), \textit{IEEE Access} (14), and \textit{Data Intelligence} (8); journal names follow DBLP abbreviations. Full venue names are listed in Appendices~A and~B.}
	\label{fig_venues_appendix}
\end{sidewaysfigure*}

\begin{sidewaysfigure*}
		\includegraphics[width=\linewidth]{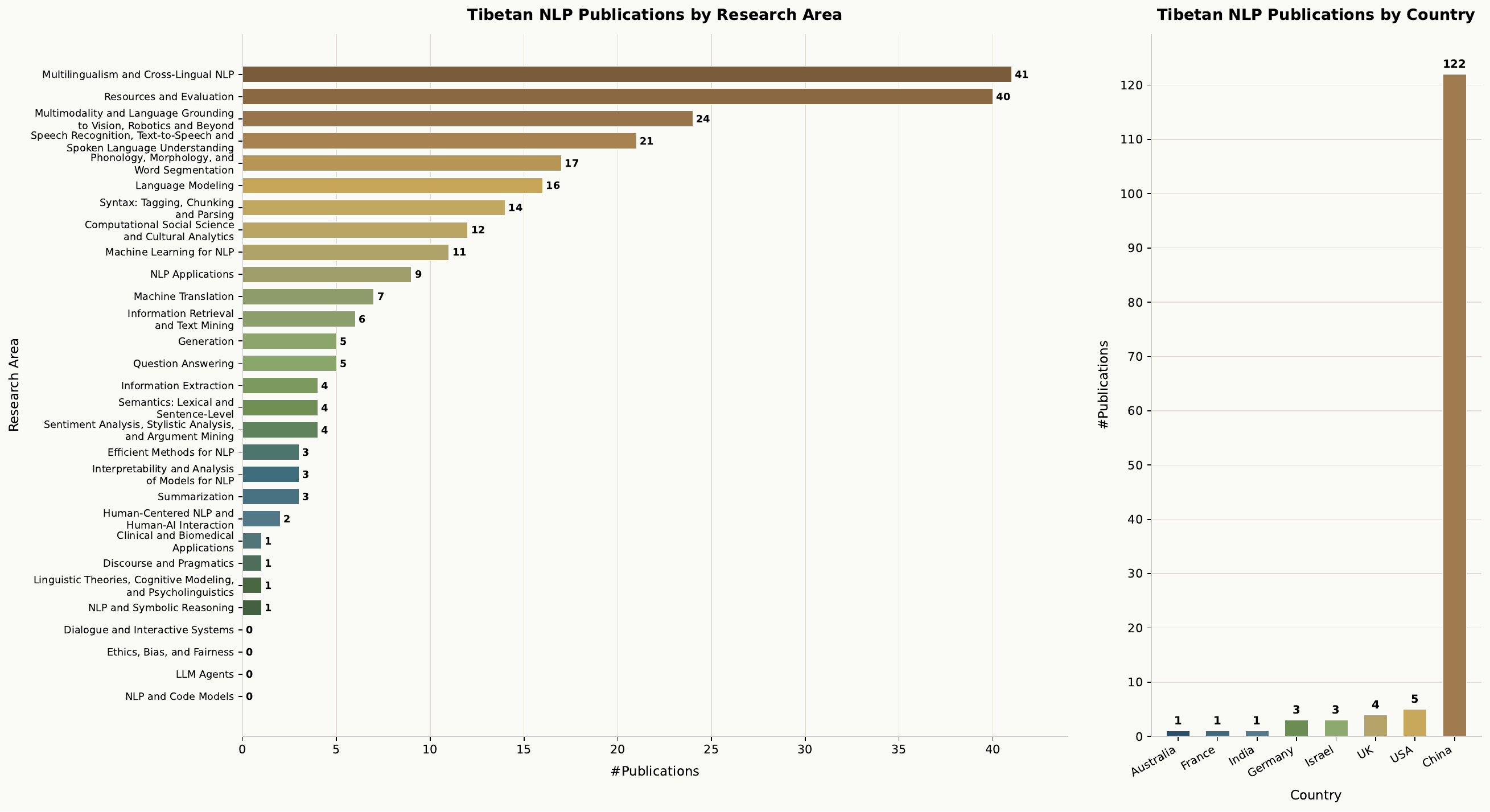}
		\captionsetup{justification=justified, singlelinecheck=false}
		\caption{Full-scale version of Figure~\ref{fig_area_country}: number of Tibetan NLP publications by ACL Rolling Review (ARR) research area and by country of the first author's primary affiliation. \textbf{Left:} the 29 ARR research areas, sorted in descending order. Each publication was assigned up to three areas (in practice, 115 received two and 25 received one), so the counts sum to 255 rather than 140. Multilingualism and Cross-Lingual NLP (41) and Resources and Evaluation (40) lead, whereas Dialogue and Interactive Systems; Ethics, Bias, and Fairness; LLM Agents; and NLP and Code Models have no publications. \textbf{Right:} countries, sorted in ascending order: China accounts for 122 (87.1\%), followed by the USA (5), the UK (4), Germany (3), Israel (3), and Australia, France, and India (one each). Per-country and per-area publication lists are given in Appendices~C and~D, respectively.}
		\label{fig_area_country_appendix}
\end{sidewaysfigure*}

\begin{sidewaysfigure*}
	\includegraphics[width=\linewidth]{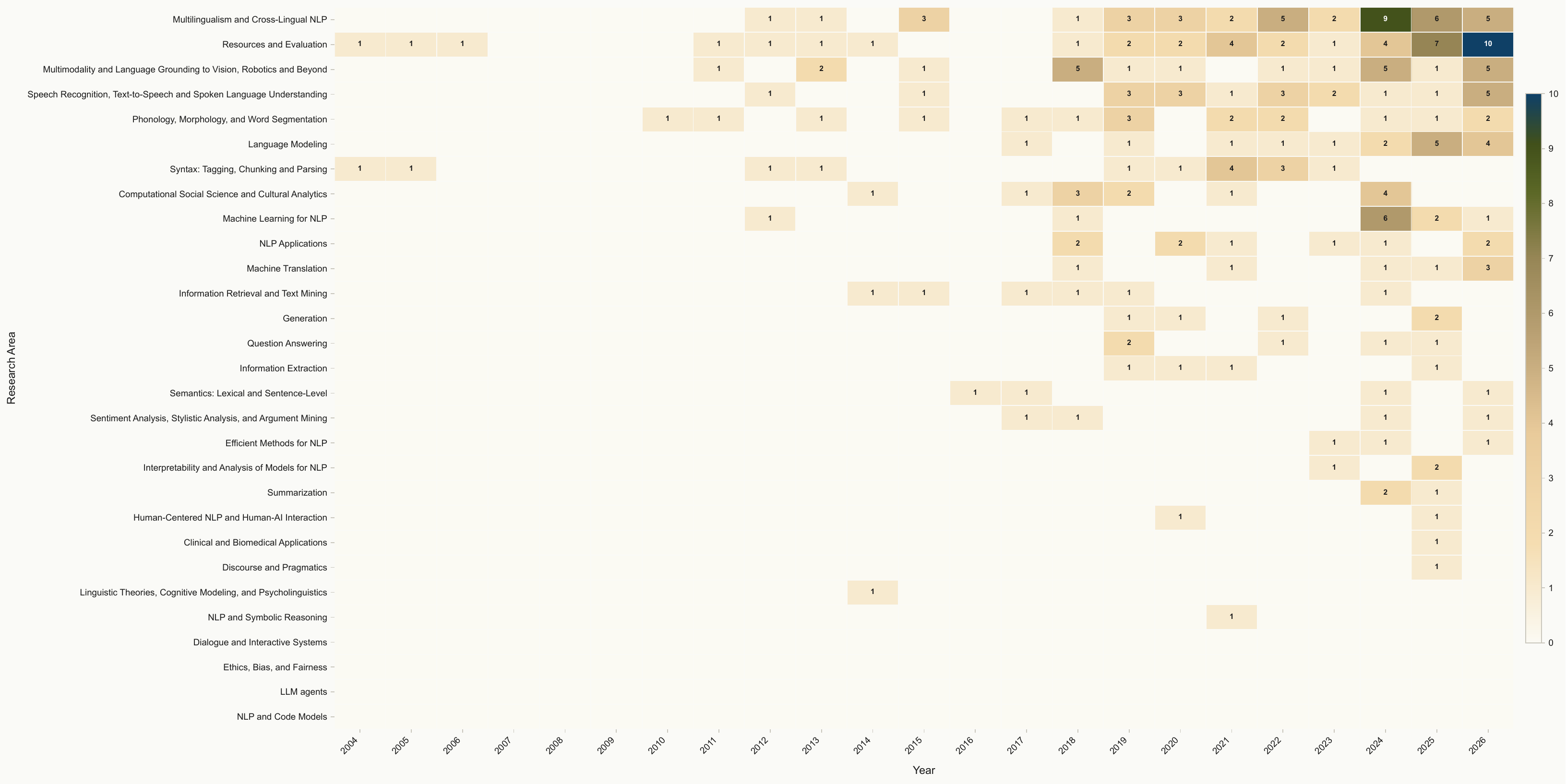}
	\captionsetup{justification=justified, singlelinecheck=false}
	\caption{Full-scale version of Figure~\ref{fig_heatmap}: number of Tibetan NLP publications in each of the 29 ARR research areas (rows) and each year (columns), 2004--2026. Rows follow the order of Figure~\ref{fig_area_country_appendix} (descending total count). Cell shading darkens with the number of publications (scale 0--10); each non-empty cell shows its count, and blank cells indicate no publications. Because a publication can belong to more than one area, the cells sum to 255. The largest values are Resources and Evaluation in 2026 (10) and Multilingualism and Cross-Lingual NLP in 2024 (9); the columns for 2007--2009 are empty, and 2026 is covered only up to \cutoffdate.}
	\label{fig_heatmap_appendix}
\end{sidewaysfigure*}

Figure \ref{fig_year_venue_appendix}, \ref{fig_venues_appendix}, \ref{fig_area_country_appendix}, and \ref{fig_heatmap_appendix} are the full-scale versions of Figure \ref{fig_year_venue}, \ref{fig_venues}, \ref{fig_area_country}, and \ref{fig_heatmap}, respectively.

\section{Open-Source Resources Released by the Surveyed Publications}
\label{appendix_open_source}

Bibliographic metadata cannot reveal whether the datasets, models, and code described in a publication are publicly released. To complement the bibliometric analysis, we manually examined each of the 140 publications for resources released by its authors; Table~\ref{tab:open-source} lists the results.

\paragraph{Compilation.} A resource is included when the authors of a publication release code, data, or models at a publicly accessible URL; resources available only on request and third-party re-uploads are excluded. For each entry, the table gives the name of the resource, the publications that introduce it, its URLs, and a one-sentence summary of its content and, where reported, its scale and licence. A resource described in several publications is listed once, under the earliest of them: this applies to the TIBMD@MUC speech corpus \citep{zhao_end--end-based_2019,zhao_open_2020} and to the PACTib/ACTib corpora and annotation pipelines \citep{meelen-roux-2020-meta,meelen_optimisation_2021,faggionato-etal-2022-nlp}. Entries are ordered by year of publication and then alphabetically by first author.

\paragraph{Coverage and trend.} The table contains 38 entries from 41 of the 140 publications (29.3\%). Public release is a recent practice in Tibetan NLP. None of the 31 publications from 2004--2018 releases a resource that meets our criteria, compared with 14 of the 48 publications from 2019--2023 (29.2\%) and 27 of the 61 from 2024--2026 (44.3\%), including 12 of the 18 publications of 2025. This rise coincides with the transition to pre-trained and large language models, a period in which releasing corpora, models, and benchmarks has become common practice across NLP.

\paragraph{What is released.} The released resources span every layer of the pipeline described in Section~\ref{sec_areas}. At the foundational layer, they include a lexicon of verbal paradigms \citep{di-etal-2019-modelling}, the PACTib/ACTib treebank and annotation pipelines \citep{meelen-roux-2020-meta,meelen_optimisation_2021,faggionato-etal-2022-nlp}, a knowledge graph \citep{sun_joint_2021}, a word segmenter \citep{gyatso_improved_2025}, and expert word-boundary annotations \citep{moore_evaluating_2026}. Script processing is represented by character datasets of historical documents, handwriting, and cursive script \citep{li_character_2022,tashi_construction_2024,zhou_tibhcr_2026}, code for few-shot character recognition \citep{bao_toward_2024}, a multimodal OCR benchmark \citep{jie_tibocr-bench_2026}, and a vision--language model \citep{xu-etal-2026-ftibsuite}. Speech is represented by the TIBMD@MUC multi-dialect corpus \citep{zhao_end--end-based_2019,zhao_open_2020} and a package for measuring recognition error at the word level \citep{moore_evaluating_2026}. Since 2022, releases have increasingly consisted of PLMs and LLMs \citep{liu_tibert_2022,liang_tibetan-bert-wwm_2024,lv_t-llama_2025,sha_tibetan-llama_2025,zhuang-sun-2025-cute}, the data used to train and align them \citep{zhang-etal-2024-mc2,zhuang_tifd_2025}, and benchmarks for evaluating them; six benchmarks date from 2025--2026 alone \citep{gao-etal-2025-tlue,xu-etal-2025-cmhg,hashiloni-etal-2025-dharmabench,zhang-etal-2025-milic,li-etal-2026-diversity,jie_tibocr-bench_2026}. Three further releases, all from one research team, support the evaluation of adversarial robustness \citep{cao-etal-2023-pay-attention,cao_tscheater_2025,cao-etal-2025-human}. Finally, Classical Tibetan and Buddhist textual heritage accounts for a distinct cluster of releases: a Sanskrit--Tibetan parallel corpus \citep{nehrdich-2022-sanstib}, tools and text analytics for Tibetan Buddhist literature \citep{trinley_taming_2021,krishna_applying_2021}, and recent resources for benchmarking, distinguishing native from translated passages, and measuring sentence similarity \citep{hashiloni-etal-2025-dharmabench,bilitski-etal-2026-automatic,cohen-etal-2026-scaling}.

\paragraph{Release rates by research area.} The share of publications that release resources varies widely across ARR areas (Appendix~D). It is highest in Resources and Evaluation (22 of 40; 55.0\%), Language Modeling (7 of 16; 43.8\%), and Multilingualism and Cross-Lingual NLP (16 of 41; 39.0\%). Among the larger areas, it is lowest in Speech Recognition, Text-to-Speech and Spoken Language Understanding (3 of 21; 14.3\%) and in Phonology, Morphology, and Word Segmentation (2 of 17; 11.8\%), and none of the publications in Information Retrieval and Text Mining (6), Sentiment Analysis, Stylistic Analysis, and Argument Mining (4), or Summarization (3) releases a resource. Speech and word segmentation are the areas most directly shaped by two defining properties of Tibetan, dialectal diversity and unmarked word boundaries, yet they have the fewest shared resources. Consent and privacy considerations can legitimately restrict the release of speech data, but the result is that TIBMD@MUC is the only entry covering the three main dialect groups, and no text resource in the table targets a specific dialect through genres such as proverbs and folk song lyrics.

\paragraph{Release rates by country and venue.} Release rates also differ by the country of the first author's affiliation (Appendix~C). Of the 18 publications from outside China, 13 (72.2\%) release resources, compared with 28 of the 122 China-affiliated publications (23.0\%). Publication venue accounts for part of this difference. Publications in the ACL Anthology release resources more often than those retrieved from DBLP (18 of 37, 48.6\%, vs.\ 23 of 103, 22.3\%), and among China-affiliated publications, the rate is 40.0\% (10 of 25) in the ACL Anthology but 18.6\% (18 of 97) elsewhere. This pattern is consistent with the artifact-release expectations of ACL venues, such as the Responsible NLP Research Checklist. In absolute terms, China-affiliated teams remain the largest contributors, accounting for 28 of the 41 releasing publications, all of the released LLMs, and five of the six recent benchmarks. All of these rates are lower bounds, because resources shared on request or through channels that the publications do not cite are not captured.

\paragraph{Accessibility and sustainability.} GitHub\footnote{\url{https://github.com/}} is the most common host (28 entries), followed by Hugging Face\footnote{\url{https://huggingface.co/}} (13 entries), and 11 entries distribute their resources across more than one platform, for example Hugging Face with GitHub \citep{zhou_tibhcr_2026} or ModelScope\footnote{\url{https://www.modelscope.cn/}} with GitHub \citep{han-etal-2026-scaling}. Only four entries are deposited in research-data repositories that assign persistent identifiers (ScienceDB\footnote{\url{https://www.scidb.cn/}} and Zenodo\footnote{\url{https://zenodo.org/}}); the others depend on platforms on which repositories can be renamed, moved, or deleted. Two datasets, TIBMD@MUC and THCU, are distributed only through Baidu Netdisk\footnote{\url{https://pan.baidu.com/}}, which can be difficult to use from outside China, while some international platforms are, conversely, not reliably accessible from mainland China. Building on the open-science recommendations in Section~\ref{sec_recommendations}, we encourage authors of Tibetan NLP resources to (1)~deposit released data and models in archives that issue persistent identifiers, (2)~state an explicit licence, (3)~document each resource with a datasheet or model card, and (4)~mirror resources on platforms accessible to researchers both inside and outside China.

\paragraph{Caveats.} Inclusion in Table~\ref{tab:open-source} indicates that a resource has been publicly released, not that it is complete, documented, or maintained; we did not evaluate the released resources themselves. As with the rest of the survey, 2026 is only partially covered. We ensure that all links in Table~\ref{tab:open-source} are accessible at \cutoffdate, but we cannot guarantee their accessibility in the future.

\section{NLP Publications on Other Bodish Languages}
\label{appendix_bodish}

Complementing the recommendation on related languages in Section~\ref{sec_recommendations}, Table~\ref{tab:bodish-languages} extends the retrieval of Section~\ref{sec_method} from Tibetan to all 82 languages of Bodish, the subgroup of Tibeto-Burman to which Ethnologue\footnote{\url{https://www.ethnologue.com/subgroup/4085/}} assigns Tibetan. For each language, we repeated the retrieval and screening with the keyword in the third column in place of \textit{Tibetan}. Entries that differ only in a qualifier, such as the three Tibetan or the three Tamang entries, share a keyword, and Sikkimese was also searched under its alternative name \textit{Bhutia}.

The keyword \textit{Tibetan} retrieves the survey corpus itself (Appendices~\ref{appendix_acl} and~\ref{appendix_dblp}). For the other 79 languages, the search yielded only 12 publications, against 140 for Tibetan: six on Dzongkha, three on Tamang, and one each on Balti, Ladakhi, and Sikkimese. They address text processing for Dzongkha and Sikkimese (e.g., word segmentation and tagged corpora), machine translation for Tamang, and speech processing for Balti and Ladakhi. The Tamang publications, whose titles and abstracts name no variety, are listed under Eastern Tamang. The keywords of the remaining 72 languages, including all those outside Central and West Bodish, yielded none. These 12 publications are outside the survey corpus and the analyses of Section~\ref{sec_results}. As retrieval covers only the ACL Anthology and DBLP and depends on the keywords, an empty cell means that no publication was retrieved, not that no research on the language exists.

\clearpage

\clearpage

\clearpage

\clearpage

\clearpage

\onecolumn
\begingroup
\scriptsize
\hbadness=10000
\urlstyle{same}
\expandafter\def\expandafter\UrlBreaks\expandafter{\UrlBreaks\do\-}
\setlength{\tabcolsep}{4pt}
\renewcommand{\arraystretch}{1.12}
\setlength{\LTcapwidth}{\textwidth}
\captionsetup{position=below}
\begin{xltabular}{\textwidth}{@{}%
		>{\raggedright\arraybackslash}p{0.13\textwidth}%
		>{\raggedright\arraybackslash}p{0.13\textwidth}%
		>{\raggedright\arraybackslash}p{0.30\textwidth}%
		>{\raggedright\arraybackslash}X@{}}
	\toprule
	\textbf{Open-source Resource} & \textbf{Paper} & \textbf{Repository URL} & \textbf{Description} \\
	\midrule
	\endfirsthead
	\multicolumn{4}{@{}l}{\textit{(continued from previous page)}} \\
	\toprule
	\textbf{Open-source Resource} & \textbf{Paper} & \textbf{Repository URL} & \textbf{Description} \\
	\midrule
	\endhead
	\midrule
	\multicolumn{4}{r@{}}{\textit{(continued on next page)}} \\
	\endfoot
	\bottomrule
	\caption{Open-source resources released by the surveyed publications, comprising corpora, datasets and benchmarks, pre-trained and large language models, and code and toolkits. The 38 entries come from 41 of the 140 publications (29.3\%); only resources released by the papers' own authors at a publicly accessible URL are included. Entries are ordered by year of publication and then alphabetically by first author. A resource described in several papers (TIBMD@MUC and PACTib/ACTib) is listed once, under its earliest paper, with all related papers cited. Descriptions summarize each resource's content and, where reported, its scale and licence.}\label{tab:open-source} \\
	\endlastfoot
	Tibetan-Verb-Lexicon &
	\citet{di-etal-2019-modelling} &
	\url{https://github.com/victoriadqj/Tibetan-Verb-Lexicon} &
	A UniMorph-style dataset of Tibetan verbal paradigms (1,433 lemmas with inflected forms) for morphological reinflection. \\
	TIBMD@MUC &
	\citet{zhao_end--end-based_2019,zhao_open_2020} &
	\url{https://pan.baidu.com/s/14CihgqjA4AFFH1QpSTjzZw} &
	A free multi-dialect Tibetan speech corpus (\"{U}-Tsang, Kham and Amdo) with a phoneme set, a pronunciation dictionary and baseline recognition code, used for joint speech content recognition, dialect identification and speaker recognition. \\
	PACTib / ACTib and annotation pipelines &
	\citet{meelen-roux-2020-meta,meelen_optimisation_2021,faggionato-etal-2022-nlp} &
	\url{https://github.com/lothelanor/actib}\newline
	\url{https://zenodo.org/record/3951503} &
	A family of Classical Tibetan corpora and tools comprising a 166M-token diachronic treebank (11th--21st c.) with segmentation, POS tags, phrase-structure parses and date metadata; ACTib v2.0, segmented and POS-tagged in small (UD) and large tag sets by a pipeline that combines rule-based, memory-based and neural methods; and an off-the-shelf pipeline from normalisation to parsing and information-structure annotation that is adaptable to Newar varieties. \\
	Children's Stories corpus &
	\citet{schmidt_grading_2020} &
	\url{https://github.com/Esukhia/children-stories} &
	A corpus of transcribed SayMore sessions of diaspora Tibetan children's speech whose word-frequency lists the Dakje readability tool uses to grade a children's picture book. \\
	bo-corpus-analytics &
	\citet{krishna_applying_2021} &
	\url{https://github.com/keutzer/bo-corpus-analytics} &
	Unicode e-texts of the Bon and Nyingma Mind-section corpora, with notebooks for the TF-IDF, cosine-similarity, LDA and $k$-means analyses of Great Perfection literature. \\
	TD50K &
	\citet{sun_joint_2021} &
	\url{https://github.com/andongBlue/TiData50K} &
	A partial release of a Tibetan knowledge-graph dataset (53,797 triples, 12,573 entities and 3,285 relations) built from a Tibetan online encyclopedia. \\
	OpenPecha Toolkit &
	\citet{trinley_taming_2021} &
	\url{https://github.com/OpenPecha/openpecha-toolkit} &
	A command-line toolkit for downloading OpenPecha texts, updating and transferring annotations, and exporting views as Markdown, DOCX, PDF or EPUB. \\
	THCU &
	\citet{li_character_2022} &
	\url{https://pan.baidu.com/s/1i1-qMcLqR0vvFJonuj5LeA}\newline
	(extraction code: \texttt{h5ga}) &
	An offline character dataset of Uchen-font Tibetan historical documents that pairs 121,214 manually annotated samples over 238 classes (THCU-M) with component-combination synthetic subsets of up to 7,238 classes (THCU-S). \\
	TiBERT &
	\citet{liu_tibert_2022} &
	\url{https://huggingface.co/CMLI-NLP/TiBERT} &
	A monolingual BERT-base Tibetan pre-trained language model trained on 3.56\,GB of Tibetan web text with a 30,005-token SentencePiece unigram vocabulary. \\
	SansTib &
	\citet{nehrdich-2022-sanstib} &
	\url{https://github.com/sebastian-nehrdich/sanstib}\newline
	\url{https://huggingface.co/buddhist-nlp/sanstib} &
	A Sanskrit--Tibetan parallel corpus of 317,289 aligned sentence pairs with 6,916 gold pairs, released with a bilingual sentence-embedding model. \\
	TSAttacker &
	\citet{cao-etal-2023-pay-attention} &
	\url{https://github.com/UTibetNLP/TSAttacker}\newline
	\url{https://github.com/metaphors/TibetanPLMsFineTuning}\newline
	\url{https://github.com/metaphors/TibetanAdversarialAttack} &
	A syllable-level black-box adversarial attack on Tibetan script for testing the robustness of Tibetan language models. \\
	fs-TCR &
	\citet{bao_toward_2024} &
	\url{https://github.com/coder-bossin/fs-TCR} &
	Code for cross-domain few-shot Tibetan character recognition that uses contrastive pre-training and meta fine-tuning with episodic synthetic query sets. \\
	TibetanQA2.0 &
	\citet{dan_tibetanqa20_2024} &
	\url{https://doi.org/10.57760/sciencedb.j00104.00120} &
	A crowd-sourced Tibetan machine reading comprehension dataset of 505 passages and 1,347 unanswerable question--answer pairs over six topics, released under CC BY 4.0. \\
	ConceptNet language adapters &
	\citet{gurgurov-etal-2024-adapting} &
	\url{https://github.com/d-gurgurov/Injecting-Commonsense-Knowledge-into-LLMs}\newline
	\url{https://github.com/d-gurgurov/Conceptnet-Embeddings} &
	Code, ConceptNet-extracted data and language adapters that inject knowledge-graph knowledge into multilingual LLMs for eight low-resource languages, including Tibetan. \\
	Tibetan-PLM &
	\citet{liang_tibetan-bert-wwm_2024} &
	\url{https://github.com/Dslab-NLP/Tibetan-PLM}\newline
	\url{https://huggingface.co/Pagewood/TBERT-pytorch}\newline
	\url{https://huggingface.co/Pagewood/TBERT-tensorflow}\newline
	\url{https://huggingface.co/Pagewood/Tibetan-BERT-wwm} &
	The Tibetan-BERT and Tibetan-BERT-wwm pre-trained models, the latter using whole word masking to learn word-level representations for Tibetan text classification. \\
	Tibetan Handwriting Khyug-yig Dataset &
	\citet{tashi_construction_2024} &
	\url{https://github.com/13209413223/Tibetan-Handwriting-Dataset} &
	A dataset of 9,874 labelled word images in the cursive Khyug-yig style written by 63 writers, with more data to be released incrementally. \\
	MC$^2$ &
	\citet{zhang-etal-2024-mc2} &
	\url{https://github.com/luciusssss/mc2_corpus}\newline
	\url{https://huggingface.co/datasets/pkupie/mc2_corpus}\newline
	\url{https://huggingface.co/pkupie/mc2-xlmr-large}\newline
	\url{https://huggingface.co/pkupie/mc2-llama-13b} &
	An open pre-training corpus for Tibetan, Uyghur, Kazakh (Arabic script) and Mongolian (traditional script). \\
	Ti-ChTrans &
	\citet{zhou_tibetan-chinese_2024} &
	\url{https://github.com/UTibetNLP/Ti-ChTrans} &
	Transformer Tibetan--Chinese translation models trained on CINO-filtered parallel data, released with an expert-reviewed benchmark test set of 1,000 sentence pairs (483 common phrases and 517 news sentences). \\
	TSCheater, TSVSDB \& AdvTS &
	\citet{cao_tscheater_2025} &
	\url{https://github.com/metaphors/TSAttack}\newline
	\url{https://huggingface.co/collections/UTibetNLP/tibetan-victim-language-models-669f614ecea872c7211c121c} &
	The TSCheater visual-similarity-based Tibetan adversarial text generator, released with the TSVSDB syllable visual-similarity database (18,833 syllables), AdvTS (the first human-proofread Tibetan adversarial robustness benchmark) and eight victim classifiers (Tibetan-BERT and CINO-small/base/large, each fine-tuned on TNCC-title and TU\_SA). \\
		HITL-GAT &
	\citet{cao-etal-2025-human} &
	\url{https://github.com/metaphors/HITL-GAT} &
	A human-in-the-loop system for generating adversarial texts, released under GPL-3.0. \\
	TLUE &
	\citet{gao-etal-2025-tlue} &
	\url{https://github.com/Vicentvankor/TLUE} &
	The first large-scale Tibetan LLM benchmark, pairing a multi-task understanding set (67 subdomains) with a safety set. \\
	Tibetan-tokenizer &
	\citet{gyatso_improved_2025} &
	\url{https://github.com/gyatso736/-Tibetan-tokenizer-} &
	A Bi-LSTM+CRF Tibetan word segmenter used to pre-segment text before the paper's agglutination-aware BPE for Tibetan--Chinese machine translation. \\
	DharmaBench &
	\citet{hashiloni-etal-2025-dharmabench} &
	\url{https://github.com/Intellexus-DSI/DharmaBench} &
	A 13-task benchmark on Buddhist texts in Sanskrit and Classical Tibetan, with datasets (CC-BY-4.0), code (Apache-2.0) and prompts. \\
	T-LLaMA &
	\citet{lv_t-llama_2025} &
	\url{https://huggingface.co/Pagewood/T-LLaMA} &
	A LLaMA~2-based Tibetan large language model with a SentencePiece-expanded Tibetan vocabulary, trained on a 2.2-billion-character Tibetan corpus. \\
	Tibetan-LLaMA 2 &
	\citet{sha_tibetan-llama_2025} &
	\url{https://github.com/Shajiu/Tibetan-LLaMA-2}\newline
	\url{https://huggingface.co/shajiu/Yak_Llama2_7B}\newline
	\url{https://huggingface.co/shajiu/Yak_Llama2_13B}\newline
	\url{https://huggingface.co/datasets/shajiu/TibetanGeneral_corpus}\newline
	\url{https://huggingface.co/datasets/shajiu/TibetanSft_corpus} &
	A LLaMA~2 model extended with a Tibetan vocabulary, continually pre-trained on a 15\,GB Tibetan corpus and instruction-tuned on 1M instructions, released with inference resources. \\
	Tibetan prompt-learning text classifier &
	\citet{tang_tibetan_2025} &
	\url{https://github.com/Greenhorntc/Text-Classification-in-Tibetan-Language-based-on-Prompt-Learning} &
	Code that recasts Tibetan text classification as prompt-based entailment and ensembles several prompt-tuned language models, evaluated on TNCC and WCM. \\
	CMHG &
	\citet{xu-etal-2025-cmhg} &
	\url{https://huggingface.co/datasets/KEVVVV/CMHG} &
	A headline-generation dataset (100K Tibetan, 50K Uyghur and 50K Mongolian) with a native-speaker-annotated test set. \\
	MiLiC-Eval &
	\citet{zhang-etal-2025-milic} &
	\url{https://github.com/luciusssss/MiLiC-Eval}\newline
	\url{https://huggingface.co/datasets/pkupie/milic-eval} &
	A 24K-instance, 9-task benchmark that evaluates LLMs on Tibetan, Uyghur, Kazakh and Mongolian in parallel. \\
	CUTE &
	\citet{zhuang-sun-2025-cute} &
	\url{https://github.com/CMLI-NLP/CUTE}\newline
	\url{https://huggingface.co/datasets/CMLI-NLP/CUTE-Datasets}\newline
	\url{https://huggingface.co/CMLI-NLP/CUTE-Llama} &
	Parallel and non-parallel 25\,GB Chinese--Uyghur--Tibetan--English corpora, released with CUTE-Llama models trained on each. \\
	TIFD &
	\citet{zhuang_tifd_2025} &
	\url{https://doi.org/10.57760/sciencedb.j00104.00146} &
	A Tibetan instruction-following dataset of 11,535 instruction--input--output records, translated from GPT-4 responses and manually proofread, for supervised fine-tuning of LLMs. \\
	ALLO/AUTO segmentation &
	\citet{bilitski-etal-2026-automatic} &
	\url{https://github.com/Intellexus-DSI/alloauto} &
	The first annotated corpus for separating native Tibetan from translated passages in Classical Tibetan texts, released with trained models and scripts. \\
	Tib-Bi-Dharma / SenSim &
	\citet{cohen-etal-2026-scaling} &
	\url{https://github.com/Intellexus-DSI/sensim} &
	Tib-Bi-Dharma, a Classical Tibetan sentence-similarity model, released with an expert gold STS set, two automatically scored STS sets and the pipeline code. \\
	Vocabulary scaling &
	\citet{han-etal-2026-scaling} &
	\url{https://github.com/White2360/vocab-scaling-low-resource}\newline
	\url{https://www.modelscope.cn/collections/lingfeng2360/ACL-2026-Findings-Vocab-Scaling} &
	The code repository for a study of joint Mongolian--Tibetan--Uyghur vocabulary size under BPE and byte-level BPE. \\
	TibOCR-Bench / TibOCR-Train &
	\citet{jie_tibocr-bench_2026} &
	\url{https://doi.org/10.57760/sciencedb.28968} &
	A multimodal Tibetan OCR benchmark covering 30 real-world scenarios plus a preference-aligned training set, which the authors state they release together with the data pipeline, checkpoints, and training and inference code. \\
	CMiLBench &
	\citet{li-etal-2026-diversity} &
	\url{https://github.com/CMLI-NLP/CMiLBench} &
	A hierarchical benchmark (24,663 instances, 17 tasks and 5 levels) testing ability, culture and safety in Tibetan, Mongolian and Uyghur. \\
	tibetan-wer / tibetan-word-segmentation-ds &
	\citet{moore_evaluating_2026} &
	\url{https://pypi.org/project/tibetan-wer/}\newline
	\url{https://huggingface.co/datasets/billingsmoore/tibetan-word-segmentation-ds} &
	A Python package for character, syllable and segmented word error rate (SWER) that scores Tibetan ASR output at the word level, released with expert word-boundary annotations of 100 NICT-Tib1 utterances (2,634 inter-syllable positions) for measuring agreement between automatic segmenters and a human annotator. \\
	FTibSuite &
	\citet{xu-etal-2026-ftibsuite} &
	\url{https://huggingface.co/onedday/FTib-VLM}\newline
	\url{https://huggingface.co/datasets/onedday/FTib-data} &
	FTibVLM, a Tibetan vision--language model built on Qwen3-VL-8B-Instruct, released with FTibData for pre-training, alignment and instruction tuning. \\
	TibHCR &
	\citet{zhou_tibhcr_2026} &
	\url{https://huggingface.co/datasets/qixiaoke/TibHCR}\newline
	\url{https://github.com/qixiaoke007/TibHCR} &
	A large-scale offline handwritten Tibetan character dataset of 141,698 samples from 235 writers across 47 classes, benchmarked with CRNN and a PP-OCRv4 cross-lingual fine-tuning baseline. \\
\end{xltabular}
\endgroup
\twocolumn

\onecolumn
\begingroup
\scriptsize
\hbadness=10000
\setlength{\tabcolsep}{4pt}
\renewcommand{\arraystretch}{1.12}
\setlength{\LTcapwidth}{\textwidth}
\captionsetup{position=below}
\begin{xltabular}{\textwidth}{@{}%
		>{\raggedright\arraybackslash}p{0.18\textwidth}%
		>{\raggedright\arraybackslash}p{0.09\textwidth}%
		>{\raggedright\arraybackslash}p{0.18\textwidth}%
		>{\raggedright\arraybackslash}X@{}}
	\toprule
	\textbf{Language} & \textbf{ISO 639-3} & \textbf{Search Keyword} & \textbf{Publications} \\
	\midrule
	\endfirsthead
	\multicolumn{4}{@{}l}{\textit{(continued from previous page)}} \\
	\toprule
	\textbf{Language} & \textbf{ISO 639-3} & \textbf{Search Keyword} & \textbf{Publications} \\
	\midrule
	\endhead
	\midrule
	\multicolumn{4}{r@{}}{\textit{(continued on next page)}} \\
	\endfoot
	\bottomrule
	\caption{The 82 Bodish languages and the NLP publications retrieved for each. Language names and ISO~\mbox{639-3} codes are taken from Ethnologue. The retrieval method follows Section~\ref{sec_method}, with the search keyword in place of \textit{Tibetan}. Entries that differ only in a qualifier share a keyword and hence their publications: those on Tibetan form the survey corpus (Appendices~\ref{appendix_acl} and~\ref{appendix_dblp}) and are not repeated here, and those on Tamang are listed under Eastern Tamang. Elsewhere, an empty cell means that no publication was retrieved.}\label{tab:bodish-languages} \\
	\endlastfoot
	\multicolumn{4}{@{}l}{\textit{Bodish, no further subgroup (5)}} \\*
	Gongduk & \texttt{goe} & Gongduk & \\
	Lhokpu & \texttt{lhp} & Lhokpu & \\
	Monpa, Kalaktang & \texttt{kkf} & Monpa & \\
	Olekha & \texttt{ole} & Olekha & \\
	Tshangla & \texttt{tsj} & Tshangla & \\
	\addlinespace
	\multicolumn{4}{@{}l}{\textit{Central Bodish (38)}} \\*
	Balti & \texttt{bft} & Balti & \citet{sharif_balti-tamko_2025}. \\
	Brokkat & \texttt{bro} & Brokkat & \\
	Brokpake & \texttt{sgt} & Brokpake & \\
	Changthang & \texttt{cna} & Changthang & \\
	Chocangacakha & \texttt{cgk} & Chocangacakha & \\
	Choni & \texttt{cda} & Choni & \\
	Dolpo & \texttt{dre} & Dolpo & \\
	Dzongkha & \texttt{dzo} & Dzongkha & \citet{wangchuk_next_2022,dhungyel_analysing_2017,arun_semantic_2014,dasgupta-etal-2012-forward,norbu-etal-2010-dzongkha,chungku-etal-2010-building}. \\
	Groma & \texttt{gro} & Groma & \\
	Gyalsumdo & \texttt{gyo} & Gyalsumdo & \\
	Humla & \texttt{hut} & Humla & \\
	Hyolmo & \texttt{scp} & Hyolmo & \\
	Jad & \texttt{jda} & Jad & \\
	Jirel & \texttt{jul} & Jirel & \\
	Kyirong & \texttt{kgy} & Kyirong & \\
	Ladakhi & \texttt{lbj} & Ladakhi & \citet{thukroo_hybrid_2023}. \\
	Lakha & \texttt{lkh} & Lakha & \\
	Layakha & \texttt{lya} & Layakha & \\
	Lhomi & \texttt{lhm} & Lhomi & \\
	Lhowa & \texttt{loy} & Lhowa & \\
	Lunanakha & \texttt{luk} & Lunanakha & \\
	Mugom-Karmarong & \texttt{muk} & Mugom-Karmarong & \\
	Naaba & \texttt{nao} & Naaba & \\
	Nubri & \texttt{kte} & Nubri & \\
	Purig & \texttt{prx} & Purig & \\
	Sherpa & \texttt{xsr} & Sherpa & \\
	Sikkimese & \texttt{sip} & Sikkimese, Bhutia & \citet{ashida-etal-2020-building}. \\
	Spiti Bhoti & \texttt{spt} & Spiti Bhoti & \\
	Stod Bhoti & \texttt{sbu} & Stod Bhoti & \\
	Syuba & \texttt{syw} & Syuba & \\
	Tibetan, Amdo & \texttt{adx} & Tibetan & \\
	Tibetan, Central & \texttt{bod} & Tibetan & \\
	Tibetan, Khams & \texttt{khg} & Tibetan & \\
	Tichurong & \texttt{tcn} & Tichurong & \\
	Tseku & \texttt{tsk} & Tseku & \\
	Tsum & \texttt{ttz} & Tsum & \\
	Walungge & \texttt{ola} & Walungge & \\
	Zangskari & \texttt{zau} & Zangskari & \\
	\addlinespace
	\multicolumn{4}{@{}l}{\textit{East Bodish (9)}} \\*
	Bumthangkha & \texttt{kjz} & Bumthangkha & \\
	Chalikha & \texttt{tgf} & Chalikha & \\
	Dakpakha & \texttt{dka} & Dakpakha & \\
	Dzalakha & \texttt{dzl} & Dzalakha & \\
	Khengkha & \texttt{xkf} & Khengkha & \\
	Kurtokha & \texttt{xkz} & Kurtokha & \\
	Monpa, Tawang & \texttt{twm} & Monpa & \\
	Nupbikha & \texttt{npb} & Nupbikha & \\
	Nyenkha & \texttt{neh} & Nyenkha & \\
	\addlinespace
	\multicolumn{4}{@{}l}{\textit{West Bodish (14)}} \\*
	Chantyal & \texttt{chx} & Chantyal & \\
	Dura & \texttt{drq} & Dura & \\
	Ghale, Northern & \texttt{ghh} & Ghale & \\
	Ghale, Southern & \texttt{ghe} & Ghale & \\
	Gurung & \texttt{gvr} & Gurung & \\
	Kaike & \texttt{kzq} & Kaike & \\
	Kuke & \texttt{ght} & Kuke & \\
	Nar Phu & \texttt{npa} & Nar Phu & \\
	Nyeshangte & \texttt{nmm} & Nyeshangte & \\
	Seke & \texttt{skj} & Seke & \\
	Tamang, Eastern & \texttt{taj} & Tamang & \citet{ghimire-etal-2026-neptam,acharya-etal-2026-english,chaudhary-etal-2020-efforts}. \\
	Tamang, Eastern Gorkha & \texttt{tge} & Tamang & \\
	Tamang, Western & \texttt{tdg} & Tamang & \\
	Thakali & \texttt{ths} & Thakali & \\
	\addlinespace
	\multicolumn{4}{@{}l}{\textit{West Himalayish (16)}} \\*
	Byangsi & \texttt{bee} & Byangsi & \\
	Chaudangsi & \texttt{cdn} & Chaudangsi & \\
	Darmiya & \texttt{drd} & Darmiya & \\
	Gahri & \texttt{bfu} & Gahri & \\
	Jangshung & \texttt{jna} & Jangshung & \\
	Kanashi & \texttt{xns} & Kanashi & \\
	Kinnauri & \texttt{kfk} & Kinnauri & \\
	Kinnauri, Bhoti & \texttt{nes} & Kinnauri & \\
	Kinnauri, Chhoyul & \texttt{tpq} & Kinnauri & \\
	Kinnauri, Chitkuli & \texttt{cik} & Kinnauri & \\
	Pattani & \texttt{lae} & Pattani & \\
	Rangkas & \texttt{rgk} & Rangkas & \\
	Rongpo & \texttt{rnp} & Rongpo & \\
	Shumcho & \texttt{scu} & Shumcho & \\
	Sunam & \texttt{ssk} & Sunam & \\
	Tinani & \texttt{lbf} & Tinani & \\
\end{xltabular}
\endgroup
\twocolumn

\end{document}